\documentclass[review]{elsarticle}
\usepackage[margin=2.9cm]{geometry}

\usepackage{lipsum}                     
\usepackage{xargs}                      

\usepackage{pdflscape}
\usepackage{url}
\usepackage{verbatim}

\usepackage{graphicx}
\usepackage{subfig}
\usepackage{refcount}
\usepackage{footnote}
\usepackage{multirow}
\usepackage{listings}
\usepackage{multicol}
\usepackage{framed}
\usepackage{listings}
\usepackage[skipbelow=\topskip,skipabove=\topskip]{mdframed}
\mdfsetup{roundcorner=0}
\usepackage{etoolbox}
\usepackage{amssymb}
\usepackage{eurosym}
\usepackage{pifont}
\usepackage{algorithmicx}
\usepackage{algorithm}
\usepackage{algpseudocode}

\usepackage{textcomp}
\usepackage{mathtools}
\usepackage{fixltx2e}

\usepackage{amsmath,xparse,mleftright}

\usepackage[colorlinks = true,
            linkcolor = blue,
            urlcolor  = blue,
            citecolor = blue,
            anchorcolor = blue]{hyperref}

\algnewcommand\algorithmicinput{\textbf{Input:}}
\algnewcommand\INPUT{\item[\algorithmicinput]}

\algnewcommand\algorithmicoutput{\textbf{Output:}}
\algnewcommand\OUTPUT{\item[\algorithmicoutput]}

\usepackage{lipsum}
\makeatletter
\def\ps@pprintTitle{%
 \let\@oddhead\@empty
 \let\@evenhead\@empty
 \def\@oddfoot{}%
 \let\@evenfoot\@oddfoot}
\makeatother

\NewDocumentCommand{\up}{som}{%
  \IfBooleanTF{#1}
    {\upext{#3}}
    {#3\IfNoValueTF{#2}{\mathord}{#2}\uparrow}%
}

\NewDocumentCommand{\down}{som}{%
  \IfBooleanTF{#1}
    {\upext{#3}}
    {#3\IfNoValueTF{#2}{\mathord}{#2}\downarrow}%
}

\usepackage{lineno,hyperref}
\modulolinenumbers[5]
\journal{Expert Systems with Applications}

\biboptions{authoryear}

\begin{document}

\begin{frontmatter}

\title{AutoWeka4MCPS-AVATAR: Accelerating Automated Machine Learning Pipeline Composition and Optimisation}

\author[mymainaddress]{Tien-Dung Nguyen \corref{mycorrespondingauthor}}
\cortext[mycorrespondingauthor]{Corresponding author}
\ead{tiendung.nguyen-2@student.uts.edu.au}
\address[mymainaddress]{Advanced Analytics Institute, University of Technology Sydney, Australia}

\author[mymainaddress]{Bogdan Gabrys}
\ead{bogdan.gabrys@uts.edu.au}

\author[mymainaddress]{Katarzyna Musial}
\ead{katarzyna.musial-gabrys@uts.edu.au}

\begin{abstract}

Automated machine learning pipeline (ML) composition and optimisation aim at automating the process of finding the most promising ML pipelines within allocated resources (i.e., time, CPU and memory).
Existing methods, such as Bayesian-based and genetic-based optimisation, which are implemented in Auto-Weka, Auto-sklearn and TPOT, evaluate pipelines by executing them. Therefore, the pipeline composition and optimisation of these methods frequently require a tremendous amount of time that prevents them from exploring complex pipelines to find better predictive models. 
To further explore this research challenge, we have conducted experiments showing that many of the generated pipelines are invalid in the first place, and attempting to execute them is a waste of time and resources.    
To address this issue, we propose a novel method to evaluate the validity of ML pipelines, without their execution, using a surrogate model (AVATAR).
The AVATAR generates a knowledge base by automatically learning the capabilities and effects of ML algorithms on datasets' characteristics. This knowledge base is used for a simplified mapping from an original ML pipeline to a surrogate model which is a Petri net based pipeline. Instead of executing the original ML pipeline to evaluate its validity, the AVATAR evaluates its surrogate model constructed by capabilities and effects of the ML pipeline components and input/output simplified mappings. Evaluating this surrogate model is less resource-intensive than the execution of the original pipeline. As a result, the AVATAR enables the pipeline composition and optimisation methods to evaluate more pipelines by quickly rejecting invalid pipelines. 
We integrate the AVATAR into the sequential model-based algorithm configuration (SMAC).
Our experiments show that when SMAC employs AVATAR, it finds better solutions than on its own. This is down to the fact that the AVATAR can evaluate more pipelines within the same time budget and allocated resources.

\end{abstract}


\end{frontmatter}


\section{Introduction}
\label{sec:intro}

Automated machine learning (AutoML) aims at automating the process of data analysis, from the collection and integration of data, through composition and optimisation of ML pipelines, to deployment and maintenance of generated solutions~\citep{kaga09,sabu18,zohu19}. Although many existing studies proposed methods to tackle the problem of pipeline composition and optimisation~\citep{sabu16,sabu16b,sabu18,olmo16,fekl15,mowe18,giya18,depi17,tsga12}, one of the drawbacks of the majority of these methods is constructing more invalid pipelines than valid ones. 
There are two main approaches to deal with this issue.


In the first class of approaches, the pipelines' structures, which define the executed order of the pipeline components, use fixed templates~\citep{kaga09,sabu18,fekl15}. Although using fixed structures can reduce the number of invalid pipelines during the composition and optimisation, these approaches limit the exploration of promising pipelines which may have different structures than the ones predefined in the templates.


In the second class of approaches, there have been several attempts to reduce the randomness of pipeline construction by generating specific ML pipeline structures based on constraints. These constraints can be represented by using context-free grammars \citep{depi17,tsga12} or AI planning to guide the construction of pipelines \citep{mowe18,giya18}. Nevertheless, all of these methods evaluate the validity of a pipeline by executing them (which we will refer to as a T-method). After executing a pipeline, if the result is an executable predictive model, the T-method evaluates the pipeline to be valid; otherwise, it is invalid. If a pipeline is complex, the number of preprocessing/predictor components within the pipeline can be high, or if the size of the dataset is large, the evaluation of the pipeline can be computationally expensive. Consequently, the optimisation process will require a significant time budget to find well-performing pipelines. 

To address the drawback of generating invalid pipeline, we propose the AVATAR to evaluate ML pipelines using their surrogate models. The AVATAR transforms a pipeline to its surrogate model and evaluates it instead of executing the original pipeline.
In the proposed approach, the evaluation of the surrogate models requires a knowledge base which is generated from a variety of synthetic datasets automatically.

AutoWeka for multi-component predictive systems (AutoWeka4MCPS) is an AutoML tool which can compose and optimise complex ML pipelines. AutoWeka4MCPS limits the generation of invalid pipelines by using a fixed pipeline template. This template includes the following components in their respective order: missing value handling $\rightarrow$ outlier removal $\rightarrow$ transformation $\rightarrow$ dimensionality reduction $\rightarrow$ sampling $\rightarrow$ predictor $\rightarrow$ meta-predictor. While the found solutions must include at least a predictor, the other data transformation steps are optional. However, if they are included in the evaluated candidate solution, they have to follow the prescribed order. Apart from introducing the fundamental concepts for the AVATAR, we integrate the AVATAR with AutoWeka4MCPS and the SMAC optimisation to demonstrate the effectiveness of the AVATAR to reduce search space by quickly rejecting invalid pipelines. 

To this end, this paper\footnote{This study is a significant extension of our previous work\citep{ngma20}, a conference paper presented at the Eighteenth International Symposium on Intelligent Data Analysis} has three main contributions:

\begin{itemize}
    \item We conduct experiments on current state-of-the-art AutoML tools to show that the construction of a large number of invalid pipelines during the pipeline composition and optimisation may result in overall bad performance of the obtained ML pipeline.

    \item We propose the AVATAR to accelerate the automatic pipeline composition and optimisation by evaluating pipelines using a surrogate model. The AVATAR generates a knowledge base by learning the capabilities and effects of ML components automatically. The AVATAR uses this knowledge base to perform the simplified mapping from an ML pipeline component to its surrogate model, then finds the pipeline validity by evaluating the surrogate model.
    
    
    
    \item We perform extensive experiments to show that the combination of the multiple configuration initialisation and the AVATAR can better stabilise the convergence of SMAC in comparison with the single configuration initialisation. In order to do that, We customise SMAC to initialise with multiple configurations. 
    
\end{itemize}

This paper is divided into five sections. After the Introduction, Section \ref{sec:related_work} reviews previous approaches to representing and evaluating ML pipelines in the context of AutoML.
Section \ref{sec:avatar_method} describes the AVATAR which we propose to evaluate ML pipelines.
Section \ref{sec:experiment} presents experiments to motivate our research and prove the efficiency of the proposed method. Finally, Section \ref{sec:conclusion} concludes this study and points in the direction of the future research in the AutoML area with an emphasis on composition and optimisation of complex ML pipelines.

\section{Related Work}
\label{sec:related_work}

In Section \ref{sec:experiment}, through our experiments, we show that the generation of invalid pipelines has a negative effect that wastes time of the ML pipeline composition and optimisation. In this section, we review previous approaches that aimed at reducing this negative impact. Particularly, we have reviewed the current state-of-the-art pipeline composition and optimisation methods which are implemented as the AutoML tools such as AutoWeka4MCPS \citep{sabu18}, ML-Plan \citep{mowe18}, P4ML \citep{giya18}, TPOT \citep{olmo16} and Auto-sklearn \citep{fekl15}. These approaches can be classified into two main types:

\textit{\textbf{Fixed pipeline structure template}}: The main idea of these approaches is to use fixed pipeline structure templates designed based on the experience of experts. A template consists of component types connected in a fixed order. A component type has ML algorithms with their hyperparameters. When generating a pipeline, there are two main steps. Firstly, a number of component types are selected. Secondly, an algorithm and its hyperparameters are selected for each component type. These approaches are implemented in AutoWeka4MCPS and Auto-sklearn. 

\begin{itemize}
    \item AutoWeka4MCPS implements an automatic pipeline composition and optimisation method of multicomponent predictive systems (MCPS) to deal with the problem of combined algorithm selection and hyperparameter optimisation (CASH) \citep{sabu18}. AutoWeka4MCPS \citep{sabu18} is developed on top of Auto-Weka 0.5 \citep{thhu13}. 
    While Auto-Weka 0.5 only supports single-component pipeline (i.e. predictors/meta-predictors), AutoWeka4MCPS extends this pipeline structure with preprocessing components.  The fixed pipeline structure template of AutoWeka4MCPS is \textit{missing value handling} $\rightarrow$ \textit{outlier removal} $\rightarrow$ \textit{transformation} $\rightarrow$ \textit{dimensionality reduction} $\rightarrow$ \textit{sampling} $\rightarrow$ \textit{predictor} $\rightarrow$ \textit{meta-predictor}. The ML algorithms for these component types are from Weka libraries. The sequential model-based optimisation (SMAC) is used to solve the CASH problem with the extension of the pipeline structure up to seven components.
    
\item Auto-sklearn also implements an automatic pipeline composition and optimisation method to deal with the CASH problem \citep{fekl15}. The fixed pipeline structure template of Auto-sklearn is \textit{one hot encoding} $\rightarrow$ \textit{missing value imputation} $\rightarrow$ \textit{rescaling} $\rightarrow$ \textit{class balancing}  $\rightarrow$ \textit{feature preprocessing} $\rightarrow$ \textit{predictor} $\rightarrow$ \textit{meta-predictor}. Auto-sklearn selects one algorithm for \textit{feature preprocessing} component to perform matrix decomposition, univariate feature selection, classification-based feature selection, feature clustering, kernel approximations, polynomial feature expansion, feature embeddings, or sparse representation transformation. The ML algorithms for these component types are from scikit-learn libraries. Similar to AutoWeka4MCPS, Auto-sklearn employs SMAC to solve the CASH problem. Being different from AutoWeka4MCPS, Auto-sklearn employs a meta-learner that uses datasets's meta-features (e.g., min, max, mean, standard deviation, class entropy, ratio of categorical, numerical and missing values) to find promising initialised configurations, whereas AutoWeka4MCPS uses random initialised configurations.
    
\end{itemize}

\textit{\textbf{Knowledge base to limit pipeline structures}}: These approaches are implemented in ML-Plan, P4ML and TPOT. The main idea of these approaches is to use constraints in forms of context-free grammars,  primitive taxonomy or ad-hoc configurations. These knowledge bases document structures with generic components (e.g., preprocessing $\rightarrow$ predictor) to a multilevel dependency of these generic components (e.g., EMImputation $\rightarrow$ Standardise $\rightarrow$ J48). Although these approaches allow more flexible pipeline structure than using fixed pipeline structure template, the knowledge bases are also designed based on the experience of experts that may not include all possibilities of valid pipelines.

\begin{itemize}
    
    \item TPOT is a pipeline optimisation tool \citep{olmo16} which implements a genetic-programming-based optimisation method \citep{deb2002fast}. TPOT generates pipelines which have data preprocessing components (e.g., data cleaning, feature selection, feature construction, and feature preprocessing), classification/regression and ensemble component. Theses components are from scikit-learn libraries. Theoretically, the pipeline can have an infinite number of components. However, the authors empirically limited the length of the pipeline to four components. The knowledge base of TPOT for generating pipelines is a Python file. 
    According to TPOT's authors, there are approximately 20\% invalid pipelines generated during the evolutionary stages. While allowing a degree of randomness of mutation and cross-over, which imitates natural evolution, can help TPOT to explore better configurations, evaluations of these invalid pipelines significantly increase the complexity for the composition and optimisation. 

    \item ML-Plan is a pipeline optimisation tool \citep{mowe18a} which employs an AI planning approach, so-called the hierarchical task network (HTN), for pipeline composition and optimisation. This method generates pipelines by performing the depth-first search on an HTN search graph. The knowledge base of ML-Plan for generating pipelines is an ad hoc configuration file to describe HTN search graph. Because of using HTN search space as a knowledge base designed by experts for guiding pipeline composition, this search space may not have better configurations which are not known by the experts yet. Therefore, it reduces the chance to find more promising pipeline structures for a given dataset. On the other hand, this approach can guarantee to generate valid pipelines; therefore, it does not waste optimisation time for evaluating invalid pipelines as in the case of TPOT.
    
    \item P4ML is another pipeline optimisation tool which uses the AI planning approach \citep{giya18}. The knowledge base of P4ML for generating pipelines is a primitive taxonomy. This taxonomy describes the dependencies between primitives (i.e., ML algorithms). P4ML constructs pipelines sequentially within 5 phases. Phase 1 uses primitives for data preprocessing. Phase 2 identifies potential classifiers. Phase 3 validates the pipeline's requirement. Phase 4 optimises hyperparameters of the pipeline. Phase 5 creates ensembles of pipelines. However, the planning-based optimisation algorithm is not presented in the study. It also lacks a detailed description of a pipeline structure. 
    
\end{itemize}

Although AutoWeka4MCPS, ML-Plan, P4ML, TPOT and Auto-sklearn evaluate pipelines by executing them, these methods have strategies to limit the generation of invalid pipelines. Auto-sklearn and AutoWeka4MCPS use a fixed pipeline structure template. TPOT, ML-Plan and P4ML use a knowledge base to limit pipeline structures, which are designed manually, to guide the construction of pipelines. Although these approaches can reduce the number of invalid pipelines, our experiments showed that the time wasted for evaluation of the invalid pipelines is significant. Moreover, using fixed templates or a knowledge base reduces search spaces of potential better performing pipelines previously not known by experts, which is a drawback during pipeline composition and optimisation. Our proposed method can automatically generate a knowledge base and reuse this knowledge base to quickly evaluate the validity of pipelines using a surrogate model. This knowledge base can be extended further with new ML components using the same methodology as the one described in the next section when automatically generating the knowledge base in the first place.
\section{AVATAR -- Evaluation of ML Pipelines Using Surrogate Models}
\label{sec:avatar_method}

To address the challenges connected with the time-consuming ML pipelines composition, optimisation and evaluation, we propose the AVATAR\footnote{\url{https://github.com/UTS-AAi/AVATAR}}. Its goal is to speed up the process by evaluating the surrogate pipelines which do not require the execution of the pipeline using the training, validation or test data.
The main idea of the AVATAR is to extend the representation of MCPS introduced in \citep{sabu17b}. The MCPS (i.e., ML pipelines), which are generated by AutoWeka4MCPS, are represented using Petri nets \citep{sabu17b}.
A Petri net is a mathematical modelling language used to represent pipelines \citep{sabu18} as well as data service compositions \citep{tafa10}.
The main idea of Petri nets is to represent transitions of states of a system. Therefore, a Petri net can be used to describe the semantic as well as execution states (i.e., how data is transformed after going through each ML component) of a pipeline. The AVATAR replaces functions of Petri nets' transitions, which in \cite{sabu17b} represent full ML components, by functions to calculate transformations of datasets' characteristics. 

The AVATAR uses a surrogate model in the form of a Petri net. This surrogate pipeline keeps the structure of the original pipeline, replaces the datasets in the form of data matrices (i.e., components' input/output simplified mappings) by the matrices of dataset-characteristics, and the ML components by transition functions to calculate the output from the input tokens (i.e., the matrices of dataset-characteristics). Because of the simplicity of the surrogate pipelines in terms of the size of the tokens and the simplicity of the transition functions, the evaluation of these pipelines is substantially less expensive than the original ones.

\subsection{The Validity of Pipelines}
We define the validity of a pipeline as follows. Given a pipeline $p$, a dataset $d$ and allocated resources~$r$ (i.e., CPU and memory), if we can execute the pipeline successfully and produce an output model, this pipeline is valid. Otherwise, it is not valid.

\begin{equation}
validity_{p} = exec(p,d,r)
\end{equation}

By this definition, there are three factors (i.e., the pipeline, the dataset and the allocated resources) that impact the validity of a pipeline. 

\begin{itemize}
    \item Pipeline and dataset - these two factors directly relate to each other. Generating a valid pipeline requires to select ML components so that the output of the previous ML component is compatible with the next ML component. This compatibility depends on the characteristics of the data. For example, let's assume that we have a dataset with missing values and two pipelines. The first pipeline is \textit{IndependentComponents} $\rightarrow$ \textit{J48}. The second pipeline is \textit{EMImputation}  $\rightarrow$ \textit{IndependentComponents} $\rightarrow$ \textit{J48}. The first pipeline is invalid because the component \textit{IndependentComponents} can not handle a dataset with missing values. The second pipeline is valid because the component \textit{EMImputation} fills in missing values. Therefore, the output of \textit{EMImputation} is another dataset without missing values, and \textit{IndependentComponents} can work with this dataset.
    
    \item Allocated resources - relate to the execution environment of the pipeline. The execution may require more resources (i.e., CPU and memory) than allocated ones due to the high complexity of ML components and large datasets. As a result, this execution may fail, leading to an invalid pipeline due to allocated resources.
    
\end{itemize}

\begin{table}[htb!]
\centering

\vspace{-0.35cm}
\caption {Descriptions of the data-characteristics.}
\label{tab:data_characteristics}

\centering
\small
\begin{tabular}{|l|l|}
\hline
\textbf{Data-characteristics}                                          & \textbf{Description}                        \\ \hline
BINARY\_CLASS                                                         & a dataset has binary classes                \\ \hline
NUMERIC\_CLASS                                                         & a dataset has numeric classes                \\ \hline
DATE\_CLASS                                                           & a dataset has date classes                  \\ \hline
MISSING\_CLASS\_VALUES                                                & a dataset has missing values in classes     \\ \hline
NOMINAL\_CLASS                                                        & a dataset has nominal classes               \\ \hline
SYMBOLIC\_CLASS                                                       & a dataset has symbolic data in classes      \\ \hline
STRING\_CLASS                                                         & a dataset has string classes                \\ \hline
UNARY\_CLASS                                                          & a dataset has unary classes                 \\ \hline
BINARY\_ATTRIBUTES                                                    & a dataset has binary attributes             \\ \hline
DATE\_ATTRIBUTES                                                      & a dataset has date attributes               \\ \hline
\begin{tabular}[c]{@{}l@{}}EMPTY\_NOMINAL\_\\ ATTRIBUTES\end{tabular} & a dataset has an empty column               \\ \hline
MISSING\_VALUES                                                       & a dataset has missing values in attributes  \\ \hline
NOMINAL\_ATTRIBUTES                                                   & a dataset has nominal attributes            \\ \hline
NUMERIC\_ATTRIBUTES                                                   & a dataset has numeric attributes            \\ \hline
UNARY\_ATTRIBUTES                                                     & a dataset has unary attributes              \\ \hline
PREDICTIVE\_MODEL                                                     & a predictive model generated by a predictor \\ \hline
\end{tabular}
\vspace{-0.35cm}

\end{table}

Because the validity of pipelines depends on the characteristics of the data, we call these characteristics as dataset-characteristics. They can be changed depending on the transformations of a given dataset by ML components. Table \ref{tab:data_characteristics} describes the dataset-characteristics used for this knowledge base. We select these dataset-characteristics because the validity of a pipeline for a given dataset depends on these characteristics. These dataset-characteristics are extracted from the capabilities of Weka algorithms\footnote{\url{http://weka.sourceforge.net/doc.dev/weka/core/Capabilities.html}}.

\subsection{The AVATAR Architecture}

\begin{figure}[htb!]
\centering
\includegraphics[width=0.8\linewidth]{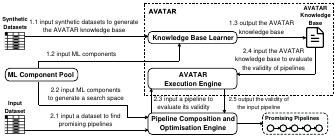}
\vspace{-0.3cm}
\caption{The overall architecture of the AVATAR}
\label{fig:avatar_architecture}
\end{figure}

Figure \ref{fig:avatar_architecture} shows the overall architecture of the AVATAR. The arrows represent data flows. There are two main flows in the architecture.

\begin{itemize}
    
    \item \textit{Learning the AVATAR knowledge base automatically (i.e., this data flow is denoted 1.x):} The {Knowledge Base Learner} employs synthetic datasets to learn the compatibility of each ML component in \textit{ML Component Pool} (i.e., the set of ML components for the pipeline composition and optimisation) with characteristics of data, and the transformations of these characteristics after being executed by this component. This knowledge is stored in the \textit{AVATAR Knowledge Base}.
    
    \item \textit{Evaluating the validity of pipelines (i.e., this data flow is denoted 2.x):} The {AVATAR Execution Engine} employs the \textit{AVATAR Knowledge Base} to evaluate the validity of pipelines. Given an input dataset and the \textit{ML Component Pool}, the \textit{pipeline Composition and Optimisation Engine} (e.g., SMAC) finds the most promising pipelines. Before evaluating the quality of pipelines by executing them, the \textit{Pipeline Composition and Optimisation Engine} employs the \textit{AVATAR Execution Engine} to quickly evaluate the validity of these pipelines.    
    
\end{itemize}

\subsection{Learning the AVATAR knowledge base automatically}
\label{subsec:metaknowledge}

First of all, we discuss how to create \textit{Synthetic Datasets}, machine learning algorithms in the \textit{ML Component Pool}, and the structure of {AVATAR Knowledge Base}. After that, we present the algorithm, which is the core of the \textit{Knowledge Base Learner}, to generate the {AVATAR Knowledge Base} from the \textit{Synthetic Datasets} and \textit{ML Component Pool} automatically.  

\subsubsection{Synthetic Datasets}

To generate the \textit{AVATAR Knowledge Base}, we propose to use synthetic datasets\footnote{\url{https://github.com/UTS-AAi/AVATAR/tree/master/synthetic-datasets}} to isolate dataset-characteristics in each dataset to evaluate which and how dataset-characteristics impact on the capabilities and effects of ML components.  
Real-world datasets usually have many dataset-characteristics that make them not suitable for our purpose. To generate the {AVATAR Knowledge Base}, one task of the AVATAR is to evaluate the changes of a dataset before and after passing through an ML component. If the dataset has many dataset-characteristics, it is difficult to understand which dataset-characteristics cause these changes.
We minimise the number of available dataset-characteristics in each synthetic dataset so that the knowledge base can be applicable in a variety of pipelines and datasets. For example, to learn the dataset-characteristic \textit{MISSING\_VALUES}, we create two datasets for classification and regression problems respectively. One dataset has one missing value attribute and one numeric class for the regression problem. Another dataset has one missing value attribute and one nominal class for the classification problem. Note that our approach can be extended with other dataset-characteristics by creating synthetic datasets containing these dataset-characteristics and numeric/nominal classes using the above methodology.

\subsubsection{ML Component Pool}

\begin{table}[htb!]

\vspace{-0.35cm}
\caption {The list of ML Components.}
\label{tab:ml_component_pool}

\centering
\scriptsize

\begin{tabular}{|l|l|l|}
\hline
\textbf{ML Component}                                     & \textbf{Data Preprocessing} & \textbf{Predictor/meta-predictor} \\ \hline
weka.filters.unsupervised.attribute.ReplaceMissingValues  & \checkmark   &                                   \\ \hline
weka.filters.unsupervised.attribute.EMImputation          & \checkmark   &                                   \\ \hline
weka.filters.supervised.instance.ClassBalancer            & \checkmark   &                                   \\ \hline
weka.filters.supervised.instance.SpreadSubsample          & \checkmark   &                                   \\ \hline
weka.filters.unsupervised.attribute.PrincipalComponents   & \checkmark   &                                   \\ \hline
weka.filters.unsupervised.instance.Resample               & \checkmark   &                                   \\ \hline
weka.filters.unsupervised.instance.ReservoirSample        & \checkmark   &                                   \\ \hline
weka.filters.unsupervised.instance.RemovePercentage       & \checkmark   &                                   \\ \hline
weka.filters.unsupervised.instance.PeriodicSampling       & \checkmark   &                                   \\ \hline
weka.filters.supervised.attribute.AttributeSelection      & \checkmark   &                                   \\ \hline
weka.filters.unsupervised.attribute.Center                & \checkmark   &                                   \\ \hline
weka.filters.unsupervised.attribute.Standardize           & \checkmark   &                                   \\ \hline
weka.filters.unsupervised.attribute.Normalize             & \checkmark   &                                   \\ \hline
weka.filters.unsupervised.attribute.IndependentComponents & \checkmark   &                                   \\ \hline
weka.filters.unsupervised.attribute.Discretize            & \checkmark   &                                   \\ \hline
weka.filters.unsupervised.attribute.NominalToBinary       & \checkmark   &                                   \\ \hline
weka.filters.unsupervised.attribute.NumericToBinary       & \checkmark   &                                   \\ \hline
weka.filters.unsupervised.attribute.NumericToNominal      & \checkmark   &                                   \\ \hline
weka.filters.unsupervised.attribute.StringToNominal       & \checkmark   &                                   \\ \hline
weka.classifiers.bayes.NaiveBayes                         &                             & \checkmark         \\ \hline
weka.classifiers.bayes.NaiveBayesUpdateable               &                             & \checkmark         \\ \hline
weka.classifiers.functions.LinearRegression               &                             & \checkmark         \\ \hline
weka.classifiers.functions.Logistic                       &                             & \checkmark         \\ \hline
weka.classifiers.functions.SimpleLogistic                 &                             & \checkmark         \\ \hline
weka.classifiers.functions.GaussianProcesses              &                             & \checkmark         \\ \hline
weka.classifiers.functions.MultilayerPerceptron           &                             & \checkmark         \\ \hline
weka.classifiers.functions.SimpleLinearRegression         &                             & \checkmark         \\ \hline
weka.classifiers.functions.SMOreg                         &                             & \checkmark         \\ \hline
weka.classifiers.functions.SMO                            &                             & \checkmark         \\ \hline
weka.classifiers.lazy.IBk                                 &                             & \checkmark         \\ \hline
weka.classifiers.lazy.KStar                               &                             & \checkmark         \\ \hline
weka.classifiers.lazy.LWL                                 &                             & \checkmark         \\ \hline
weka.classifiers.meta.AdaBoostM1                          &                             & \checkmark         \\ \hline
weka.classifiers.meta.AttributeSelectedClassifier         &                             & \checkmark         \\ \hline
weka.classifiers.meta.Bagging                             &                             & \checkmark         \\ \hline
weka.classifiers.meta.ClassificationViaRegression         &                             & \checkmark         \\ \hline
weka.classifiers.meta.LogitBoost                          &                             & \checkmark         \\ \hline
weka.classifiers.meta.RandomCommittee                     &                             & \checkmark         \\ \hline
weka.classifiers.meta.RandomSubSpace                      &                             & \checkmark         \\ \hline
weka.classifiers.rules.DecisionTable                      &                             & \checkmark         \\ \hline
weka.classifiers.rules.JRip                               &                             & \checkmark         \\ \hline
weka.classifiers.rules.OneR                               &                             & \checkmark         \\ \hline
weka.classifiers.rules.PART                               &                             & \checkmark         \\ \hline
weka.classifiers.rules.ZeroR                              &                             & \checkmark         \\ \hline
weka.classifiers.trees.DecisionStump                      &                             & \checkmark         \\ \hline
weka.classifiers.trees.J48                                &                             & \checkmark         \\ \hline
weka.classifiers.trees.LMT                                &                             & \checkmark         \\ \hline
weka.classifiers.trees.RandomForest                       &                             & \checkmark         \\ \hline
weka.classifiers.trees.RandomTree                         &                             & \checkmark         \\ \hline
weka.classifiers.trees.REPTree                            &                             & \checkmark         \\ \hline
\end{tabular}

\vspace{-0.35cm}

\end{table}

The \textit{ML Component Pool} is a set of ML components. These components are data preprocessing algorithms and predictors/meta-predictors. Table \ref{tab:ml_component_pool} presents ML components that are used in this study. We use Weka libraries for the implementation of the \textit{ML Component Pool}.
The \textit{ML Component Pool} plays two roles in the AVATAR architecture. Firstly, the \textit{ML Component Pool} is used for the {Knowledge Base Learner} to learn how ML components transform dataset-characteristics and store the knowledge into the \textit{AVATAR Knowledge Base}. Secondly, the \textit{ML Component Pool} is used by the \textit{Pipeline Composition and Optimisation Engine} to build search spaces.

\subsubsection{The AVATAR Knowledge Base}

The purpose of the \textit{AVATAR Knowledge Base}\footnote{\url{https://github.com/UTS-AAi/AVATAR/blob/master/avatar-knowledge-base/avatar_knowledge_base.json}} is to describe the logic of transition functions of the surrogate pipelines. The logic includes the capabilities and effects of ML components (i.e., pipeline components). The capabilities and effects are constructed by dataset-characteristics. Listing \ref{lst:knowledge_base_example} shows a segment of the AVATAR knowledge that describes the capabilities and effects of \textit{EMImputation}.

The capabilities are used to verify whether an algorithm can be applied to a given dataset without making any changes to either of them. For example, whether the linear regression algorithm can be used on data with missing values and numeric attributes or not? The capabilities for a given algorithm are represented as a list of dataset-characteristics. The value of each capability-related dataset-characteristic is either 0 (i.e., the algorithm cannot work with the dataset which has this dataset-characteristic) or 1 (i.e., the algorithm can work with the dataset which has this dataset-characteristic). Based on the capabilities, we can determine which components of a pipeline (i.e., ML algorithms) are not able to process specific dataset-characteristics of a dataset. 
For example, \textit{EMImputation} is compatible with missing values. Therefore, we have \textit{MISSING\_VALUES = 1} in the capabilities.

The effects describe data transformations. Similar to the capabilities, the effects are represented by a list of dataset-characteristics. Each effect-related dataset-characteristic can have three values, 0 (i.e., does not transform this dataset-characteristic), 1 (i.e., transforms one or more attributes/classes to this dataset-characteristic), or -1 (i.e., disables the effect of this dataset-characteristic on one or more attributes/classes).
For example, \textit{EMImputation} fills in all missing values of attributes. As a result, the output dataset of \textit{EMImputation} does not have missing values. Therefore, we have \textit{MISSING\_VALUES = -1} in the effects.

\begin{minipage}{\linewidth}

\begin{lstlisting}[frame=single,basicstyle=\scriptsize,caption={A segment of the AVATAR Knowledge Base},label={lst:knowledge_base_example}]
...
    "componentId" : "weka.filters.unsupervised.attribute.EMImputation",
    "componentName" : "EMImputation",
    "listOfCapabilities" : [ {
      "mLComponentCapability" : "NOMINAL_CLASS",
      "value" : 0
    }, {
      "mLComponentCapability" : "NUMERIC_CLASS",
      "value" : 1
    }, {
      "mLComponentCapability" : "MISSING_VALUES",
      "value" : 1
    }, {
      "mLComponentCapability" : "NOMINAL_ATTRIBUTES",
      "value" : 0
    }, {
      "mLComponentCapability" : "NUMERIC_ATTRIBUTES",
      "value" : 1
    }
    ...
    ],
    "listOfEffects" : [ {
      "mLComponentCapability" : "NOMINAL_CLASS",
      "value" : 0
    }, {
      "mLComponentCapability" : "NUMERIC_CLASS",
      "value" : 0
    }, {
      "mLComponentCapability" : "MISSING_VALUES",
      "value" : -1
    }, {
      "mLComponentCapability" : "NOMINAL_ATTRIBUTES",
      "value" : 0
    }, {
      "mLComponentCapability" : "NUMERIC_ATTRIBUTES",
      "value" : 0
    } 
    ...
    ],
...
\end{lstlisting}

\end{minipage}

\subsubsection{The Knowledge Base Learner}

\label{sec:algorithm_generating_metaknowledge}

\begin{figure}[htb!]
\centering
\includegraphics[width=0.75\linewidth]{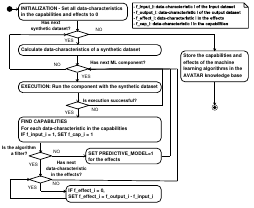}
\vspace{-0.3cm}
\caption{Algorithm to generate the knowledge base for evaluating surrogate pipelines.}
\label{fig:generate_meta_knowledge}
\end{figure}

The \textit{Knowledge Base Learner} implements an algorithm to generate the \textit{AVATAR Knowledge Base} from the \textit{ML Component Pool} and \textit{Synthetic Datasets}. 
Figure \ref{fig:generate_meta_knowledge} presents the algorithm to generate the \textit{AVATAR Knowledge Base}. This algorithm has four main stages:

\begin{enumerate}

    \item Initialisation: The first stage initialises all dataset-characteristics available in the lists of capabilities and effects to 0.
    \item Execution: Run ML components with every synthetic dataset and get outputs (i.e., output datasets or predictive models).
    \item Find capabilities: If the execution is successful, the algorithm calculates the values of dataset-characteristics of the input dataset. After that, the algorithm sets the active dataset-characteristics (i.e., the dataset-characteristics have the value 1) of the input dataset for the ones in the capabilities. For example, the algorithm executes \textit{EMImputation} with a dataset successfully. The algorithm calculates the values of dataset-characteristics of the input dataset. One of the dataset-characteristics is MISSING\_VALUE\_ATTRIBUTE which equals 1. The algorithm sets MISSING\_VALUE\_ATTRIBUTE to 1 in the capabilities. It means that \textit{EMImputation} can work with a dataset containing missing values.
    
    \item Find effects: If an ML component is a predictor/meta-predictor, we set \textit{PREDICTIVE\_MODEL} for its effects. If the ML component is a data preprocessing component and its current value is a default value (the default value is 0 by the initialisation), we set this effect-related dataset-characteristic to the value of a difference between the value of this dataset-characteristic of the output and input dataset. For example, the algorithm executes \textit{EMImputation} with an input dataset successfully. This input dataset has missing values (i.e., the token of the input dataset has MISSING\_VALUE\_ATTRIBUTE=1), and the output dataset does not have a missing value (i.e., the token of the output dataset has MISSING\_VALUE\_ATTRIBUTE=0) because the missing values are filled in by \textit{EMImputation}. The value of MISSING\_VALUE\_ATTRIBUTE in the effect of \textit{EMImputation} equals 0-1=-1.

\end{enumerate}

The \textit{AVATAR Knowledge Base} can be extended with the knowledge for others ML components by adding the effects and capabilities of these components manually, or automatically by using the \textit{Knowledge Base Learner}. The \textit{AVATAR Knowledge Base} can also be extended with other dataset-characteristics by creating synthetic datasets to isolate these dataset-characteristics and functions to calculate the values of these dataset-characteristics, then using the \textit{Knowledge Base Learner} to learn these new dataset-characteristics for all ML Components in the \textit{ML Component Pool}. 
  
\subsection{Evaluating the validity of pipelines}

The {AVATAR Execution Engine} receives requests from the {Pipeline Composition and Optimisation Engine} to evaluate the validity of the pipelines. First of all, it maps the ML pipeline to its surrogate pipeline. After that, it evaluates this surrogate pipeline by firing all transitions.
We claim that a Petri net is the most promising method to represent a surrogate pipeline. The reason is that it is fast to verify the validity of a Petri net based on simplified ML pipeline.

\begin{figure}[htb!]
\centering
\includegraphics[width=1.0\linewidth]{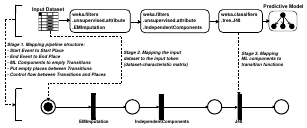}
\vspace{-0.3cm}
\caption{Mapping a ML pipeline to its surrogate model.}
\label{fig:mapping_surrogate}
\end{figure}

\subsubsection{Mapping an ML pipeline to its surrogate model}
The AVATAR maps an ML pipeline to a Petri net pipeline in three steps as presented in Figure~\ref{fig:mapping_surrogate}.

\begin{enumerate}

\item The structure of the original ML pipeline is mapped to the respective structure of the Petri net surrogate pipeline. The start and end events are mapped to the start and end places, respectively. The components are mapped to empty transitions. Empty places (i.e., places without tokens) are put between all transitions. Finally, all flows are mapped to arcs.

\item The values of dataset-characteristics are calculated from the input dataset to form a dataset-characteristic matrix which is the input token in the start place of the surrogate pipeline.

\item The transition functions are mapped from the original components. In this stage, only the corresponding component information (i.e., component name, identifier and parameters) is mapped to the transition function.    
\end{enumerate}

\subsubsection{Evaluating a surrogate model}

\begin{figure}[htb]
\centering
\includegraphics[width=0.75\linewidth]{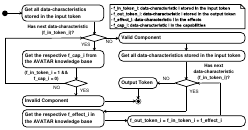}
\vspace{-0.3cm}
\caption{Algorithm for firing a transition of the surrogate model.}
\label{fig:alg_transition}
\end{figure}

The evaluation of a surrogate model executes a Petri net pipeline. This execution starts by firing each transition of the Petri net pipeline and transforming the input token.
As shown in Figure \ref{fig:alg_transition}, firing a transition consists of two tasks: (i) the evaluation of the capabilities of each component and (ii) the calculation of the output token.
The first task verifies the validity of the component using the following rules. If the value of a dataset-characteristic stored in the input token ($f\_in\_token\_i$) is 1 and the corresponding dataset-characteristic in the component's capabilities ($f\_cap\_i$) is 0, this component is invalid. Otherwise, this component is valid. If a component is invalid, the surrogate pipeline is evaluated as invalid.
For example, if the input dataset has missing values in a class (i.e., MISSING\_CLASS\_VALUES=1) and the component can handle missing values in a class (i.e., MISSING\_CLASS\_VALUES=1), the component is valid. 
If the input dataset has missing values in a class (i.e., MISSING\_CLASS\_VALUES=1) and the component cannot handle missing values in a class (i.e., MISSING\_CLASS\_VALUES=0), the component is invalid. 
If the input dataset does not have missing values in a class (i.e., MISSING\_CLASS\_VALUES=0), the component is always valid.

The second task calculates each dataset-characteristic stored in the output token (\textit{f\_out\_token\_i}) in the next place from the input token by adding the value of a dataset-characteristic stored in the input token (\textit{f\_in\_token\_i}) and the respective dataset-characteristic in the component's effects (\textit{f\_effect\_i}). 
For example, it is assumed that an effect-related dataset-characteristic $X$ is MISSING\_CLASS\_VALUES. 
If the input dataset has missing values in a class (i.e., $MISSING\_CLASS\_VALUES = 1$) and the component can remove/replace missing values in a class (i.e., $MISSING\_CLASS\_VALUES = -1$), so that the output dataset will not have missing values in a class (the dataset-characteristic of the output token is 1-1=0).
If the input dataset has missing values in a class (i.e., the dataset-characteristic $X$ of the input token is 1) and the algorithm of the component cannot remove/replace missing values in a class (i.e., $MISSING\_CLASS\_VALUES = 0$), the output dataset still has missing values in a class (i.e., the dataset-characteristic $X$ of the output token is 1+0=1). If the input dataset has no missing values in a class (i.e., the dataset-characteristic $X$ of the input token is 0), the output dataset also has no missing value in a class (i.e., 0-1=-1). In this case, because the values of dataset-characteristics of a dataset are either 0 or 1, the minimum value 0 is set to $MISSING\_CLASS\_VALUES$.

\subsection{An example to illustrate the pipeline evaluation of the AVATAR}

\begin{figure}[htb]
\centering
\includegraphics[width=1.0\linewidth]{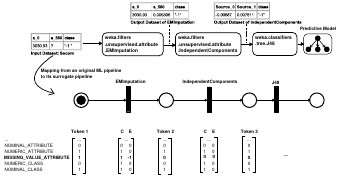}
\vspace{-0.8cm}
\caption{An example of evaluating a ML pipeline using the AVATAR}
\label{fig:avatar_example}

\end{figure}

Figure \ref{fig:avatar_example} shows an example of how the AVATAR works. The original pipeline is an ML pipeline including three components: EMImputation, IndependentComponents and J48. The input dataset is $secom$ which has numeric attributes, missing values in these numeric attributes and nominal values in the class/target attribute. We show the first row of two selected attributes and the target column of this dataset. We also show the first row of the transformed datasets after data preprocessing steps.

Firstly, the AVATAR maps the ML pipeline to the Petri net based pipeline. The input token for the Petri net based pipeline is \textit{token 1}. The values of NOMINAL\_ATTRIBUTE and NUMERIC\_CLASS equal 0. It means that the input data does not have nominal attribute and numeric class. The values of NUMERIC\_ATTRIBUTE, MISSING\_VALUE\_ATTRIBUTE and NOMINAL\_CLASS equal 1. It means that the input data have numeric attribute, missing value attribute and nominal class. The capability (C) and effect (E) of each ML component is retrieved from the AVATAR Knowledge Base.  

Secondly, executing the Petri net based pipeline will fire all transitions. For example, token 1 has MISSING\_VALUE\_ATTRIBUTE=1. EMImputation has C of  MISSING\_VALUE\_ATTRIBUTE=1. It means that EMImputation can work with missing value attributes. EMImputation has E of MISSING\_VALUE\_ATTRIBUTE=-1. It means that after passing token 1 to EMImputation, the value of MISSING\_VALUE\_ATTRIBUTE is 1 + (-1) = 0 (i.e., the missing values are removed).

If IndependentComponents is put in front of EMImputation, we will have an invalid pipeline. This is because the IndependentComponents has C of MISSING\_VALUE\_ATTRIBUTE=0, which means it cannot handle missing value attributes.

\section{Experiments and Discussion}
\label{sec:experiment}

To show the capabilities of the AVATAR, we have conducted four sets of experiments. The first set of experiments investigates the time required (wasted) to evaluate invalid pipelines which are generated during the traditional ML pipeline composition and optimisation process.
After that, we have conducted the second set of experiments to compare the pipeline evaluation time of the AVATAR and the traditional approaches requiring pipelines executions to validate pipelines. The third set of experiments compares the quality (i.e., classification error) of the most promising pipelines found by SMAC with and without the AVATAR. Finally, the fourth set of experiments proposes and investigates the convergence speed of SMAC using the AVATAR and multiple initialisations of pipeline configurations during a single search for the best ML pipeline. We also provide a supplementary document\footnote{\url{https://github.com/UTS-AAi/autoweka/blob/master/autoweka4mcps/doc/supplementary_document.pdf}} with more details and experimental results which, due to the document size constraints, have not been included in this paper. 

\begin{table}[htb!]

\vspace{-0.35cm}
\caption {Summary of datasets' characteristics: the number of numeric attributes, nominal attributes, the number of distinct classes, instances in training and testing sets.}
\label{tab:exp_datasets}

\centering
\scriptsize

\begin{tabular}{|l|l|l|l|l|l|}
\hline
\textbf{Dataset}  & \textbf{Numeric} & \textbf{Nominal} & \textbf{No of distinct classes} & \textbf{Train} & \textbf{Test} \\ \hline
abalone           & 7                & 1                & 28                              & 2,924          & 1,253         \\ \hline
adult             & 6                & 8                & 2                               & 32,561         & 16,281        \\ \hline
amazon            & 10,000           & 0                & 50                              & 1,050          & 450           \\ \hline
car               & 0                & 6                & 4                               & 1,210          & 518           \\ \hline
cifar10small      & 3,072            & 0                & 10                              & 10,000         & 10,000        \\ \hline
convex            & 784              & 0                & 2                               & 8,000          & 50,000        \\ \hline
dexter            & 20,000           & 0                & 2                               & 420            & 180           \\ \hline
dorothea          & 100,000          & 0                & 2                               & 805            & 345           \\ \hline
gcredit           & 7                & 13               & 2                               & 700            & 300           \\ \hline
gisette           & 5,000            & 0                & 2                               & 4,900          & 2,100         \\ \hline
kddcup & 192              & 38               & 2                               & 35,000         & 15,000        \\ \hline
krvskp            & 0                & 36               & 2                               & 2,238          & 958           \\ \hline
madelon           & 500              & 0                & 2                               & 1,820          & 780           \\ \hline
mnist             & 784              & 0                & 10                              & 12,000         & 50,000        \\ \hline
secom             & 590              & 0                & 2                               & 1,097          & 470           \\ \hline
semeion           & 256              & 0                & 10                              & 1,116          & 477           \\ \hline
shuttle           & 9                & 0                & 7                               & 43,500         & 14,500        \\ \hline
waveform          & 40               & 0                & 3                               & 3,500          & 1,500         \\ \hline
winequality       & 11               & 0                & 11                              & 3,429          & 1,469         \\ \hline
yeast             & 8                & 0                & 10                              & 1,039          & 445           \\ \hline
\end{tabular}

\vspace{-0.35cm}

\end{table}

Table \ref{tab:exp_datasets} summarises characteristics of datasets\footnote{\url{https://archive.ics.uci.edu}} used for experiments. We have selected these datasets as they were used in previous studies \citep{sabu18,olmo16,fekl15} and therefore allow for easier comparative analysis. We have run all the experiments on AWS EC2 $t3a.medium$ virtual machines which have 2 vCPU and 4GB memory. 

\subsection{Experiments to investigate the time required to evaluate invalid pipelines}

\subsubsection{Experiment settings}
The AutoML tools used for the experiments are AutoWeka4MCPS \citep{sabu18} and Auto-sklearn \citep{fekl15}. These tools have been selected because of their abilities to construct and optimise hyperparameters of complex ML pipelines, and they have been empirically proven to be effective in a number of previous studies \citep{sabu18,fekl15,baal18}. However, these previous experiments did not investigate the time needed for the evaluation of invalid pipelines and its impact on the pipeline composition and optimisation process. This is the goal of our first set of experiments.

To investigate this issue, we use five iterations (Iter) for the first set of experiments. Each iteration uses a different seed number.
We set the time budget to 2 hours and the memory to 1GB per iteration.
We evaluate the pipelines produced by the AutoML tools using three criteria: (1) the number of invalid/valid pipelines generated in each iteration, (2) the total evaluation time of invalid/ valid pipelines (seconds) and (3) the wasted evaluation time (\%). The wasted evaluation time is calculated by the percentage of the total evaluation time of invalid pipelines over the total optimisation time. This criterion represents the percentage of the wasted time used to evaluate invalid pipelines.

\subsubsection{Experiment results}

\begin{table}[htb!]

\centering

\vspace{-0.35cm}
\caption {Negative impacts of invalid pipelines using AutoWeka4MCPS. (1): the number of invalid/ valid pipelines, (2): the total evaluation time of invalid/ valid pipelines (s), (3): the wasted evaluation time (\%).}
\label{tab:neg_impact_invalid_pipeline_aw}

\tiny

\begin{tabular}{|l|l|l|l|l|l|l|}
\hline
\textbf{Dataset}                       & \textbf{Criteria} & \textbf{Iter 1} & \textbf{Iter 2} & \textbf{Iter 3} & \textbf{Iter 4} & \textbf{Iter 5} \\ \hline
\multirow{3}{*}{\textbf{abalone}}      & \textbf{(1)}      & 9/54            & 25/66           & 16/53           & 29/97           & 3/13            \\ \cline{2-7} 
                                       & \textbf{(2)}      & 16/7,284        & 8,639/322       & 7,571/286       & 3,361/2,351     & 8,070/74        \\ \cline{2-7} 
                                       & \textbf{(3)}      & 0.21            & 96.40           & 96.36           & 58.84           & 99.10           \\ \hline
\multirow{3}{*}{\textbf{adult}}        & \textbf{(1)}      & 5/25            & 4/23            & 22/23           & 8/25            & 4/19            \\ \cline{2-7} 
                                       & \textbf{(2)}      & 3,982/4,308     & 12/8,188        & 5,493/805       & 7,321/356       & 3,656/6,229     \\ \cline{2-7} 
                                       & \textbf{(3)}      & 48.04           & 0.14            & 87.22           & 95.36           & 36.98           \\ \hline
\multirow{3}{*}{\textbf{amazon}}       & \textbf{(1)}      & 12/23           & 6/4             & 26/0            & 17/19           & 1/0             \\ \cline{2-7} 
                                       & \textbf{(2)}      & 5,413/3,056     & 7,662/66        & 4,928/0         & 7,701/1,689     & 3,603/0         \\ \cline{2-7} 
                                       & \textbf{(3)}      & 63.92           & 99.67           & 100.00          & 82.01           & 100.00          \\ \hline
\multirow{3}{*}{\textbf{car}}          & \textbf{(1)}      & 22/94           & 26/123          & 31/133          & 71/263          & 32/128          \\ \cline{2-7} 
                                       & \textbf{(2)}      & 4,353/1,416     & 8,239/224       & 4,206/5,039     & 2,745/3,614     & 13,092/1,443    \\ \cline{2-7} 
                                       & \textbf{(3)}      & 75.48           & 97.35           & 45.49           & 43.17           & 90.07           \\ \hline
\multirow{3}{*}{\textbf{cifar10small}} & \textbf{(1)}      & 2/6             & 11/30           & 16/0            & 1/6             & 10/17           \\ \cline{2-7} 
                                       & \textbf{(2)}      & 56/9,700        & 6,376/854       & 6,590/0         & 1,474/4,563     & 2,006/7,602     \\ \cline{2-7} 
                                       & \textbf{(3)}      & 0.57            & 88.19           & 100.00          & 24.41           & 20.88           \\ \hline
\multirow{3}{*}{\textbf{convex}}       & \textbf{(1)}      & 5/24            & 4/40            & 5/2             & 2/15            & 10/15           \\ \cline{2-7} 
                                       & \textbf{(2)}      & 3,625/4,309     & 5,640/3,787     & 4,124/2,129     & 2,229/4,968     & 6,622/3,364     \\ \cline{2-7} 
                                       & \textbf{(3)}      & 45.69           & 59.83           & 65.95           & 30.97           & 66.31           \\ \hline
\multirow{3}{*}{\textbf{dexter}}       & \textbf{(1)}      & 14/55           & 3/18            & 4/0             & 0/4             & 6/13            \\ \cline{2-7} 
                                       & \textbf{(2)}      & 1,827/4,246     & 3,604/3,852     & 7,205/0         & 0/8,796         & 7,210/89        \\ \cline{2-7} 
                                       & \textbf{(3)}      & 30.08           & 48.34           & 100.00          & 0.00            & 98.78           \\ \hline
\multirow{3}{*}{\textbf{dorothea}}     & \textbf{(1)}      & 5/17            & 9/3             & 4/0             & 1/0             & 5/16            \\ \cline{2-7} 
                                       & \textbf{(2)}      & 3,627/1,477     & 231/7,261       & 7,208/0         & 3,602/0         & 3,639/3,511     \\ \cline{2-7} 
                                       & \textbf{(3)}      & 71.05           & 3.09            & 100.00          & 100.0           & 50.89           \\ \hline
\multirow{3}{*}{\textbf{gcredit}}      & \textbf{(1)}      & 75/314          & 52/209          & 86/378          & 24/165          & 139/706         \\ \cline{2-7} 
                                       & \textbf{(2)}      & 479/5,513       & 5,233/941       & 1,387/5,908     & 3,363/1,975     & 1,827/2,664     \\ \cline{2-7} 
                                       & \textbf{(3)}      & 8.00            & 84.77           & 19.01           & 63.00           & 40.68           \\ \hline
\multirow{3}{*}{\textbf{gisette}}      & \textbf{(1)}      & 5/22            & 3/16            & 5/6             & 2/5             & 11/28           \\ \cline{2-7} 
                                       & \textbf{(2)}      & 930/7,853       & 638/6,218       & 5,239/3,522     & 4,119/6,116     & 4,533/3,185     \\ \cline{2-7} 
                                       & \textbf{(3)}      & 10.59           & 9.31            & 59.80           & 40.24           & 58.74           \\ \hline
\multirow{3}{*}{\textbf{kddcup}}       & \textbf{(1)}      & 8/18            & 46/22           & 11/32           & 4/0             & 5/2             \\ \cline{2-7} 
                                       & \textbf{(2)}      & 4,294/2,870     & 3,778/3,739     & 3,927/6,250     & 7,309/0         & 71/7,219        \\ \cline{2-7} 
                                       & \textbf{(3)}      & 59.94           & 50.26           & 38.58           & 100.00          & 0.97            \\ \hline
\multirow{3}{*}{\textbf{krvskp}}       & \textbf{(1)}      & 10/46           & 31/142          & 22/117          & 20/74           & 28/99           \\ \cline{2-7} 
                                       & \textbf{(2)}      & 3,614/1,223     & 4,292/1,506     & 8,788/766       & 5,609/224       & 3,792/1,243     \\ \cline{2-7} 
                                       & \textbf{(3)}      & 74.73           & 74.03           & 91.99           & 96.16           & 75.32           \\ \hline
\multirow{3}{*}{\textbf{madelon}}      & \textbf{(1)}      & 11/47           & 12/38           & 6/4             & 11/33           & 20/65           \\ \cline{2-7} 
                                       & \textbf{(2)}      & 6,474/3,034     & 7,170/3,257     & 5,108/514       & 3,705/2,469     & 6,022/3,151     \\ \cline{2-7} 
                                       & \textbf{(3)}      & 68.09           & 68.77           & 90.86           & 60.01           & 65.65           \\ \hline
\multirow{3}{*}{\textbf{mnist}}        & \textbf{(1)}      & 2/11            & 5/5             & 11/0            & 0/13            & 0/2             \\ \cline{2-7} 
                                       & \textbf{(2)}      & 29/7,171        & 10,641/3,660    & 5,780/0         & 0/7,228         & 0/7,214         \\ \cline{2-7} 
                                       & \textbf{(3)}      & 0.40            & 74.41           & 100.00          & 0.00            & 0.00            \\ \hline
\multirow{3}{*}{\textbf{secom}}        & \textbf{(1)}      & 19/89           & 2/14            & 17/97           & 17/118          & 3/31            \\ \cline{2-7} 
                                       & \textbf{(2)}      & 3,974/4,236     & 2,534/5,676     & 6,382/2,404     & 4,405/5,523     & 4/9,414         \\ \cline{2-7} 
                                       & \textbf{(3)}      & 48.40           & 30.86           & 72.64           & 44.37           & 0.04            \\ \hline
\multirow{3}{*}{\textbf{semeion}}      & \textbf{(1)}      & 27/84           & 5/0             & 22/23           & 20/51           & 14/23           \\ \cline{2-7} 
                                       & \textbf{(2)}      & 1,935/6,224     & 9,103/0         & 2,759/3,459     & 3,165/3,647     & 3,311/1,731     \\ \cline{2-7} 
                                       & \textbf{(3)}      & 23.71           & 100.00          & 44.37           & 46.46           & 65.67           \\ \hline
\multirow{3}{*}{\textbf{shuttle}}      & \textbf{(1)}      & 16/57           & 2/0             & 11/0            & 10/49           & 1/10            \\ \cline{2-7} 
                                       & \textbf{(2)}      & 3,727/1,620     & 4,873/0         & 7210/0          & 7,220/2,276     & 1/7,915         \\ \cline{2-7} 
                                       & \textbf{(3)}      & 69.71           & 100.00          & 100.00          & 76.03           & 0.01            \\ \hline
\multirow{3}{*}{\textbf{waveform}}     & \textbf{(1)}      & 18/42           & 5/5             & 11/0            & 25/84           & 20/33           \\ \cline{2-7} 
                                       & \textbf{(2)}      & 4,639/173       & 4,158/410       & 5,511/0         & 4,496/1,058     & 5,146/2,714     \\ \cline{2-7} 
                                       & \textbf{(3)}      & 96.41           & 91.02           & 100.00          & 80.95           & 65.47           \\ \hline
\multirow{3}{*}{\textbf{wineqw}}       & \textbf{(1)}      & 27/90           & 2/0             & 12/0            & 15/62           & 20.32           \\ \cline{2-7} 
                                       & \textbf{(2)}      & 6,082/725       & 4,444/0         & 8,376/0         & 4,573/306       & 8,526/1,412     \\ \cline{2-7} 
                                       & \textbf{(3)}      & 89.35           & 100.00          & 100.00          & 98.73           & 85.79           \\ \hline
\multirow{3}{*}{\textbf{yeast}}        & \textbf{(1)}      & 39/178          & 5/0             & 49/83           & 49/173          & 28/77           \\ \cline{2-7} 
                                       & \textbf{(2)}      & 5,828/325       & 7,033/0         & 8,821/224       & 10,630/1,172    & 7,501/660       \\ \cline{2-7} 
                                       & \textbf{(3)}      & 94.73           & 100.00          & 97.52           & 86.13           & 91.92           \\ \hline
\end{tabular}

\end{table}

\begin{table}[htb!]

\centering


\caption {Negative impacts of invalid pipelines using Auto-sklearn. (1): the number of invalid/ valid pipelines, (2): the total evaluation time of invalid/ valid pipelines (s), (3): the wasted evaluation time (\%).}
\label{tab:neg_impact_invalid_pipeline_autosklearn}
\scriptsize

\begin{tabular}{|l|l|l|l|l|l|l|}
\hline
\textbf{Dataset}                       & \textbf{Criteria} & \textbf{Iter 1} & \textbf{Iter 2} & \textbf{Iter 3} & \textbf{Iter 4} & \textbf{Iter 5} \\ \hline
\textbf{abalone}                       & \textbf{}         & crashed         & crashed         & crashed         & crashed         & crashed         \\ \hline
\textbf{adult}                         &                   & crashed         & crashed         & crashed         & crashed         & crashed         \\ \hline
\textbf{amazon}                        & \textbf{}         & crashed         & crashed         & crashed         & crashed         & crashed         \\ \hline
\textbf{car}                           & \textbf{}         & crashed         & crashed         & crashed         & crashed         & crashed         \\ \hline
\multirow{3}{*}{\textbf{cifar10small}} & \textbf{(1)}      & 4/4             & 7/1             & 6/4             & 6/7             & 6/2             \\ \cline{2-7} 
                                       & \textbf{(2)}      & 4579/2457       & 6819/263        & 5885/1194       & 3867/3253       & 6613/508        \\ \cline{2-7} 
                                       & \textbf{(3)}      & 65.08           & 96.29           & 83.14           & 54.31           & 92.86           \\ \hline
\multirow{3}{*}{\textbf{convex}}       & \textbf{(1)}      & 9/19            & 9/18            & 9/16            & 8/18            & 9/14            \\ \cline{2-7} 
                                       & \textbf{(2)}      & 2692/4475       & 2625/4539       & 2880/4291       & 2646/4524       & 3121/4052       \\ \cline{2-7} 
                                       & \textbf{(3)}      & 37.57           & 36.65           & 40.16           & 36.90           & 43.51           \\ \hline
\multirow{3}{*}{\textbf{dexter}}       & \textbf{(1)}      & 9/44            & 9/42            & 9/44            & 9/44            & 9/44            \\ \cline{2-7} 
                                       & \textbf{(2)}      & 4917/2177       & 3773/3333       & 4742/2366       & 4704/2415       & 4028/3079       \\ \cline{2-7} 
                                       & \textbf{(3)}      & 69.31           & 53.10           & 66.72           & 66.08           & 56.68           \\ \hline
\textbf{dorothea}                      & \textbf{}         & crashed         & crashed         & crashed         & crashed         & crashed         \\ \hline
\textbf{gcredit}                       & \textbf{}         & crashed         & crashed         & crashed         & crashed         & crashed         \\ \hline
\multirow{3}{*}{\textbf{gisette}}      & \textbf{(1)}      & 6/7             & 5/6             & 4/3             & 5/4             & 3/5             \\ \cline{2-7} 
                                       & \textbf{(2)}      & 5227/1920       & 4380/2723       & 6734/387        & 6125/1002       & 5156/1988       \\ \cline{2-7} 
                                       & \textbf{(3)}      & 73.13           & 61.67           & 94.56           & 85.94           & 72.17           \\ \hline
\textbf{kddcup}                        & \textbf{}         & crashed         & crashed         & crashed         & crashed         & crashed         \\ \hline
\textbf{krvskp}                        & \textbf{}         & crashed         & crashed         & crashed         & crashed         & crashed         \\ \hline
\multirow{3}{*}{\textbf{madelon}}      & \textbf{(1)}      & 9/50            & 11/47           & 9/49            & 9/49            & 9/49            \\ \cline{2-7} 
                                       & \textbf{(2)}      & 4215/2925       & 4763/2374       & 4914/2225       & 4774/2366       & 3718/3423       \\ \cline{2-7} 
                                       & \textbf{(3)}      & 59.03           & 66.74           & 68.84           & 66.86           & 52.07           \\ \hline
\multirow{3}{*}{\textbf{mnist}}        & \textbf{(1)}      & 3/8             & 6/10            & 6/10            & 7/5             & 5/9             \\ \cline{2-7} 
                                       & \textbf{(2)}      & 2802/4355       & 3379/3795       & 4013/3160       & 5837/1337       & 5516/1658       \\ \cline{2-7} 
                                       & \textbf{(3)}      & 39.15           & 47.10           & 55.94           & 81.37           & 76.89           \\ \hline
\textbf{secom}                         & \textbf{}         & crashed         & crashed         & crashed         & crashed         & crashed         \\ \hline
\multirow{3}{*}{\textbf{semeion}}      & \textbf{(1)}      & 4/32            & 12/55           & 7/51            & 5/37            & 4/26            \\ \cline{2-7} 
                                       & \textbf{(2)}      & 6122/1042       & 5906/1228       & 5412/1731       & 5715/1444       & 5943/1230       \\ \cline{2-7} 
                                       & \textbf{(3)}      & 85.46           & 82.79           & 75.77           & 79.83           & 82.85           \\ \hline
\multirow{3}{*}{\textbf{shuttle}}      & \textbf{(1)}      & 3/10            & 3/10            & 2/18            & 2/15            & 3/19            \\ \cline{2-7} 
                                       & \textbf{(2)}      & 4329/2853       & 4312/2870       & 2087/5087       & 1897/5282       & 1957/5215       \\ \cline{2-7} 
                                       & \textbf{(3)}      & 60.27           & 60.03           & 29.09           & 26.42           & 27.28           \\ \hline
\multirow{3}{*}{\textbf{waveform}}     & \textbf{(1)}      & 3/21            & 4/27            & 3/27            & 3/30            & 3/22            \\ \cline{2-7} 
                                       & \textbf{(2)}      & 4081/3098       & 4295/2873       & 3679/3489       & 3890/3277       & 3876/3299       \\ \cline{2-7} 
                                       & \textbf{(3)}      & 56.84           & 59.92           & 51.32           & 54.28           & 54.02           \\ \hline
\multirow{3}{*}{\textbf{wineqw}}       & \textbf{(1)}      & 3/54            & 2/39            & 3/55            & 2/57            & 2/43            \\ \cline{2-7} 
                                       & \textbf{(2)}      & 1891/5244       & 2845/4309       & 1815/5322       & 9/7127          & 2140/5012       \\ \cline{2-7} 
                                       & \textbf{(3)}      & 26.50           & 39.77           & 25.44           & 0.12            & 29.92           \\ \hline
\textbf{yeast}                         & \textbf{}         & crashed         & crashed         & crashed         & crashed         & crashed         \\ \hline
\end{tabular}

\vspace{-0.35cm}

\end{table}

\begin{landscape}

\begin{table*}[htb!]
\centering

\scriptsize
\begin{tabular}{|l|l|l|l|l|l|l|l|l|l|l|l|l|l|l|l|}
\hline
\multirow{2}{*}{\textbf{Dataset}} & \multicolumn{3}{l|}{\textbf{Number of invalid pipelines}} & \multicolumn{3}{l|}{\textbf{Number of valid pipelines}} & \multicolumn{3}{c|}{\textbf{\begin{tabular}[c]{@{}c@{}}Total evaluation time of\\ invalid pipelines (s)\end{tabular}}} & \multicolumn{3}{c|}{\textbf{\begin{tabular}[c]{@{}c@{}}Total evaluation time of\\ valid pipelines (s)\end{tabular}}} & \multicolumn{3}{l|}{\textbf{Wasted evaluation time (\%)}} \\ \cline{2-16} 
                                  & \textbf{Mean}     & \textbf{StdDev}     & \textbf{Max}    & \textbf{Mean}    & \textbf{StdDev}    & \textbf{Max}    & \textbf{Mean}                         & \textbf{StdDev}                         & \textbf{Max}                         & \textbf{Mean}                         & \textbf{StdDev}                        & \textbf{Max}                        & \textbf{Mean}     & \textbf{StdDev}     & \textbf{Max}    \\ \hline
\textbf{abalone}                  & 16                & 10                  & 29              & 57               & 27                 & 97              & 5,531                                 & 3,328                                   & 8,639                                & 2,063                                 & 2,738                                  & 7,284                               & 70.18             & 38.04               & 99.10           \\ \hline
\textbf{adult}                    & 9                 & 7                   & 22              & 23               & 2                  & 25              & 4,093                                 & 2,418                                   & 7,321                                & 3,977                                 & 3,036                                  & 8,188                               & 53.55             & 34.75               & 95.36           \\ \hline
\textbf{amazon}                   & 12                & 9                   & 26              & 9                & 10                 & 23              & 5,861                                 & 1,600                                   & 7,701                                & 962                                   & 1,230                                  & 3,056                               & 89.12             & 14.38               & 100.00          \\ \hline
\textbf{car}                      & 36                & 18                  & 71              & 148              & 59                 & 263             & 6,527                                 & 3,754                                   & 13,092                               & 2,347                                 & 1,735                                  & 5,039                               & 70.31             & 22.36               & 97.35           \\ \hline
\textbf{cifar10small}             & 8                 & 6                   & 16               & 12               & 11                 & 30              & 3,300                                 & 2,676                                   & 6,590                                & 4,544                                 & 3,747                                  & 9,700                               & 46.81             & 39.63               & 100.00          \\ \hline
\textbf{convex}                   & 5                 & 3                   & 10              & 19               & 13                 & 40              & 4,448                                 & 1,540                                   & 6,622                                & 3,711                                 & 956                                    & 4,968                               & 53.75             & 13.62               & 66.31           \\ \hline
\textbf{dexter}                   & 5                 & 5                   & 14              & 18               & 20                 & 55              & 3,969                                 & 2,879                                   & 7,210                                & 3,397                                 & 3,242                                  & 8,796                               & 55.44             & 39.07               & 100.00          \\ \hline
\textbf{dorothea}                 & 5                 & 3                   & 9               & 7                & 8                  & 17              & 3,661                                 & 2,207                                   & 7,208                                & 2,450                                 & 2,728                                  & 7,261                               & 65.01             & 36.11               & 100.00          \\ \hline
\textbf{gcredit}                  & 75                & 38                  & 139             & 354              & 191                & 706             & 2,458                                 & 1,672                                   & 5,233                                & 3,400                                 & 1,968                                  & 5,908                               & 43.09             & 28.11               & 84.77           \\ \hline
\textbf{gisette}                  & 5                 & 3                   & 11              & 15               & 9                  & 28              & 3,092                                 & 1,920                                   & 5,239                                & 5,379                                 & 1,768                                  & 7,853                               & 35.74             & 22.18               & 59.80           \\ \hline
\textbf{kddcup}                   & 15                & 16                  & 46              & 15               & 12                 & 32              & 3,876                                 & 2,300                                   & 7,309                                & 4,016                                 & 2,560                                  & 7,219                               & 49.95             & 32.04               & 100.00          \\ \hline
\textbf{krvskp}                   & 22                & 7                   & 31              & 96               & 33                 & 142             & 5,219                                 & 1,916                                   & 8,788                                & 992                                   & 452                                    & 1,506                               & 82.45             & 9.59                & 96.16           \\ \hline
\textbf{madelon}                  & 12                & 5                   & 20              & 37               & 20                 & 65              & 5,696                                 & 1,200                                   & 7,170                                & 2,485                                 & 1,022                                  & 3,257                               & 70.68             & 10.55               & 90.86           \\ \hline
\textbf{mnist}                    & 4                 & 4                   & 11              & 6                & 5                  & 13              & 3,290                                 & 4,302                                   & 10,641                               & 5,055                                 & 2,876                                  & 7,228                               & 34.96             & 43.42               & 100.00          \\ \hline
\textbf{secom}                    & 12                & 7                   & 19              & 70               & 40                 & 118             & 3,460                                 & 2,121                                   & 6,382                                & 5,451                                 & 2,303                                  & 9,414                               & 39.26             & 23.80               & 72.64           \\ \hline
\textbf{semeion}                  & 18                & 8                   & 27              & 36               & 29                 & 84              & 4,055                                 & 2,569                                   & 9,103                                & 3,012                                 & 2,080                                  & 6,224                               & 56.04             & 25.68               & 100.00          \\ \hline
\textbf{shuttle}                  & 8                 & 6                   & 16              & 23               & 25                 & 57              & 4,606                                 & 2,671                                   & 7,220                                & 2,362                                 & 2,917                                  & 7,915                               & 69.15             & 36.69               & 100.00          \\ \hline
\textbf{waveform}                 & 16                & 7                   & 25              & 33               & 30                 & 84              & 4,790                                 & 481                                     & 5,511                                & 871                                   & 989                                    & 2,714                               & 86.77             & 12.44               & 100.00          \\ \hline
\textbf{wineqw}                   & 15                & 8                   & 27              & 38               & 39                 & 90              & 6,400                                 & 1,771                                   & 8,526                                & 489                                   & 533                                    & 1,412                               & 94.77             & 6.01                & 100.00          \\ \hline
\textbf{yeast}                    & 34                & 16                  & 49              & 102              & 67                 & 178             & 7,963                                 & 1,642                                   & 10,630                               & 476                                   & 408                                    & 1,172                               & 94.06             & 4.80                & 100.00          \\ \hline
\end{tabular}
\caption {Mean, standard deviation (StdDev), and Maximum (Max) of the number of invalid and valid pipelines, the total evaluation time of invalid and valid pipelines, and the wasted evaluation time using AutoWeka4MCPS.}
\label{tab:exp1_mean_aw}
\vspace{-0.5cm}
\end{table*}

\end{landscape}
\begin{landscape}

\begin{table*}[htb!]
\centering

\scriptsize
\begin{tabular}{|l|l|l|l|l|l|l|l|l|l|l|l|l|l|l|l|}
\hline
\multirow{2}{*}{\textbf{Dataset}} & \multicolumn{3}{l|}{\textbf{Number of invalid pipelines}} & \multicolumn{3}{l|}{\textbf{Number of valid pipelines}} & \multicolumn{3}{c|}{\textbf{\begin{tabular}[c]{@{}c@{}}Total evaluation time of\\ invalid pipelines (s)\end{tabular}}} & \multicolumn{3}{c|}{\textbf{\begin{tabular}[c]{@{}c@{}}Total evaluation time of\\ valid pipelines (s)\end{tabular}}} & \multicolumn{3}{l|}{\textbf{Wasted evaluation time (\%)}} \\ \cline{2-16} 
                                  & \textbf{Mean}     & \textbf{StdDev}     & \textbf{Max}    & \textbf{Mean}    & \textbf{StdDev}    & \textbf{Max}    & \textbf{Mean}                         & \textbf{StdDev}                         & \textbf{Max}                         & \textbf{Mean}                         & \textbf{StdDev}                        & \textbf{Max}                        & \textbf{Mean}     & \textbf{StdDev}     & \textbf{Max}    \\ \hline
\textbf{abalone}                  & -                 & -                   & -               & -                & -                  & -               & -                                     & -                                       & -                                    & -                                     & -                                      & -                                   & -                 & -                   & -               \\ \hline
\textbf{adult}                    & -                 & -                   & -               & -                & -                  & -               & -                                     & -                                       & -                                    & -                                     & -                                      & -                                   & -                 & -                   & -               \\ \hline
\textbf{amazon}                   & -                 & -                   & -               & -                & -                  & -               & -                                     & -                                       & -                                    & -                                     & -                                      & -                                   & -                 & -                   & -               \\ \hline
\textbf{car}                      & -                 & -                   & -               & -                & -                  & -               & -                                     & -                                       & -                                    & -                                     & -                                      & -                                   & -                 & -                   & -               \\ \hline
\textbf{cifar10small}             & 6                 & 1                   & 7               & 4                & 2                  & 7               & 5,553                                 & 1,151                                   & 6,819                                & 1,535                                 & 1,148                                  & 3,253                               & 78.34             & 16.18               & 96.29           \\ \hline
\textbf{convex}                   & 9                 & 0                   & 9               & 17               & 2                  & 19              & 2,793                                 & 187                                     & 3,121                                & 4,376                                 & 185                                    & 4,539                               & 38.96             & 2.59                & 43.51           \\ \hline
\textbf{dexter}                   & 9                 & 0                   & 9               & 44               & 1                  & 44              & 4,433                                 & 448                                     & 4,917                                & 2,674                                 & 449                                    & 3,333                               & 62.38             & 6.31                & 69.31           \\ \hline
\textbf{dorothea}                 & -                 & -                   & -               & -                & -                  & -               & -                                     & -                                       & -                                    & -                                     & -                                      & -                                   & -                 & -                   & -               \\ \hline
\textbf{gcredit}                  & -                 & -                   & -               & -                & -                  & -               & -                                     & -                                       & -                                    & -                                     & -                                      & -                                   & -                 & -                   & -               \\ \hline
\textbf{gisette}                  & 5                 & 1                   & 6               & 5                & 1                  & 7               & 5,524                                 & 819                                     & 6,734                                & 1,604                                 & 818                                    & 2,723                               & 77.49             & 11.49               & 94.56           \\ \hline
\textbf{kddcup}                   & -                 & -                   & -               & -                & -                  & -               & -                                     & -                                       & -                                    & -                                     & -                                      & -                                   & -                 & -                   & -               \\ \hline
\textbf{krvskp}                   & -                 & -                   & -               & -                & -                  & -               & -                                     & -                                       & -                                    & -                                     & -                                      & -                                   & -                 & -                   & -               \\ \hline
\textbf{madelon}                  & 9                 & 1                   & 11              & 49               & 1                  & 50              & 4,477                                 & 448                                     & 4,914                                & 2,663                                 & 449                                    & 3,423                               & 62.71             & 6.29                & 68.84           \\ \hline
\textbf{mnist}                    & 5                 & 1                   & 7               & 8                & 2                  & 10              & 4,309                                 & 1,184                                   & 5,837                                & 2,861                                 & 1,180                                  & 4,355                               & 60.09             & 16.49               & 81.37           \\ \hline
\textbf{secom}                    & -                 & -                   & -               & -                & -                  & -               & -                                     & -                                       & -                                    & -                                     & -                                      & -                                   & -                 & -                   & -               \\ \hline
\textbf{semeion}                  & 6                 & 3                   & 12              & 40               & 11                 & 55              & 5,820                                 & 241                                     & 6,122                                & 1,335                                 & 235                                    & 1,731                               & 81.34             & 3.31                & 85.46           \\ \hline
\textbf{shuttle}                  & 3                 & 0                   & 3               & 14               & 4                  & 19              & 2,916                                 & 1,148                                   & 4,329                                & 4,261                                 & 1,145                                  & 5,282                               & 40.62             & 15.97               & 60.27           \\ \hline
\textbf{waveform}                 & 3                 & 0                   & 4               & 25               & 3                  & 30              & 3,964                                 & 209                                     & 4,295                                & 3,207                                 & 208                                    & 3,489                               & 55.28             & 2.91                & 59.92           \\ \hline
\textbf{wineqw}                   & 2                 & 0                   & 3               & 50               & 7                  & 57              & 1,740                                 & 939                                     & 2,845                                & 5,403                                 & 933                                    & 7,127                               & 24.35             & 13.13               & 39.77           \\ \hline
\textbf{yeast}                    & -                 & -                   & -               & -                & -                  & -               & -                                     & -                                       & -                                    & -                                     & -                                      & -                                   & -                 & -                   & -               \\ \hline
\end{tabular}

\caption {Mean, standard deviation (StdDev), and Maximum (Max) of the number of invalid and valid pipelines, the total evaluation time of invalid and valid pipelines, and the wasted evaluation time using Auto-sklearn. Symbol '-' presents crashed experiments.}
\label{tab:exp1_mean_ak}
\vspace{-0.5cm}
\end{table*}

\end{landscape}


Table \ref{tab:neg_impact_invalid_pipeline_aw} and  \ref{tab:neg_impact_invalid_pipeline_autosklearn} present the number of invalid pipelines and the wasted time used to evaluate these invalid pipelines in ML pipeline composition and optimisation of AutoWeka4MCPS and Auto-sklearn. Table \ref{tab:exp1_mean_aw} and Table \ref{tab:exp1_mean_ak} present mean, standard deviation and maximum of the number of invalid and valid pipelines, the evaluation time of the invalid and valid pipelines, and the wasted evaluation time.
Firstly, these tables show that not all of the constructed pipelines are valid. We can see the presence of invalid pipelines for both AutoML tools across different datasets and iterations. For example, there are 9 invalid and 54 valid pipelines in the case of using AutoWeka4MCPS for the dataset \textit{abalone} in \textit{Iter 1}. Mean, standard deviation and maximum of the number of invalid pipelines in the case of using AutoWeka4MCPS for the dataset \textit{abalone} are 16, 10 and 29 respectively.
There are 9 invalid and 19 valid pipelines in the case of using Auto-sklearn, the dataset \textit{convex} and \textit{Iter 1}. 
Mean, standard deviation and maximum of the number of invalid pipelines in the case of using Auto-sklearn for the dataset \textit{convex} is 9, 0 and 9 respectively.
Secondly, the evaluation time of these invalid pipelines may be significant in several cases. For example, the wasted evaluation time is 75.48\% in the case of using the dataset \textit{car} and \textit{Iter 5}. We can see that changing seed numbers has a strong impact on the wasted evaluation time in the case of AutoWeka4MCPS. For example, the experiments with the dataset \textit{abalone} show that the wasted evaluation time is in the range between 0.21\% and 99.10\%. The reason is that Weka libraries themselves can evaluate the compatibility of a single component pipeline without execution. If the initialisation of the pipeline composition and optimisation with a specific seed number results in pipelines consisting of only one predictor, and these pipelines are well-performing, it tends to exploit similar ML pipelines. As a result, the wasted evaluation time is low. However, if the initialisation results in a complex pipeline which is invalid, there is no guidance for the next promising pipelines. Therefore, the next promising pipelines are randomly selected. As a result, the upcoming pipelines that will be evaluated may also be invalid. 
This impact is negligible in the case of Auto-sklearn. In other words, the impact of changing seed numbers on the variance of the wasted evaluation in the case of Auto-sklearn is less than in the case of AutoWeka4MCPS. For example, standard deviation of the number of invalid pipelines in the case of using Auto-sklearn and the dataset \textit{cifar10small, convex and dexter} are 1, 0 and 0 respectively. However, standard deviation of the number of invalid pipelines in the case of using AutoWeka4MCPS and the dataset \textit{cifar10small, convex and dexter} are 6, 3, 5 respectively.
The reason is that Auto-sklearn uses meta-learning to initialise with promising ML pipelines.
The experiments with the datasets \textit{abalone}, \textit{adult}, \textit{amazon}, \textit{car}, \textit{dorothea}, \textit{gcredit}, \textit{kddcup}, \textit{krvskp}, \textit{secom},  and \textit{yeast} show that Auto-sklearn limits the generation of invalid pipelines by making assumption about cleaned input datasets. The experiments crash if the input datasets have data quality issues (i.e. missing values) or not transformed into a specific, required format (i.e., all attributes must be in numeric format). Similar to AutoWeka4MCPS, the Auto-sklearn can not handle invalid pipelines effectively even with the initialisation using the meta-learning. In conclusion, we empirically prove the presence of invalid pipelines and that the wasted time used to evaluate these invalid pipelines can be significant.

\subsection{Experiments to compare the performance of the AVATAR and the traditional method that requires executions}
\subsubsection{Experiment settings}

In order to demonstrate the efficiency of the AVATAR, we compare the performance of the AVATAR and the traditional method (which we refer to as T-method) that requires the executions of pipelines. The T-method is used to evaluate the validity of pipelines in the pipeline composition and optimisation of AutoWeka4MCPS and Auto-sklearn. We use the same datasets as in the first set of experiments. The goal of the second set of experiments is to show that the AVATAR can evaluate the validity of pipelines faster than the T-method and the evaluated results of the AVATAR are similar to the T-method. 

We randomly generate ML pipelines which have up to six components (i.e. these component types are missing value handling, dimensionality reduction, outlier removal, data transformation, data sampling and predictor). The predictor is put at the end of the pipelines because a valid pipeline always has a predictor at the end.
Each pipeline is evaluated using both the AVATAR and the T-method. We set the time budget to 10 hours per dataset instead of 2 hours in the first set of experiments.
The reason is that we want to use the same total time budget of the first set of experiments. Each iteration of the first set of experiments has 2 hours optimisation time for each dataset. Therefore, five iterations have 10 hours of optimisation time.
Please note that the generation of random pipelines in this set of experiments is not a part of the full ML pipeline composition and optimisation process because each random pipeline is evaluated by both the AVATAR and the T-method. We use the following criteria to compare the performance: (1) the number of invalid/ valid pipelines, (2) the total evaluation time of invalid/ valid pipelines (seconds), (3) the number of pipelines that have the same evaluated results between the AVATAR and the T-method, and (4) the percentage of the pipelines that the AVATAR can validate accurately (\%) in comparison to the T-method. 

\subsubsection{Experiment results}

\begin{table}[htb!]

\vspace{-0.35cm}
\caption {Comparison of the performance of the AVATAR and T-method}
\label{tab:exp_avatar_tmethod}

\centering
\tiny

\begin{tabular}{|l|l|l|l|l|l|l|}
\hline
\multirow{2}{*}{\textbf{Dataset}} & \multicolumn{2}{l|}{\textbf{T-method}}                                                                                                                                                  & \multicolumn{2}{l|}{\textbf{AVATAR}}                                                                                                                                                    & \multirow{2}{*}{\textbf{\begin{tabular}[c]{@{}l@{}}Pipelines have\\ different/similar\\ evaluated results\end{tabular}}} & \multirow{2}{*}{\textbf{\begin{tabular}[c]{@{}l@{}}The percentage of\\ pipelines that the\\ AVATAR can evaluate\\ accurately (\%)\end{tabular}}} \\ \cline{2-5}
                                  & \textbf{\begin{tabular}[c]{@{}l@{}}Invalid/valid\\ pipelines\end{tabular}} & \textbf{\begin{tabular}[c]{@{}l@{}}Total evaluation\\ time of invalid/\\ valid pipelines (s)\end{tabular}} & \textbf{\begin{tabular}[c]{@{}l@{}}Invalid/valid\\ pipelines\end{tabular}} & \textbf{\begin{tabular}[c]{@{}l@{}}Total evaluation\\ time of invalid/\\ valid pipelines (s)\end{tabular}} &                                                                                                                          &                                                                                                                                                  \\ \hline
\textbf{abalone}                  & 553/780                                                                    & 18,609/17,336                                                                                              & 537/796                                                                    & 23/27                                                                                                      & 16/1,317                                                                                                                 & 98.80                                                                                                                                            \\ \hline
\textbf{adult}                    & 441/663                                                                    & 6,041/29,823                                                                                               & 430/674                                                                    & 28/169                                                                                                     & 11/1,093                                                                                                                 & 99.00                                                                                                                                            \\ \hline
\textbf{amazon}                   & 40/60                                                                      & 552/38,053                                                                                                 & 38/62                                                                      & 18/94                                                                                                      & 2/98                                                                                                                     & 98.00                                                                                                                                            \\ \hline
\textbf{car}                      & 3,159/4,858                                                                & 13,561/22,277                                                                                              & 3,156/4,861                                                                & 65/92                                                                                                      & 3/8,014                                                                                                                  & 99.96                                                                                                                                            \\ \hline
\textbf{cifar10small}             & 20/31                                                                      & 817/34,939                                                                                                 & 19/32                                                                      & 96/205                                                                                                     & 1/50                                                                                                                     & 98.04                                                                                                                                            \\ \hline
\textbf{convex}                   & 159/248                                                                    & 9,655/26,170                                                                                               & 154/253                                                                    & 33/219                                                                                                     & 5/402                                                                                                                    & 98.70                                                                                                                                            \\ \hline
\textbf{dexter}                   & 274/412                                                                    & 2,137/33,528                                                                                               & 272/414                                                                    & 54/290                                                                                                     & 2/684                                                                                                                    & 99.71                                                                                                                                            \\ \hline
\textbf{dorothea}                 & 95/100                                                                     & 4,677/30,354                                                                                               & 83/112                                                                     & 293/1,423                                                                                                  & 12/183                                                                                                                   & 93.85                                                                                                                                            \\ \hline
\textbf{gcredit}                  & 3,616/5,373                                                                & 16,033/19,755                                                                                              & 3,613/5,376                                                                & 82/125                                                                                                     & 3/8,986                                                                                                                  & 99.97                                                                                                                                            \\ \hline
\textbf{gisette}                  & 83/122                                                                     & 2,418/33,179                                                                                               & 79/126                                                                     & 134/386                                                                                                    & 4/201                                                                                                                    & 98.05                                                                                                                                            \\ \hline
\textbf{kddcup}                   & 38/30                                                                      & 3,238/33,326                                                                                               & 34/34                                                                      & 236/235                                                                                                    & 4/64                                                                                                                     & 94.12                                                                                                                                            \\ \hline
\textbf{krvskp}                   & 3,106/4,858                                                                & 14,084/21,703                                                                                              & 3,096/4,868                                                                & 74/132                                                                                                     & 10/7,954                                                                                                                 & 99.87                                                                                                                                            \\ \hline
\textbf{madelon}                  & 1,032/1,633                                                                & 5,846/29,893                                                                                               & 1,028/1,637                                                                & 53/209                                                                                                     & 4/2,661                                                                                                                  & 99.85                                                                                                                                            \\ \hline
\textbf{mnist}                    & 70/135                                                                     & 1,014/36,940                                                                                               & 70/135                                                                     & 48/195                                                                                                     & 0/205                                                                                                                    & 100.00                                                                                                                                           \\ \hline
\textbf{secom}                    & 860/1,337                                                                  & 6,599/32,203                                                                                               & 852/1,345                                                                  & 59/203                                                                                                     & 8/2,189                                                                                                                  & 99.64                                                                                                                                            \\ \hline
\textbf{semeion}                  & 1,141/1,783                                                                & 5,755/30,123                                                                                               & 1,141/1,783                                                                & 35/86                                                                                                      & 0/2,924                                                                                                                  & 100.00                                                                                                                                           \\ \hline
\textbf{shuttle}                  & 553/829                                                                    & 9,237/27,299                                                                                               & 540/842                                                                    & 19/64                                                                                                      & 13/1,369                                                                                                                 & 99.06                                                                                                                                            \\ \hline
\textbf{waveform}                 & 2,096/3,201                                                                & 12,185/23,604                                                                                              & 2,031/3,266                                                                & 62/147                                                                                                     & 65/5,232                                                                                                                 & 98.77                                                                                                                                            \\ \hline
\textbf{wineqw}                   & 837/1,319                                                                  & 14,246/22,188                                                                                              & 823/1,333                                                                  & 24/37                                                                                                      & 14/2,142                                                                                                                 & 99.35                                                                                                                                            \\ \hline
\textbf{yeast}                    & 2,868/4,527                                                                & 13,672/22,177                                                                                              & 2,849/4,546                                                                & 60/87                                                                                                      & 19/7,376                                                                                                                 & 99.74                                                                                                                                            \\ \hline
\end{tabular}

\vspace{-0.35cm}

\end{table}

Table \ref{tab:exp_avatar_tmethod} compares the performance of the AVATAR and the T-method using the above criteria. We can see that the total evaluation time of invalid/valid pipelines of the AVATAR is significantly lower than the T-method. 
For example, in the case of using the dataset \textit{convex}, the number of invalid pipelines of the AVATAR and the T-method are approximately the same, 154 in comparison with 159. However, the T-method requires more time than the AVATAR to evaluate these invalid pipelines significantly, 9,655 seconds in comparison with 33 seconds. The number of valid pipelines of the AVATAR and the T-method are also approximately the same, 248 in comparison with 253. However, the T-method requires more time than the AVATAR to evaluate these valid pipelines significantly, 26,170 seconds in comparison with 219 seconds.
While the evaluation time of pipelines of the AVATAR is quite stable, the evaluation time of pipelines of the T-method is much higher and depends on the size of the datasets. 
For example, the dataset \textit{kddcup} is larger than \textit{abalone} in terms of the number of instances and the number of attributes. The sum of the evaluation time of invalid and valid pipelines in case of using the T-method with the dataset \textit{kddcup} and \textit{abalone} are approximately the same, 36,564 seconds in comparison with 35,945 seconds. However, the sum of the number of invalid and valid pipelines in case of using the T-method with the dataset \textit{kddcup} and \textit{abalone} are 68 and 1,333 respectively.

It means that the AVATAR is faster than the T-method in evaluating both invalid and valid pipelines regardless of the size of datasets.
The reason is that the AVATAR does not execute pipelines with a dataset. The AVATAR maps these pipelines to Petri net based pipelines. After that, the AVATAR finds the validity by firing the transitions of these surrogate pipelines with the input token mapped from the input dataset. These mappings and calculations are less time-consuming than executions of the original pipelines.   

In order to prove that the AVATAR and the T-method deliver similar evaluated results, we compare the evaluated results of pipelines between the AVATAR and the T-method. We calculate the accuracy of the AVATAR by the percentage of the number of pipelines that have similar evaluated results over total number pipelines.  
Table \ref{tab:exp_avatar_tmethod} shows that the percentage
of pipelines that the AVATAR can evaluate in the same way as T-method is more than 98\% on all datasets excluding the cases of using the datasets \textit{dorothea} and \textit{kddcup} for which it is approximately 94\%. We have carefully reviewed the pipelines which have different evaluation results between the AVATAR and the T-method. Interestingly, the AVATAR evaluates all of these pipelines to be valid, and the T-method evaluates these pipelines to be invalid. The reason is that executions of these pipelines cause the out of memory problem. In other words, the AVATAR does not consider the allocated memory as an impact on the validity of a pipeline.
To deal with the pipelines that were classified as invalid because of insufficient memory to validate them, a promising solution is to reduce the size of an input dataset by adding a sampling component with appropriate hyperparameters. If the sampling size is too small, we may miss important features. If the sampling size is too large, we may continue to run into the out of memory problem.  
We cannot conclude that if we allocate more memory, whether the executions of these pipelines would be successful or not. It shows that the validity of a pipeline also depends on its execution environment such as memory. These factors have not been considered yet in the AVATAR. This is an interesting research gap that will be addressed in the future.  

\begin{table}[htb!]

\centering

\vspace{-0.35cm}
\caption {Five invalid pipelines with the longest evaluation time using the T-method on the dataset \textit{mnist}}
\label{tab:top5_worst_pipeline}

\scriptsize

\begin{tabular}{|l|l|l|l|l|l|}
\hline
\textbf{Pipeline}     & \#1    & \#2    & \#3    & \#4    & \#5    \\ \hline
\textbf{T-method (s)} & 39.779 & 39.334 & 30.904 & 30.418 & 29.438 \\ \hline
\textbf{AVATAR (s)}   & 0.375  & 0.016  & 0.062  & 0.063  & 0.032  \\ \hline
\end{tabular}
\vspace{-0.35cm}

\end{table}

Finally, we take a detailed look at the invalid pipelines with the longest evaluation time using the T-method on the dataset \textit{mnist}, as shown in Table \ref{tab:top5_worst_pipeline}. Pipeline \#1 (39.779s) has the structure \textit{Discretize} \(\rightarrow\) \textit{RemovePercentage} \(\rightarrow\) \textit{RemoveUseless} \(\rightarrow\) \textit{RemoveOutliers} \(\rightarrow\)  \textit{EMImputation} \(\rightarrow\)  \textit{LogitBoost}. All of the attributes of the dataset \textit{mnist} are numeric. The component \textit{Discretize} transforms numeric to nominal attributes in the forms of bins. The components \textit{RemovePercentage}, \textit{RemoveUseless}, \textit{RemoveOutliers} are compatible with nominal attributes, and do not have any effect on nominal attributes. The component \textit{EMImputation} is not compatible with nominal attributes in the output dataset of the component \textit{RemoveOutliers} that leads to an invalid pipeline.

This pipeline is invalid because \textit{EMImputation} does not work with nominal attributes, and there is no component transforming the nominal to numeric attributes. We can see that the AVATAR is able to evaluate the validity of this pipeline without executing it in just 0.375 s.


\subsection{Experiments to compare the performance of SMAC with and without the AVATAR}
\subsubsection{Experiment Settings}
\label{sec:exp_settings}

The AutoML tool used for the experiments is AutoWeka4MCPS\footnote{\url{https://github.com/UTS-AAi/autoweka}} with integrated AVATAR.
We use the same experiment settings as for the first set of experiments. We use SMAC as a pipeline composition and optimisation method. We evaluate the experimental results of the five iterations with and without the AVATAR. We compare the performance of SMAC with and without the AVATAR using the following criteria:

\begin{itemize}
    \item Mean, standard deviation and min of the error rate; and 
    \item Mean, standard deviation and max of the number of the successfully evaluated pipelines.
\end{itemize}

\subsubsection{Experiment Results}

Table \ref{tab:exp_awa_2h} presents mean, standard deviation and minimum of the error rate, and mean, standard deviation and maximum of the number of evaluated pipelines of 5 iterations with and without the AVATAR using the datasets presented in Table \ref{tab:exp_datasets}.
We can see that using the AVATAR can accelerate the ML pipeline composition and optimisation. Additionally the results support the following observations: 

\begin{itemize}
    \item SMAC can evaluate more pipelines with the AVATAR than without it. The mean of the number of evaluated pipelines, when AVATAR is used, is higher than in the cases without the AVATAR for all datasets.  
    
    
    \item SMAC finds better pipelines with the AVATAR than without it. Table \ref{tab:exp_awa_2h} shows that mean of the error rate in experiments with the AVATAR is smaller than in the cases without the AVATAR in 15/20 datasets. 

    \item The impact of SMAC's initialisation on the convergence.
    Standard deviation of the error rate indicates the degree of the difference of the iterations' convergences given different initialisation configurations of SMAC. This initialisation randomly selects first configurations based on seed numbers.
    The high value of this statistical metric indicates that the iterations are not able to converge given the time budget.
    Table \ref{tab:exp_awa_2h} shows that the standard deviation of the error rate for two datasets in both cases with and without the AVATAR is greater than 5\%. They are the iterations using the  \textit{amazon} and \textit{cifar10small} datasets. It means that the convergence of the results for these datasets highly depends on the initialisation of SMAC because these datasets are large and the time budget for the optimisation is not enough.  We have investigated the impact the initialisation of SMAC with multiple configurations in the fourth set of experiments.
\end{itemize}

\begin{figure}[h] 
  
    \subfloat{%
        \includegraphics[width=0.3\linewidth]{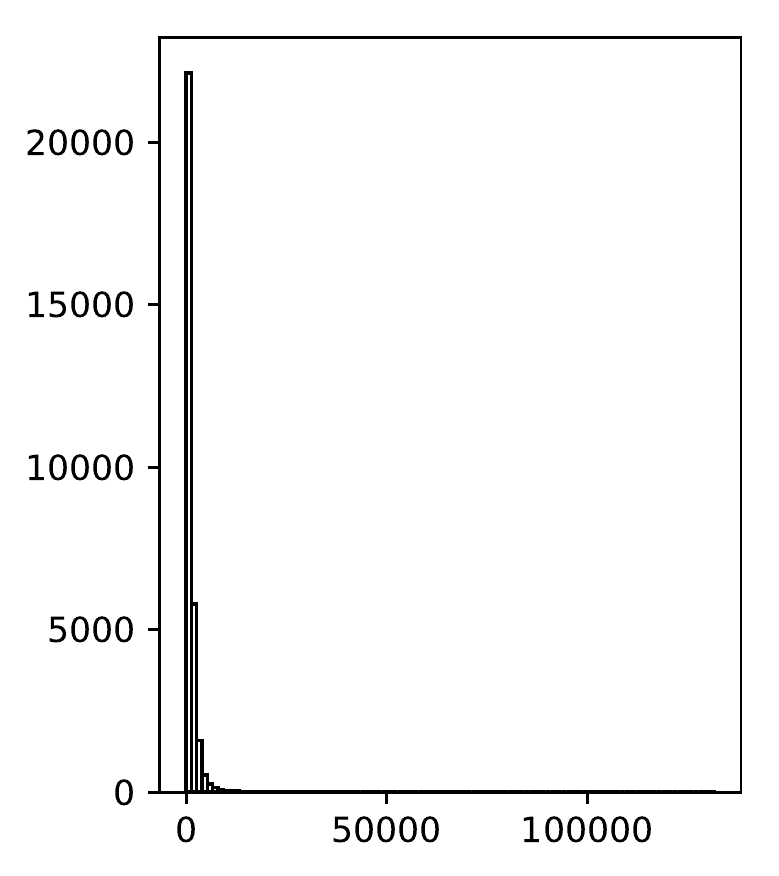}
      }%
    \hfill
    \subfloat{%
        \includegraphics[width=0.315\linewidth]{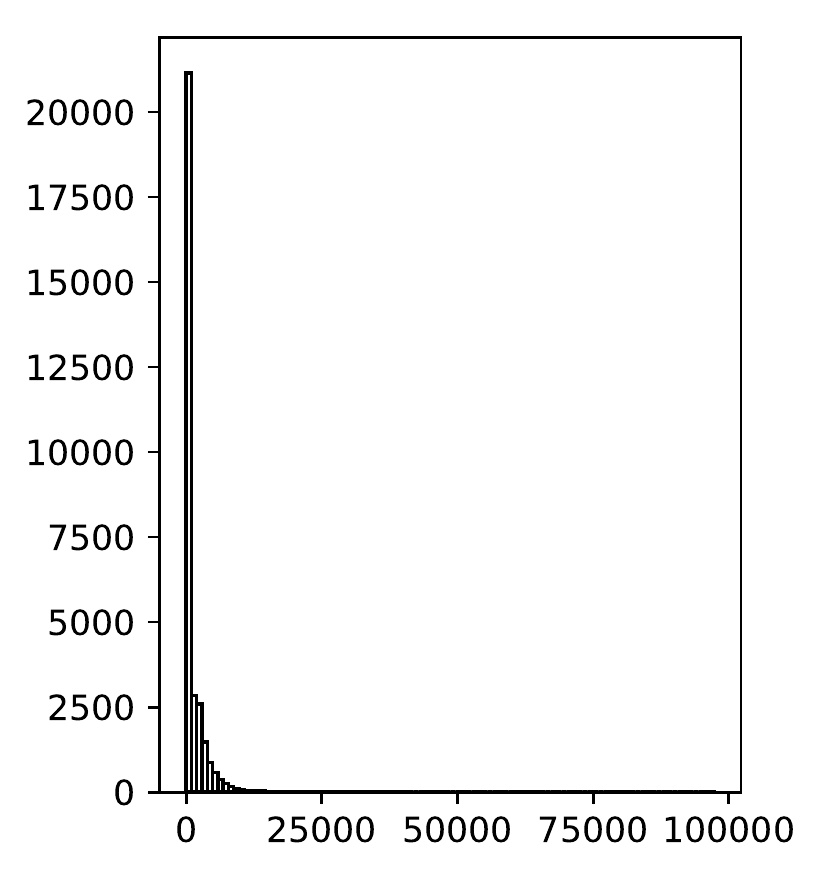}
        }%
    \hfill
    \subfloat{%
        \includegraphics[width=0.3\linewidth]{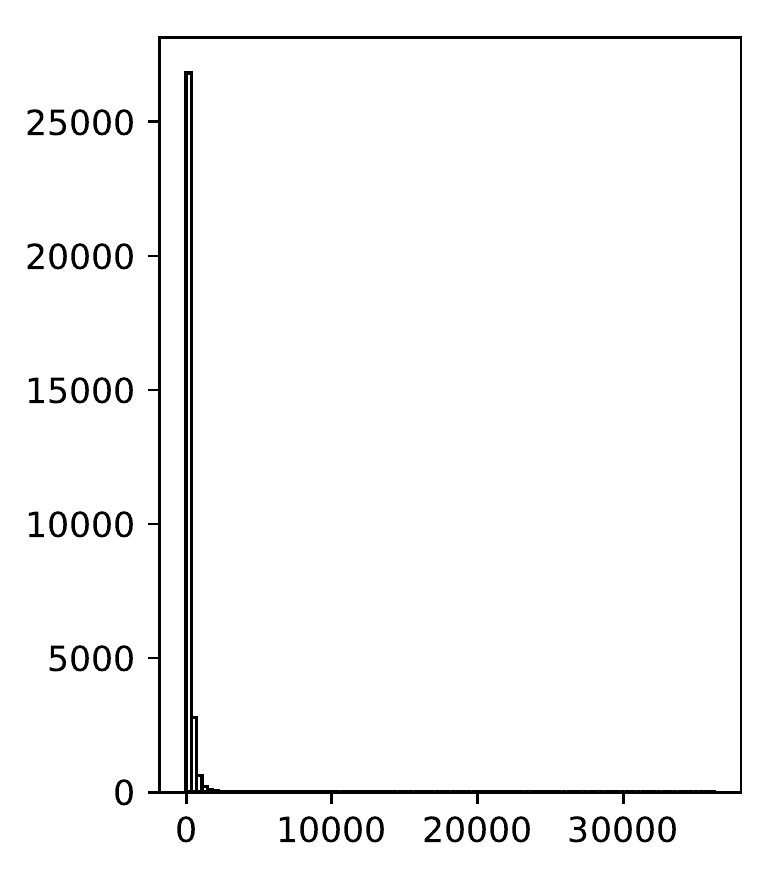}
      }%
   
   \caption{The histograms of three selected attributes of the dataset \textit{kddcup}.}
    \label{fig:histogram_kddcup}
\end{figure}

Table \ref{tab:exp_awa_bestpp} shows the best pipelines of each iteration with and without the AVATAR. This experiment results confirm that the most promising pipelines are often short pipelines (i.e. the pipelines only have one predictor/meta-predictor). There are two cases where the most promising pipelines include data preprocessing components. They are (kddcup, Iter 1) and (secom, Iter 2).
We try to interpret the semantics of the first pipeline structure, \textit{ClassBalancer $\rightarrow$ RemoveOutliers $\rightarrow$ InterquartileRange $\rightarrow$ AttributeSelection $\rightarrow$ Resample $\rightarrow$ PART}. 
The dataset \textit{kddcup} is imbalanced, therefore, using the ML component \textit{ClassBalancer} has a class balancing effect. Moreover, this dataset is high dimensional, therefore, using ML components such as \textit{InterquartileRange} and \textit{AttributeSelection} can reduce the number of dimensions. This is a large dataset; therefore, \textit{ReSample} can reduce the size of the dataset to accelerate the training of predictive models. Figure \ref{fig:histogram_kddcup} presents the histograms of three selected attributes of this dataset. We see that there are potential outliers which can negatively affect the ML model training. Therefore, using the component \textit{RemoveOutliers} can remove these outliers.

Table \ref{tab:exp_awa_details} shows the error rate and the number of evaluated pipelines of five iterations with and without the AVATAR. We group the experiment results into three main cases:

\begin{itemize}
    \item \textit{The error rate is lower with the AVATAR} (55/100 iterations) [Denoted by $\uparrow$]: In these cases, the AVATAR shows its effectiveness to quickly ignore invalid pipelines. Therefore, the number of evaluated pipelines are higher than without the AVATAR. As a result, the convergence of SMAC with AVATAR is faster. An example of these cases is the dataset \textit{cifar10small} with Iter 5. This case clearly shows that the number of evaluated pipelines is higher with the AVATAR than without, 50 compared to 27. As a result, the error rate is lower with the AVATAR, 64.94 compared to 84.29. The results confirm that within the same time budget, the AVATAR allows an ML pipeline and optimisation method (i.e., SMAC) to explore the search space faster than without it. 
    
    \item \textit{The same error rate with and without the AVATAR} (39/100 iterations) [Do not have either $\uparrow$ or $\downarrow$]:
    \begin{itemize}
        \item \textit{The cases with a similar number of evaluated pipelines.} (22/39 iterations): These are cases that have the difference between the number of pipelines less than or equal 10. An example of these cases is the dataset \textit{dorothea} with Iter 5. The error rate of both cases is 9.32\%. The number of evaluated pipelines with and without the AVATAR is 22 and 21 respectively. All of the datasets in this group are large and require long training time. Therefore, the number of evaluated pipelines are small. As a result, there is not a significant difference between the cases with and without the AVATAR.  
        
        
        \item \textit{The cases with significantly different number of evaluated pipelines.} (7/39 iterations): These are cases that have the difference between the number of pipelines greater than or equal 100. An example from this group is the dataset \textit{secom} with Iter 1. The error rate of both cases is 6.11\%, and the number of evaluated pipelines with and without the AVATAR are 307 and 108 respectively. We have observed that in this group of cases, the convergence to very good solutions was quick with and without the AVATAR because it is quite easy to find well-performing pipelines for these datasets.
  
    \end{itemize}
    
    \item \textit{The error rate is higher with the AVATAR} (6/100 iterations) [Denoted by $\downarrow$]: These cases are (kddcup, Iter 2), (abalone, Iter 3), (abalone, Iter 4), (adult, Iter 4), (shuttle, Iter 4) and (madelon, Iter 5). What is common for these cases is that they have quite a small number of evaluated pipelines and a similar number of evaluated pipelines. The reason is that the training time of these pipelines is significant. Therefore, there is an element of luck related to the initialisation and the benefits of quick validity check that AVATAR provides cannot be realised.
    
\end{itemize}

Table \ref{tab:exp_awa_details} also shows the not completed iterations (i.e., represented by - ) where no result is returned. The reason for that is that no valid, fully trained pipeline has been returned by SMAC. In these cases, we do not use these runs to calculate the error rate and relevant metrics.
There are 21 and 40 incomplete iterations with and without the AVATAR respectively. The number of incomplete cases is almost double without using the AVATAR. Using the AVATAR has helped SMAC to overcome these situations in 19 out of 40 incomplete iterations when AVATAR was not used. 

Figure \ref{fig:exp_convergence_confidence_avatar_vs_noavatar} visualises the convergence of the error rate of the best pipelines found with and without the AVATAR. The lower and upper boundaries representing the min and the max error rates of the five iterations are shown in the form of the shaded area. The solid and the dashed lines represent the mean of the error rate of the best pipelines of the five iterations found at the time points with and without the AVATAR respectively. There are 5/20 cases (\textit{adult}, \textit{car}, \textit{gisette}, \textit{krvskp}, \textit{secom}) for which SMAC has a similar convergence with and without the AVATAR. There are 15/20 cases for which SMAC converges faster by using the AVATAR and 2 hours of optimisation time. In addition, we can see that the shaded areas of SMAC with the AVATAR are narrower and lay lower than without the AVATAR in these 15 cases. It means that the stability of SMAC is better by using the AVATAR. For example, in the case of the dataset \textit{madelon}, SMAC without the AVATAR converges to the error rate between 20\% and 50\% after 80 minutes, whereas SMAC with the AVATAR converges to the error rate less than 20\% after 30 minutes.


Overall, we can observe that the use of AVATAR enables SMAC to find better pipelines faster. A disadvantage of SMAC is that SMAC does not work effectively with large datasets within a limited time budget. We can observe this for a number of cases which have a small number of evaluations (e.g., kddcup, Iter2). Although the AVATAR can quickly find valid pipelines in these cases, finding both valid and well-performing pipelines with limited time budget and resources (i.e., virtual machines) still remains a challenge for future studies.


\begin{table}[htb!]

\vspace{-0.35cm}
\caption {Mean, standard deviation (StdDev) and minimum (min) of the error rate, and mean, standard deviation and maximum (max) of the number of evaluated pipelines of AutoWeka4MCPS with (y) and without (n) the AVATAR using 2 hours time budget. The better results are in bold.}
\label{tab:exp_awa_2h}

\centering
\scriptsize

\begin{tabular}{|l|l|l|l|l|l|l|l|l|l|l|l|l|}
\hline
\multirow{3}{*}{\textbf{Dataset}} & \multicolumn{6}{l|}{\textbf{\% Error Rate}}                                                                   & \multicolumn{6}{l|}{\textbf{Number of Evaluated Pipelines}}                                                   \\ \cline{2-13} 
                                  & \multicolumn{2}{l|}{\textbf{Mean}} & \multicolumn{2}{l|}{\textbf{StdDev}} & \multicolumn{2}{l|}{\textbf{Min}} & \multicolumn{2}{l|}{\textbf{Mean}} & \multicolumn{2}{l|}{\textbf{StdDev}} & \multicolumn{2}{l|}{\textbf{Max}} \\ \cline{2-13} 
                                  & \textbf{y}       & \textbf{n}      & \textbf{y}        & \textbf{n}       & \textbf{y}      & \textbf{n}      & \textbf{y}        & \textbf{n}     & \textbf{y}        & \textbf{n}       & \textbf{y}        & \textbf{n}    \\ \hline
\textbf{abalone}                  & 73.86            & \textbf{73.47}  & 0.94              & 0.23             & \textbf{72.91}  & 73.22           & \textbf{106}      & 73             & 73                & 36               & \textbf{229}      & 126           \\ \hline
\textbf{adult}                    & 14.60            & \textbf{14.38}  & 0.60              & 0.70             & \textbf{13.41}  & \textbf{13.41}  & 60                & 32             & 28                & 7                & 93                & 45            \\ \hline
\textbf{amazon}                   & \textbf{46.03}   & 58.57           & 10.68             & 6.38             & \textbf{38.48}  & 52.19           & \textbf{44}       & 22             & 38                & 14               & \textbf{103}      & 36            \\ \hline
\textbf{car}                      & \textbf{0.58}    & 2.73            & 0.30              & 0.95             & \textbf{0.33}   & \textbf{1.07}   & \textbf{1136}     & 185            & 122               & 77               & \textbf{1269}     & 334           \\ \hline
\textbf{cifar10small}             & \textbf{71.94}   & 75.44           & 8.76              & 8.86             & \textbf{64.94}  & 66.58           & \textbf{35}       & 20             & 24                & 13               & \textbf{67}       & 41            \\ \hline
\textbf{convex}                   & \textbf{33.52}   & 38.99           & 5.74              & 2.39             & \textbf{25.69}  & 35.93           & \textbf{49}       & 24             & 30                & 12               & \textbf{82}       & 44            \\ \hline
\textbf{dexter}                   & \textbf{8.09}    & 8.81            & 0.85              & 0.00             & \textbf{6.90}   & \textbf{8.81}   & \textbf{33}       & 23             & 23                & 24               & 66                & 69            \\ \hline
\textbf{dorothea}                 & \textbf{8.02}    & 9.32            & 1.31              & 0.00             & \textbf{6.71}   & \textbf{9.32}   & \textbf{17}       & 12             & 14                & 9                & \textbf{42}       & 22            \\ \hline
\textbf{gcredit}                  & \textbf{22.26}   & 22.83           & 0.39              & 0.28             & \textbf{21.71}  & 22.43           & \textbf{808}      & 430            & 270               & 229              & \textbf{1298}     & 845           \\ \hline
\textbf{gisette}                  & \textbf{2.58}    & 2.93            & 0.19              & 0.54             & \textbf{2.39}   & \textbf{2.39}   & \textbf{27}       & 20             & 20                & 12               & \textbf{54}       & 39            \\ \hline
\textbf{kddcup}                   & \textbf{1.80}    & \textbf{1.80}   & 0.00              & 0.00             & \textbf{1.80}   & \textbf{1.80}   & \textbf{67}       & 30             & 33                & 24               & \textbf{108}      & 68            \\ \hline
\textbf{krvskp}                   & \textbf{0.44}    & 0.67            & 0.04              & 0.12             & \textbf{0.40}   & \textbf{0.49}   & \textbf{898}      & 118            & 128               & 40               & \textbf{1071}     & 173           \\ \hline
\textbf{madelon}                  & \textbf{22.86}   & 26.05           & 0.22              & 3.34             & \textbf{22.75}  & \textbf{22.75}  & \textbf{279}      & 49             & 59                & 24               & \textbf{350}      & 85            \\ \hline
\textbf{mnist}                    & \textbf{16.38}   & \textbf{-}      & 0.00              & -                & \textbf{16.38}  & \textbf{-}      & \textbf{23}       & 18             & 27                & 17               & \textbf{77}       & 50            \\ \hline
\textbf{secom}                    & \textbf{6.11}    & \textbf{6.11}   & 0.00              & 0.00             & \textbf{6.11}   & \textbf{6.11}   & \textbf{277}      & 81             & 160               & 47               & \textbf{490}      & 135           \\ \hline
\textbf{semeion}                  & \textbf{4.95}    & 8.37            & 0.41              & 1.98             & \textbf{4.66}   & 5.64            & \textbf{338}      & 54             & 118               & 36               & \textbf{515}      & 111           \\ \hline
\textbf{shuttle}                  & 0.09             & \textbf{0.03}   & 0.10              & 0.01             & 0.03            & \textbf{0.02}   & \textbf{91}       & 31             & 48                & 29               & \textbf{159}      & 73            \\ \hline
\textbf{waveform}                 & \textbf{12.53}   & 12.72           & 0.13              & 0.06             & \textbf{12.46}  & 12.66           & \textbf{333}      & 49             & 138               & 37               & \textbf{492}      & 109           \\ \hline
\textbf{winequality}              & \textbf{33.44}   & 37.85           & 0.38              & 5.65             & \textbf{33.10}  & 33.39           & \textbf{432}      & 52             & 208               & 42               & \textbf{744}      & 117           \\ \hline
\textbf{yeast}                    & \textbf{38.02}   & 39.15           & 0.60              & 0.53             & \textbf{36.96}  & 38.50           & \textbf{1116}     & 136            & 349               & 80               & \textbf{1785}     & 222           \\ \hline
\end{tabular}

\vspace{-0.35cm}

\end{table}

\begin{landscape}

\begin{table*}[htb!]

\vspace{-0.35cm}
\captionsetup{font=scriptsize}
\caption {The best pipelines of five iterations with and without the AVATAR.}
\label{tab:exp_awa_bestpp}

\centering
\tiny

\begin{tabular}{|l|l|l|l|l|l|l|l|l|l|l|}
\hline
\multirow{2}{*}{\textbf{Dataset}}                                     & \multicolumn{2}{l|}{\textbf{Iter 1}}                                                                                                                                                                                                                                                                                       & \multicolumn{2}{l|}{\textbf{Iter 2}}                                                                                               & \multicolumn{2}{l|}{\textbf{Iter 3}}                                                                                                 & \multicolumn{2}{l|}{\textbf{Iter 4}}                                                                                                 & \multicolumn{2}{l|}{\textbf{Iter 5}}                                                                                                 \\ \cline{2-11} 
                                                                      & \textbf{\begin{tabular}[c]{@{}l@{}}With\\ AVATAR\end{tabular}}                                                                                                             & \textbf{\begin{tabular}[c]{@{}l@{}}Without\\ AVATAR\end{tabular}}                                                                             & \textbf{\begin{tabular}[c]{@{}l@{}}With\\ AVATAR\end{tabular}} & \textbf{\begin{tabular}[c]{@{}l@{}}Without\\ AVATAR\end{tabular}} & \textbf{\begin{tabular}[c]{@{}l@{}}With\\ AVATAR\end{tabular}}   & \textbf{\begin{tabular}[c]{@{}l@{}}Without\\ AVATAR\end{tabular}} & \textbf{\begin{tabular}[c]{@{}l@{}}With\\ AVATAR\end{tabular}}   & \textbf{\begin{tabular}[c]{@{}l@{}}Without\\ AVATAR\end{tabular}} & \textbf{\begin{tabular}[c]{@{}l@{}}With\\ AVATAR\end{tabular}}   & \textbf{\begin{tabular}[c]{@{}l@{}}Without\\ AVATAR\end{tabular}} \\ \hline
\textbf{abalone}                                                      & SMO                                                                                                                                                                        & SimpleLogistic                                                                                                                                & DecisionTable                                                  & DecisionTable                                                     & NaiveBayes                                                       & Logistic                                                          & Kstar                                                            & SimpleLogistic                                                    & \textbf{-}                                                       & -                                                                 \\ \hline
\textbf{adult}                                                        & PART                                                                                                                                                                       & PART                                                                                                                                          & Logistic                                                       & Logistic                                                          & PART                                                             & Logistic                                                          & NaiveBayes                                                       & DecisionTable                                                     & SGD                                                              & -                                                                 \\ \hline
\textbf{amazon}                                                       & PART                                                                                                                                                                       & Jrip                                                                                                                                          & -                                                              & -                                                                 & \begin{tabular}[c]{@{}l@{}}NaiveBayes\\ Multinomial\end{tabular} & -                                                                 & \begin{tabular}[c]{@{}l@{}}NaiveBayes\\ Multinomial\end{tabular} & NaiveBayes                                                        & -                                                                & -                                                                 \\ \hline
\textbf{car}                                                          & SMO                                                                                                                                                                        & LMT                                                                                                                                           & SMO                                                            & LMT                                                               & SMO                                                              & SMO                                                               & SMO                                                              & SMO                                                               & SMO                                                              & SMO                                                               \\ \hline
\textbf{cifar10small}                                                 & -                                                                                                                                                                          & \textbf{-}                                                                                                                                    & -                                                              & -                                                                 & DecisionStump                                                    & -                                                                 & \textbf{-}                                                       & -                                                                 & RandomForest                                                     & DecisionStump                                                     \\ \hline
\textbf{convex}                                                       & RandomTree                                                                                                                                                                 & VotedPerceptron                                                                                                                               & Jrip                                                           & Jrip                                                              & \textbf{-}                                                       & -                                                                 & -                                                                & \textbf{-}                                                        & RandomForest                                                     & PART                                                              \\ \hline
\textbf{dexter}                                                       & SGD                                                                                                                                                                        & \begin{tabular}[c]{@{}l@{}}NaiveBayes\\ Multinomial\end{tabular}                                                                              & SimpleLogistic                                                 & -                                                                 & \textbf{-}                                                       & -                                                                 & \textbf{-}                                                       & \textbf{-}                                                        & \begin{tabular}[c]{@{}l@{}}NaiveBayes\\ Multinomial\end{tabular} & -                                                                 \\ \hline
\textbf{dorothea}                                                     & DecisionStump                                                                                                                                                              & \textbf{-}                                                                                                                                    & \textbf{-}                                                     & \textbf{-}                                                        & \textbf{-}                                                       & -                                                                 & -                                                                & -                                                                 & PART                                                             & PART                                                              \\ \hline
\textbf{gcredit}                                                      & RandomForest                                                                                                                                                               & NaiveBayes                                                                                                                                    & SMO                                                            & NaiveBayes                                                        & SMO                                                              & NaiveBayes                                                        & SMO                                                              & SMO                                                               & RandomForest                                                     & SMO                                                               \\ \hline
\textbf{gisette}                                                      & VotedPerceptron                                                                                                                                                            & VotedPerceptron                                                                                                                               & \textbf{-}                                                     & -                                                                 & -                                                                & -                                                                 & -                                                                & \textbf{-}                                                        & VotedPerceptron                                                  & SGD                                                               \\ \hline
\textbf{\begin{tabular}[c]{@{}l@{}}kddcup\\ \end{tabular}} & \begin{tabular}[c]{@{}l@{}}ClassBalancer\\  $\rightarrow$ RemoveOutliers\\  $\rightarrow$ InterquartileRange\\  $\rightarrow$ AttributeSelection\\  $\rightarrow$ Resample\\  $\rightarrow$ PART\end{tabular} & \textbf{-}                                                                                                                                    & -                                                              & DecisionStump                                                     & ZeroR                                                            & DecisionStump                                                     & DecisionStump                                                    & -                                                                 & -                                                                & -                                                                 \\ \hline
\textbf{krvskp}                                                       & J48                                                                                                                                                                        & J48                                                                                                                                           & J48                                                            & LMT                                                               & J48                                                              & LMT                                                               & J48                                                              & J48                                                               & J48                                                              & RandomForest                                                      \\ \hline
\textbf{madelon}                                                      & Jrip                                                                                                                                                                       & REPTree                                                                                                                                       & Jrip                                                           & REPTree                                                           & Jrip                                                             & -                                                                 & Jrip                                                             & Jrip                                                              & Jrip                                                             & Jrip                                                              \\ \hline
\textbf{mnist}                                                        & -                                                                                                                                                                          & \textbf{-}                                                                                                                                    & -                                                              & \textbf{-}                                                        & PART                                                             & -                                                                 & -                                                                & \textbf{-}                                                        & -                                                                & -                                                                 \\ \hline
\textbf{secom}                                                        & SimpleLogistic                                                                                                                                                             & \begin{tabular}[c]{@{}l@{}}ClassBalancer\\  $\rightarrow$ EMImputation\\  $\rightarrow$ Normalize\\  $\rightarrow$ PrincipalComponent\\  $\rightarrow$ Kstar\end{tabular} & DecisionStump                                                  & -                                                                 & SMO                                                              & SMO                                                               & RandomForest                                                     & DecisionStump                                                     & REPTree                                                          & J48                                                               \\ \hline
\textbf{semeion}                                                      & SMO                                                                                                                                                                        & SMO                                                                                                                                           & SMO                                                            & -                                                                 & SMO                                                              & SimpleLogistic                                                    & SMO                                                              & \begin{tabular}[c]{@{}l@{}}Multilayer\\ Perceptron\end{tabular}   & SMO                                                              & LMT                                                               \\ \hline
\textbf{shuttle}                                                      & RandomForest                                                                                                                                                               & REPTree                                                                                                                                       & RandomForest                                                   & -                                                                 & DecisionTable                                                    & -                                                                 & REPTree                                                          & RandomForest                                                      & -                                                                & \textbf{-}                                                        \\ \hline
\textbf{waveform}                                                     & LMT                                                                                                                                                                        & Logistic                                                                                                                                      & SimpleLogistic                                                 & -                                                                 & SimpleLogistic                                                   & -                                                                 & SimpleLogistic                                                   & LMT                                                               & SimpleLogistic                                                   & SimpleLogistic                                                    \\ \hline
\textbf{winequality}                                                  & RandomForest                                                                                                                                                               & RandomForest                                                                                                                                  & RandomForest                                                   & -                                                                 & RandomForest                                                     & -                                                                 & Kstar                                                            & Logistic                                                          & Kstar                                                            & RandomForest                                                      \\ \hline
\textbf{yeast}                                                        & RandomForest                                                                                                                                                               & -                                                                                                                                             & RandomForest                                                   & -                                                                 & RandomForest                                                     & RandomForest                                                      & RandomForest                                                     & SMO                                                               & SMO                                                              & SMO                                                               \\ \hline
\end{tabular}

\end{table*}

\end{landscape}

\begin{landscape}

\begin{table*}[htb!]

\vspace{-0.35cm}
\captionsetup{font=scriptsize}
\caption {Error rate (\%) and the number of evaluated pipelines of five iterations with (y) and without (n) the AVATAR. The better results are in bold. $\uparrow$ and $\downarrow$ shows the lower and higher error rate by using the AVATAR respectively.}
\label{tab:exp_awa_details}

\centering
\scriptsize
\begin{tabular}{|l|l|l|l|l|l|l|l|l|l|l|l|l|l|l|l|l|l|l|l|l|}
\hline
{\color{black} }                                   & \multicolumn{4}{l|}{{\color{black} \textbf{Iter 1}}}                                                                                            & \multicolumn{4}{l|}{{\color{black} \textbf{Iter 2}}}                                                                                            & \multicolumn{4}{l|}{{\color{black} \textbf{Iter 3}}}                                                                                            & \multicolumn{4}{l|}{{\color{black} \textbf{Iter 4}}}                                                                                            & \multicolumn{4}{l|}{{\color{black} \textbf{Iter 5}}}                                                                                            \\ \cline{2-21} 
{\color{black} }                                   & \multicolumn{2}{l|}{{\color{black} \textbf{\% error}}}                  & \multicolumn{2}{l|}{{\color{black} \textbf{pipelines}}}        & \multicolumn{2}{l|}{{\color{black} \textbf{\% error}}}                  & \multicolumn{2}{l|}{{\color{black} \textbf{pipelines}}}        & \multicolumn{2}{l|}{{\color{black} \textbf{\% error}}}                  & \multicolumn{2}{l|}{{\color{black} \textbf{pipelines}}}        & \multicolumn{2}{l|}{{\color{black} \textbf{\% error}}}                  & \multicolumn{2}{l|}{{\color{black} \textbf{pipelines}}}        & \multicolumn{2}{l|}{{\color{black} \textbf{\% error}}}                  & \multicolumn{2}{l|}{{\color{black} \textbf{pipelines}}}        \\ \cline{2-21} 
\multirow{-3}{*}{{\color{black} \textbf{Dataset}}} & {\color{black} \textbf{y}}      & {\color{black} \textbf{n}}     & {\color{black} \textbf{y}} & {\color{black} \textbf{n}} & {\color{black} \textbf{y}}      & {\color{black} \textbf{n}}     & {\color{black} \textbf{y}} & {\color{black} \textbf{n}} & {\color{black} \textbf{y}}      & {\color{black} \textbf{n}}     & {\color{black} \textbf{y}} & {\color{black} \textbf{n}} & {\color{black} \textbf{y}}      & {\color{black} \textbf{n}}     & {\color{black} \textbf{y}} & {\color{black} \textbf{n}} & {\color{black} \textbf{y}}      & {\color{black} \textbf{n}}     & {\color{black} \textbf{y}} & {\color{black} \textbf{n}} \\ \hline
{\color{black} \textbf{abalone}}                   & {\color{black} \textbf{$\uparrow$72.91}} & {\color{black} 73.26}          & {\color{black} 229}        & {\color{black} 63}         & {\color{black} \textbf{73.60}}  & {\color{black} \textbf{73.60}} & {\color{black} 132}        & {\color{black} 91}         & {\color{black} $\downarrow$75.41}          & {\color{black} \textbf{73.78}} & {\color{black} 48}         & {\color{black} 69}         & {\color{black} $\downarrow$73.50}          & {\color{black} \textbf{73.22}} & {\color{black} 99}         & {\color{black} 126}        & {\color{black} -}      & {\color{black} -}              & {\color{black} 21}         & {\color{black} 16}         \\ \hline
{\color{black} \textbf{adult}}                     & {\color{black} \textbf{13.41}}  & {\color{black} \textbf{13.41}} & {\color{black} 82}         & {\color{black} 30}         & {\color{black} \textbf{14.89}}  & {\color{black} \textbf{14.89}} & {\color{black} 27}         & {\color{black} 27}         & {\color{black} \textbf{$\uparrow$14.11}} & {\color{black} 14.89}          & {\color{black} 70}         & {\color{black} 45}         & {\color{black} $\downarrow$14.82}          & {\color{black} \textbf{14.33}} & {\color{black} 93}         & {\color{black} 33}         & {\color{black} \textbf{$\uparrow$15.79}} & {\color{black} -}              & {\color{black} 28}         & {\color{black} 23}         \\ \hline
{\color{black} \textbf{amazon}}                    & {\color{black} \textbf{$\uparrow$61.14}} & {\color{black} 64.95}          & {\color{black} 37}         & {\color{black} 35}         & {\color{black} -}      & {\color{black} -}     & {\color{black} 10}         & {\color{black} 10}         & {\color{black} \textbf{$\uparrow$38.48}} & {\color{black} -}              & {\color{black} 103}        & {\color{black} 26}         & {\color{black} \textbf{$\uparrow$38.48}} & {\color{black} 52.19}          & {\color{black} 71}         & {\color{black} 36}         & {\color{black} -}               & {\color{black} -}              & {\color{black} 1}          & {\color{black} 1}          \\ \hline
{\color{black} \textbf{car}}                       & {\color{black} \textbf{$\uparrow$0.33}}  & {\color{black} 3.88}           & {\color{black} 1209}       & {\color{black} 116}        & {\color{black} \textbf{$\uparrow$0.41}}  & {\color{black} 3.31}           & {\color{black} 1269}       & {\color{black} 149}        & {\color{black} \textbf{$\uparrow$1.16}}  & {\color{black} 2.48}           & {\color{black} 1123}       & {\color{black} 164}        & {\color{black} \textbf{$\uparrow$0.59}}  & {\color{black} 1.07}           & {\color{black} 912}        & {\color{black} 334}        & {\color{black} \textbf{$\uparrow$0.41}}  & {\color{black} 2.89}           & {\color{black} 1167}       & {\color{black} 160}        \\ \hline
{\color{black} \textbf{cifar10small}}              & {\color{black} -}      & {\color{black} -}     & {\color{black} 8}          & {\color{black} 8}          & {\color{black} \textbf{66.58}}  & {\color{black} \textbf{66.58}} & {\color{black} 44}         & {\color{black} 41}         & {\color{black} \textbf{$\uparrow$84.29}} & {\color{black} -}              & {\color{black} 67}         & {\color{black} 16}         & {\color{black} -}      & {\color{black} -}              & {\color{black} 8}          & {\color{black} 7}          & {\color{black} \textbf{$\uparrow$64.94}} & {\color{black} 84.29}          & {\color{black} 50}         & {\color{black} 27}         \\ \hline
{\color{black} \textbf{convex}}                    & {\color{black} \textbf{$\uparrow$35.59}} & {\color{black} 35.93}          & {\color{black} 71}         & {\color{black} 29}         & {\color{black} \textbf{39.28}}  & {\color{black} \textbf{39.28}} & {\color{black} 66}         & {\color{black} 44}         & {\color{black} -}      & {\color{black} -}              & {\color{black} 14}         & {\color{black} 7}          & {\color{black} -}               & {\color{black} -}     & {\color{black} 10}         & {\color{black} 17}         & {\color{black} \textbf{$\uparrow$25.69}} & {\color{black} 41.76}          & {\color{black} 82}         & {\color{black} 25}         \\ \hline
{\color{black} \textbf{dexter}}                    & {\color{black} \textbf{$\uparrow$6.90}}  & {\color{black} 8.81}           & {\color{black} 66}         & {\color{black} 69}         & {\color{black} \textbf{$\uparrow$8.57}}  & {\color{black} -}              & {\color{black} 52}         & {\color{black} 21}         & {\color{black} -}      & {\color{black} -}              & {\color{black} 19}         & {\color{black} 4}          & {\color{black} -}      & {\color{black} -}     & {\color{black} 4}          & {\color{black} 4}          & {\color{black} \textbf{$\uparrow$8.81}}  & {\color{black} -}              & {\color{black} 24}         & {\color{black} 19}         \\ \hline
{\color{black} \textbf{dorothea}}                  & {\color{black} \textbf{$\uparrow$6.71}}  & {\color{black} -}     & {\color{black} 42}         & {\color{black} 22}         & {\color{black} -}      & {\color{black} -}     & {\color{black} 12}         & {\color{black} 12}         & {\color{black} -}      & {\color{black} -}              & {\color{black} 8}          & {\color{black} 4}          & {\color{black} -}               & {\color{black} -}              & {\color{black} 1}          & {\color{black} 1}          & {\color{black} \textbf{9.32}}   & {\color{black} \textbf{9.32}}  & {\color{black} 22}         & {\color{black} 21}         \\ \hline
{\color{black} \textbf{gcredit}}                   & {\color{black} \textbf{$\uparrow$22.86}} & {\color{black} 23.14}          & {\color{black} 503}        & {\color{black} 389}        & {\color{black} \textbf{$\uparrow$21.71}} & {\color{black} 23.00}          & {\color{black} 861}        & {\color{black} 261}        & {\color{black} \textbf{$\uparrow$22.00}} & {\color{black} 23.00}          & {\color{black} 686}        & {\color{black} 464}        & {\color{black} \textbf{$\uparrow$22.29}} & {\color{black} 22.57}          & {\color{black} 691}        & {\color{black} 189}        & {\color{black} \textbf{22.43}}  & {\color{black} \textbf{22.43}} & {\color{black} 1298}       & {\color{black} 845}        \\ \hline
{\color{black} \textbf{gisette}}                   & {\color{black} \textbf{2.39}}   & {\color{black} \textbf{2.39}}  & {\color{black} 47}         & {\color{black} 27}         & {\color{black} -}      & {\color{black} -}              & {\color{black} 21}         & {\color{black} 19}         & {\color{black} -}      & {\color{black} -}              & {\color{black} 8}          & {\color{black} 6}          & {\color{black} -}      & {\color{black} -}     & {\color{black} 6}          & {\color{black} 7}          & {\color{black} \textbf{$\uparrow$2.76}}  & {\color{black} 3.47}           & {\color{black} 54}         & {\color{black} 39}         \\ \hline
{\color{black} \textbf{kddcup}}                    & {\color{black} \textbf{$\uparrow$1.80}}  & {\color{black} -}     & {\color{black} 108}        & {\color{black} 26}         & {\color{black} $\downarrow$-}              & {\color{black} \textbf{1.80}}  & {\color{black} 66}         & {\color{black} 68}         & {\color{black} \textbf{1.80}}   & {\color{black} \textbf{1.80}}  & {\color{black} 83}         & {\color{black} 43}         & {\color{black} \textbf{$\uparrow$1.80}}  & {\color{black} -}              & {\color{black} 73}         & {\color{black} 4}          & {\color{black} -}      & {\color{black} -}              & {\color{black} 7}          & {\color{black} 7}          \\ \hline
{\color{black} \textbf{krvskp}}                    & {\color{black} \textbf{$\uparrow$0.40}}  & {\color{black} 0.72}           & {\color{black} 761}        & {\color{black} 56}         & {\color{black} \textbf{$\uparrow$0.40}}  & {\color{black} 0.58}           & {\color{black} 997}        & {\color{black} 173}        & {\color{black} \textbf{$\uparrow$0.49}}  & {\color{black} 0.80}           & {\color{black} 1071}       & {\color{black} 139}        & {\color{black} \textbf{$\uparrow$0.40}}  & {\color{black} 0.76}           & {\color{black} 918}        & {\color{black} 94}         & {\color{black} \textbf{0.49}}   & {\color{black} \textbf{0.49}}  & {\color{black} 745}        & {\color{black} 127}        \\ \hline
{\color{black} \textbf{madelon}}                   & {\color{black} \textbf{$\uparrow$22.75}} & {\color{black} 30.11}          & {\color{black} 253}        & {\color{black} 58}         & {\color{black} \textbf{$\uparrow$22.75}} & {\color{black} 28.57}          & {\color{black} 326}        & {\color{black} 50}         & {\color{black} \textbf{$\uparrow$22.75}} & {\color{black} -}              & {\color{black} 350}        & {\color{black} 10}         & {\color{black} \textbf{22.75}}  & {\color{black} \textbf{22.75}} & {\color{black} 182}        & {\color{black} 44}         & {\color{black} $\downarrow$23.30}          & {\color{black} \textbf{22.75}} & {\color{black} 282}        & {\color{black} 85}         \\ \hline
{\color{black} \textbf{mnist}}                     & {\color{black} -}      & {\color{black} -}     & {\color{black} 13}         & {\color{black} 13}         & {\color{black} -}               & {\color{black} -}     & {\color{black} 11}         & {\color{black} 50}         & {\color{black} \textbf{$\uparrow$16.38}} & {\color{black} -}              & {\color{black} 77}         & {\color{black} 11}         & {\color{black} -}      & {\color{black} -}     & {\color{black} 13}         & {\color{black} 13}         & {\color{black} -}      & {\color{black} -}     & {\color{black} 2}          & {\color{black} 2}          \\ \hline
{\color{black} \textbf{secom}}                     & {\color{black} \textbf{6.11}}   & {\color{black} \textbf{6.11}}  & {\color{black} 307}        & {\color{black} 108}        & {\color{black} \textbf{$\uparrow$6.11}}  & {\color{black} -}              & {\color{black} 37}         & {\color{black} 16}         & {\color{black} \textbf{6.11}}   & {\color{black} \textbf{6.11}}  & {\color{black} 490}        & {\color{black} 114}        & {\color{black} \textbf{6.11}}   & {\color{black} \textbf{6.11}}  & {\color{black} 164}        & {\color{black} 135}        & {\color{black} \textbf{6.11}}   & {\color{black} \textbf{6.11}}  & {\color{black} 385}        & {\color{black} 34}         \\ \hline
{\color{black} \textbf{semeion}}                   & {\color{black} \textbf{$\uparrow$4.93}}  & {\color{black} 5.64}           & {\color{black} 339}        & {\color{black} 111}        & {\color{black} \textbf{$\uparrow$4.75}}  & {\color{black} -}              & {\color{black} 341}        & {\color{black} 5}          & {\color{black} \textbf{$\uparrow$5.74}}  & {\color{black} 10.03}          & {\color{black} 515}        & {\color{black} 45}         & {\color{black} \textbf{$\uparrow$4.66}}  & {\color{black} 7.34}           & {\color{black} 353}        & {\color{black} 71}         & {\color{black} \textbf{$\uparrow$4.66}}  & {\color{black} 10.48}          & {\color{black} 143}        & {\color{black} 37}         \\ \hline
{\color{black} \textbf{shuttle}}                   & {\color{black} \textbf{0.04}}   & {\color{black} \textbf{0.04}}  & {\color{black} 92}         & {\color{black} 73}         & {\color{black} \textbf{$\uparrow$0.03}}  & {\color{black} -}              & {\color{black} 117}        & {\color{black} 2}          & {\color{black} \textbf{$\uparrow$0.26}}  & {\color{black} -}              & {\color{black} 72}         & {\color{black} 11}         & {\color{black} $\downarrow$0.03}           & {\color{black} \textbf{0.02}}  & {\color{black} 159}        & {\color{black} 59}         & {\color{black} -}      & {\color{black} -}     & {\color{black} 16}         & {\color{black} 11}         \\ \hline
{\color{black} \textbf{waveform}}                  & {\color{black} \textbf{12.80}}  & {\color{black} \textbf{12.80}} & {\color{black} 84}         & {\color{black} 60}         & {\color{black} \textbf{$\uparrow$12.46}} & {\color{black} -}              & {\color{black} 325}        & {\color{black} 10}         & {\color{black} \textbf{$\uparrow$12.46}} & {\color{black} -}              & {\color{black} 492}        & {\color{black} 11}         & {\color{black} \textbf{$\uparrow$12.46}} & {\color{black} 12.66}          & {\color{black} 420}        & {\color{black} 109}        & {\color{black} \textbf{$\uparrow$12.49}} & {\color{black} 12.69}          & {\color{black} 346}        & {\color{black} 53}         \\ \hline
{\color{black} \textbf{winequality}}               & {\color{black} \textbf{33.39}}  & {\color{black} \textbf{33.39}} & {\color{black} 582}        & {\color{black} 117}        & {\color{black} \textbf{$\uparrow$34.18}} & {\color{black} -}              & {\color{black} 194}        & {\color{black} 2}          & {\color{black} \textbf{$\uparrow$33.10}} & {\color{black} -}              & {\color{black} 405}        & {\color{black} 12}         & {\color{black} \textbf{$\uparrow$33.30}} & {\color{black} 45.82}          & {\color{black} 744}        & {\color{black} 77}         & {\color{black} \textbf{$\uparrow$33.22}} & {\color{black} 34.33}          & {\color{black} 234}        & {\color{black} 52}         \\ \hline
{\color{black} \textbf{yeast}}                     & {\color{black} \textbf{$\uparrow$36.96}} & {\color{black} 39.94}          & {\color{black} 776}        & {\color{black} 217}        & {\color{black} \textbf{$\uparrow$38.59}} & {\color{black} -}              & {\color{black} 1092}       & {\color{black} 5}          & {\color{black} \textbf{$\uparrow$37.83}} & {\color{black} 38.50}          & {\color{black} 954}        & {\color{black} 132}        & {\color{black} \textbf{$\uparrow$38.11}} & {\color{black} 39.27}          & {\color{black} 1785}       & {\color{black} 222}        & {\color{black} \textbf{$\uparrow$38.60}} & {\color{black} 38.89}          & {\color{black} 972}        & {\color{black} 105}        \\ \hline
\end{tabular}

\end{table*}

\end{landscape}

\begin{figure}[!htbp] 
  
    \subfloat[abalone]{%
        \includegraphics[width=0.45\linewidth]{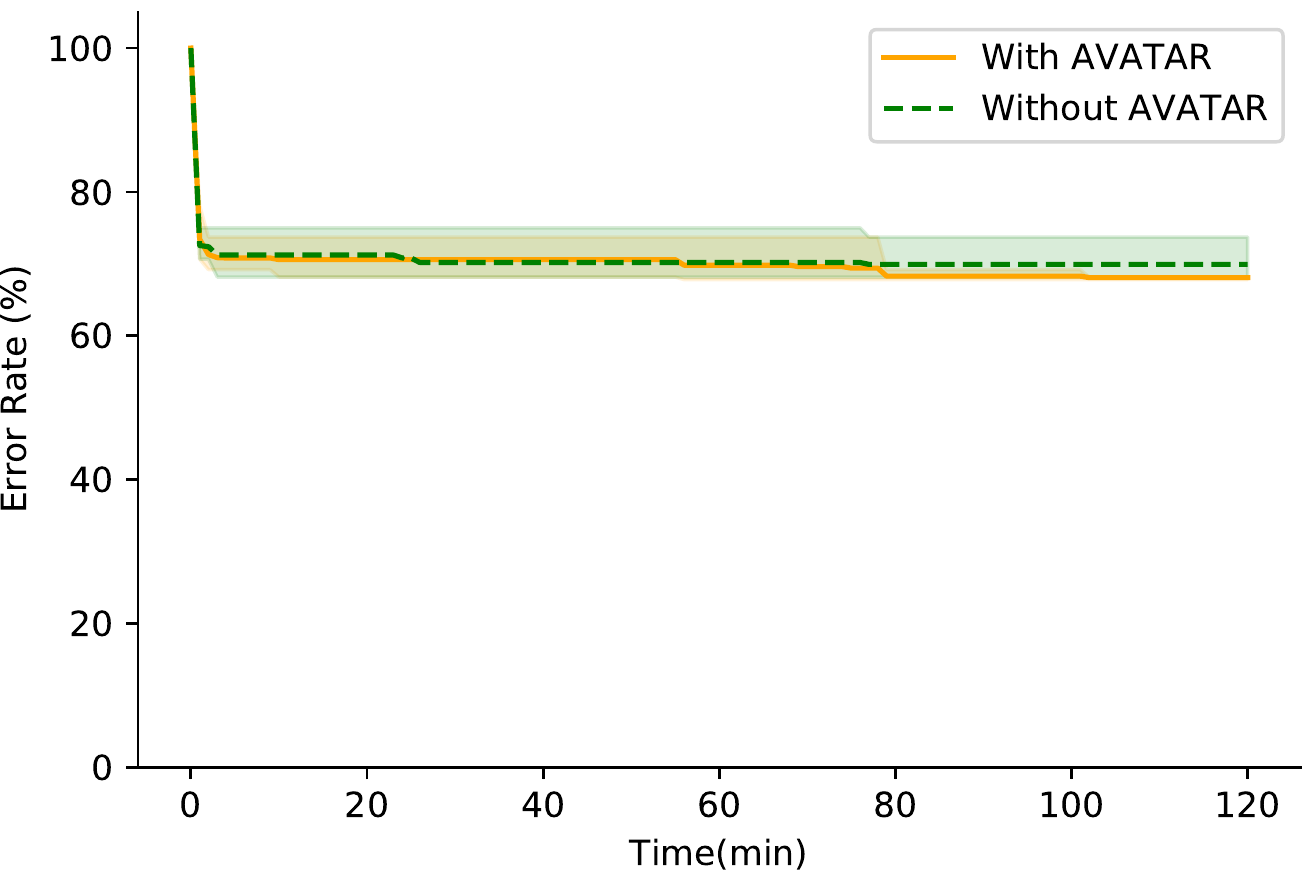}
      }%
    \hfill
    \subfloat[adult]{%
        \includegraphics[width=0.45\linewidth]{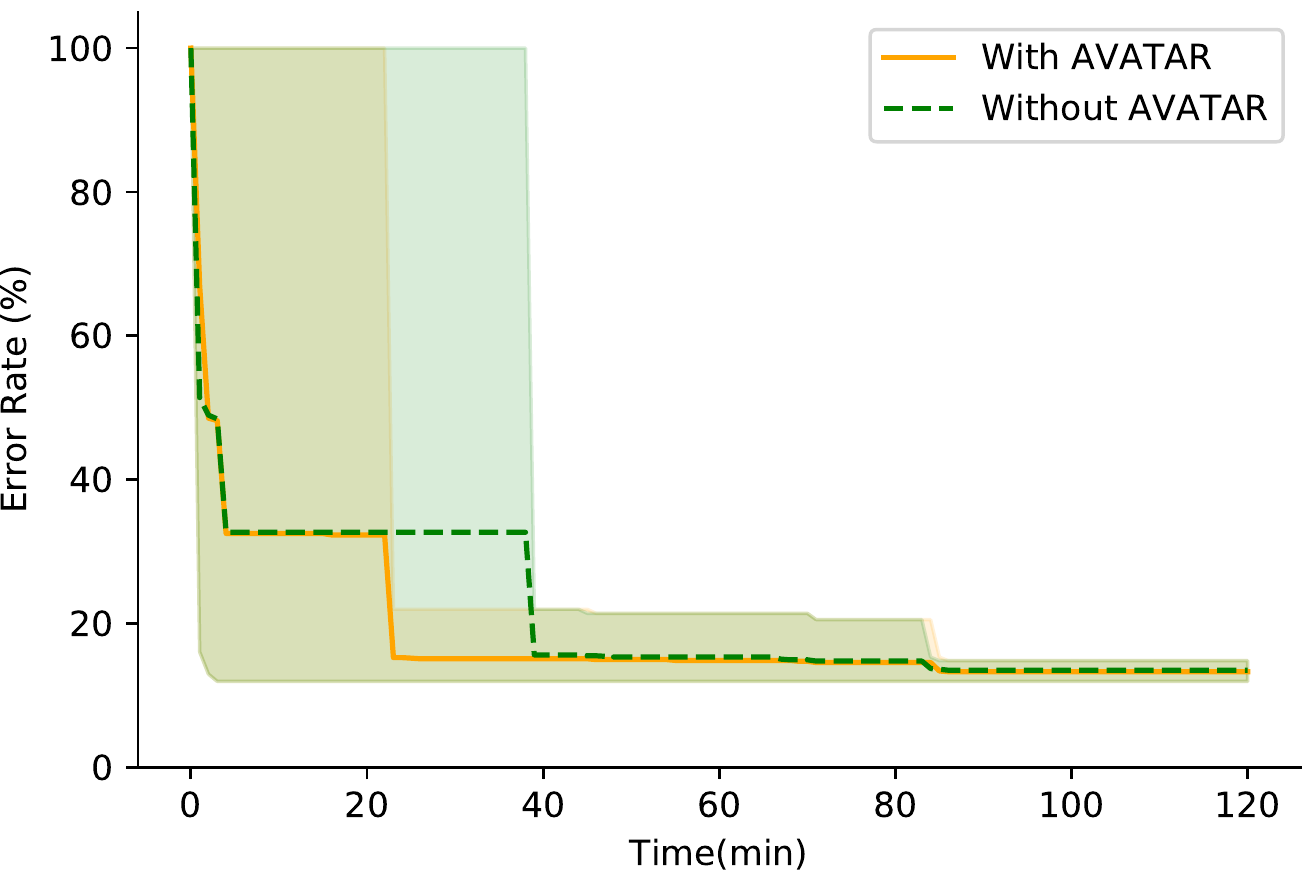}
      }%
      
    \subfloat[amazon]{%
        \includegraphics[width=0.45\linewidth]{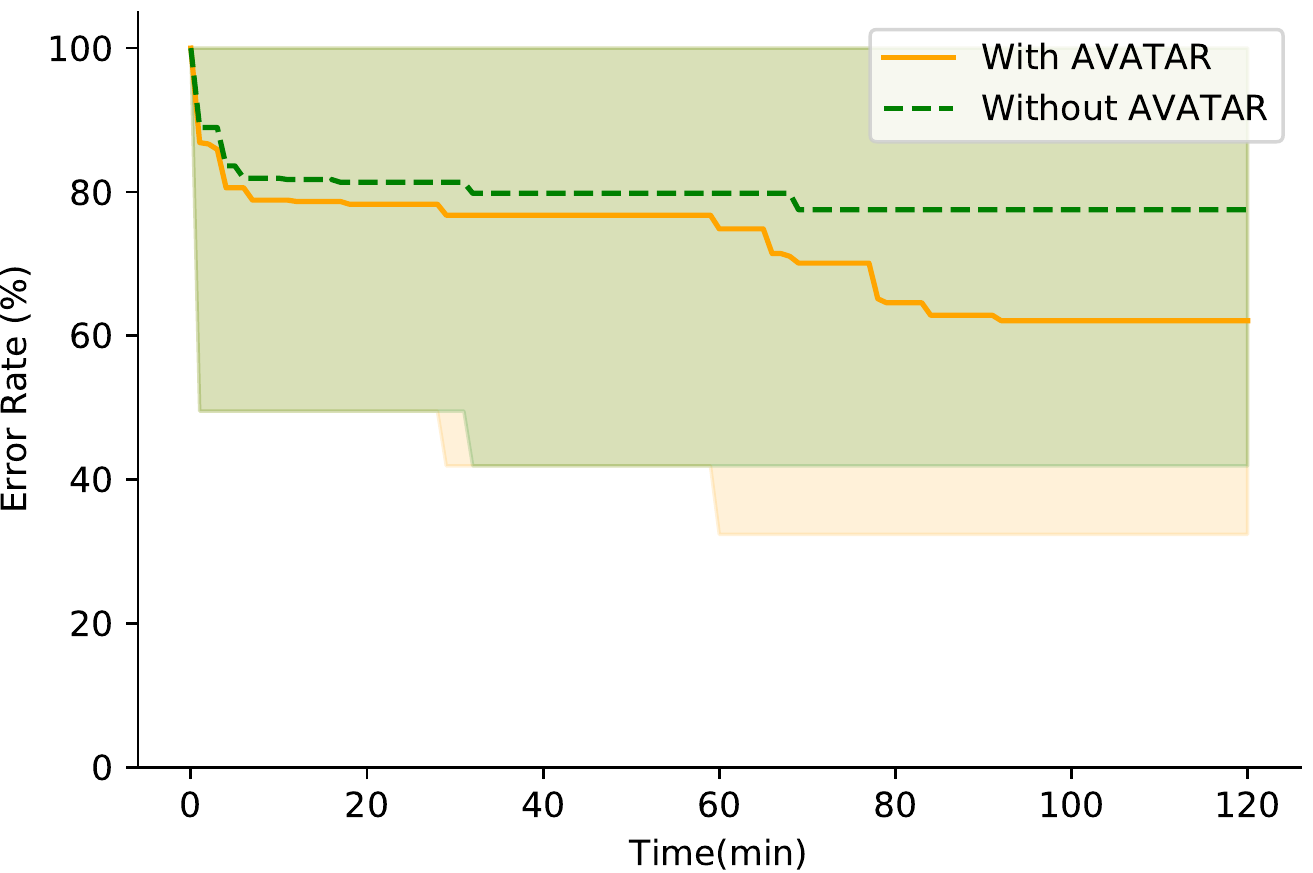}
        }%
    \hfill
    \subfloat[car]{%
        \includegraphics[width=0.45\linewidth]{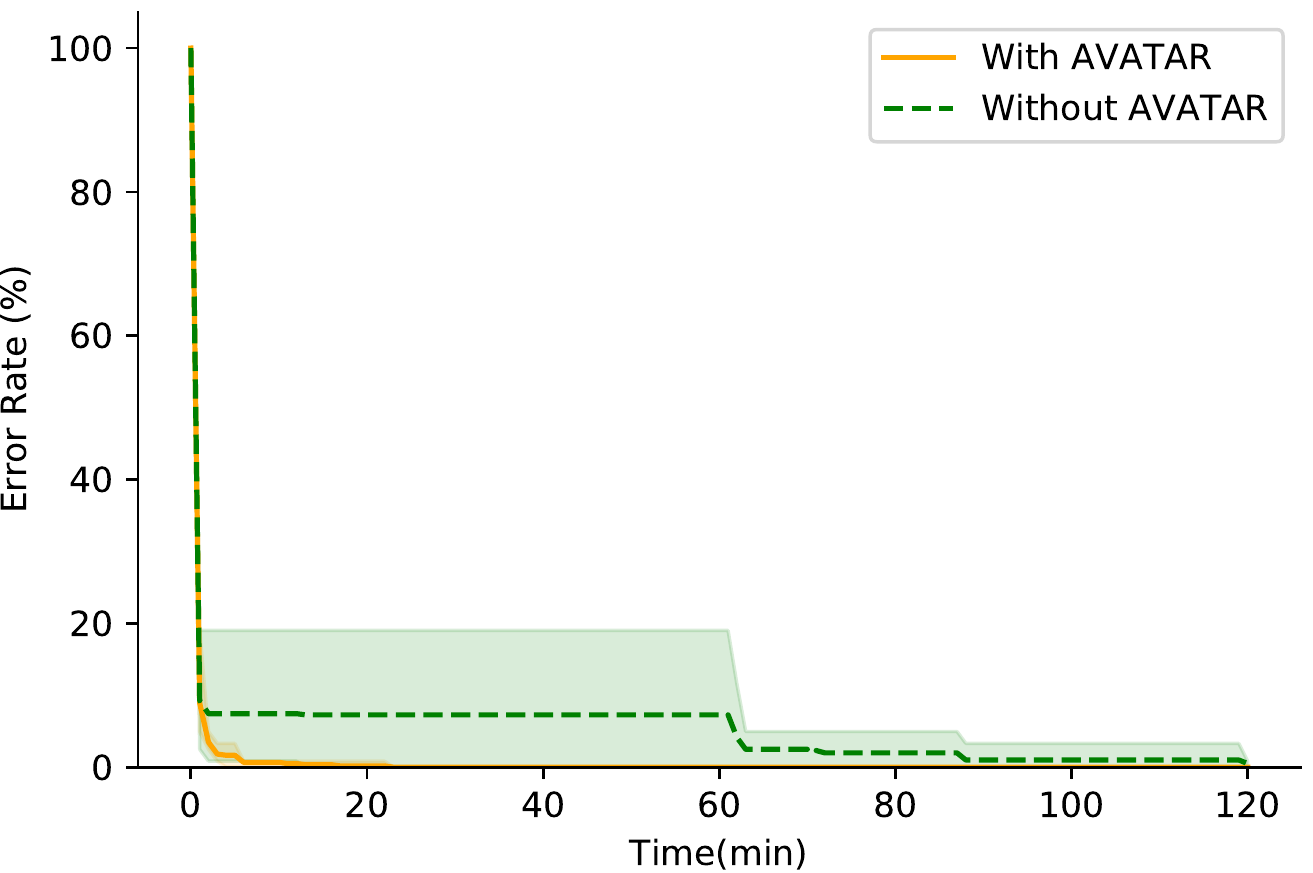}
      }%
    
    \subfloat[cifar10small]{%
        \includegraphics[width=0.45\linewidth]{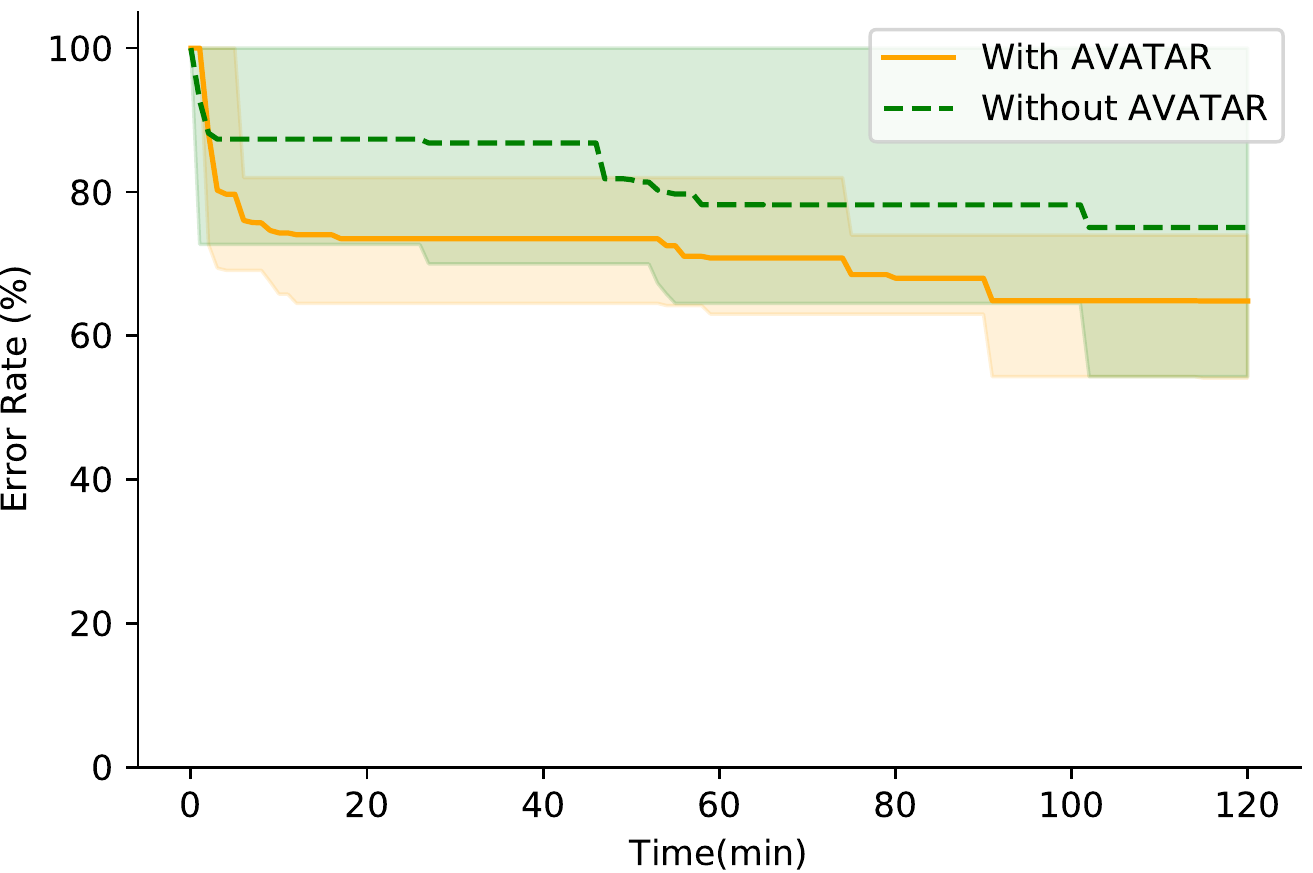}
        }%
    \hfill    
    \subfloat[convex]{%
        \includegraphics[width=0.45\linewidth]{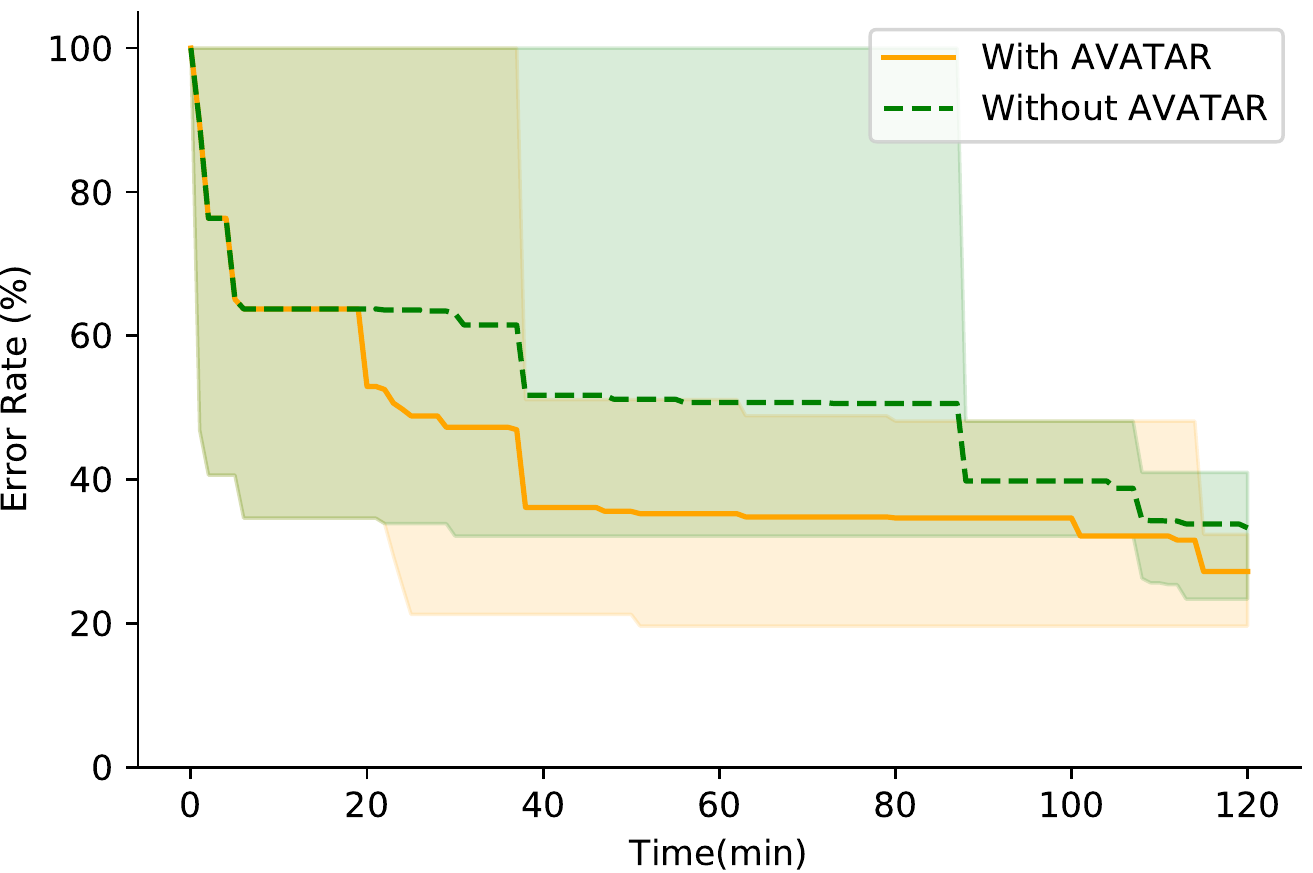}
      }%
    
    \subfloat[dexter]{%
        \includegraphics[width=0.45\linewidth]{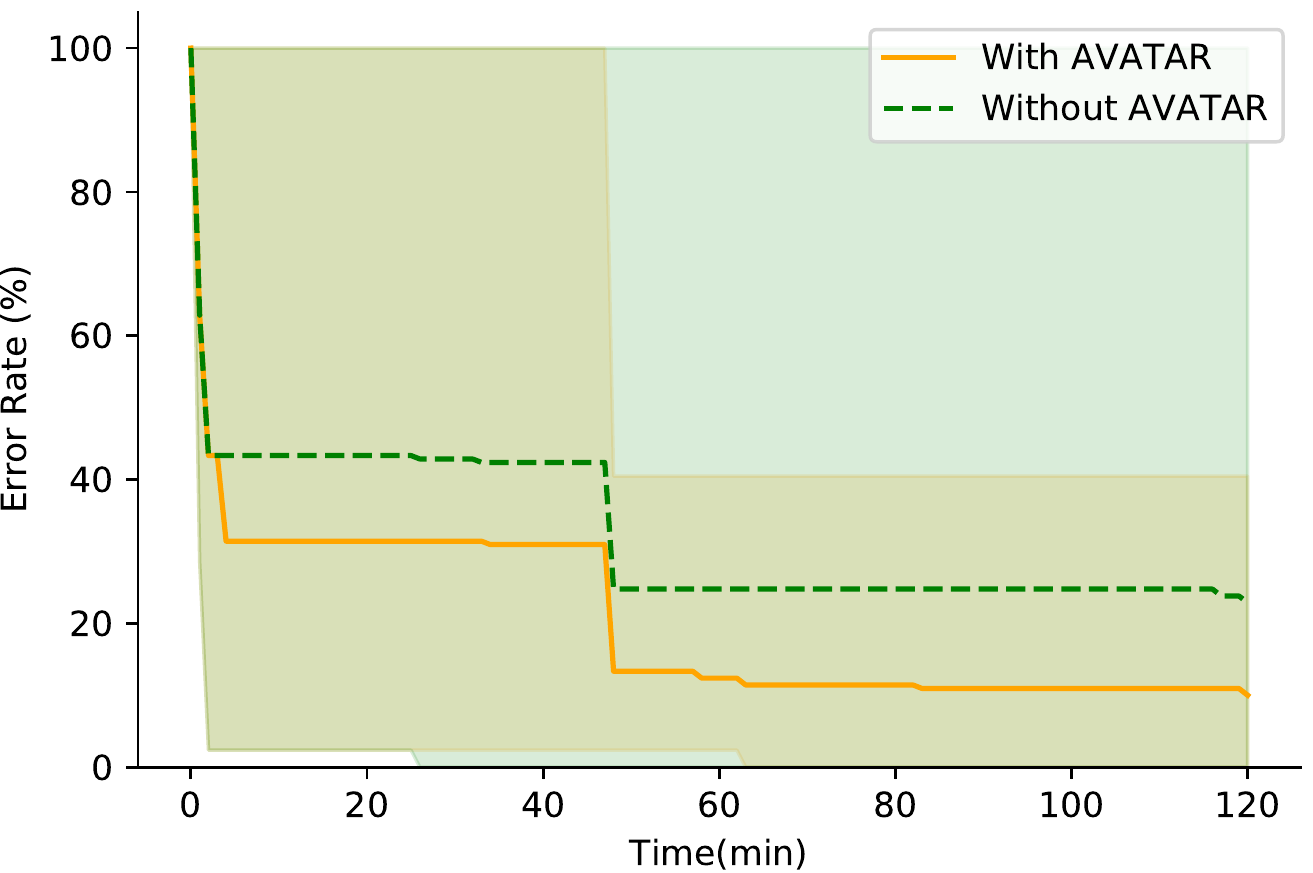}
        }%
     \hfill   
     \subfloat[dorothea]{%
        \includegraphics[width=0.45\linewidth]{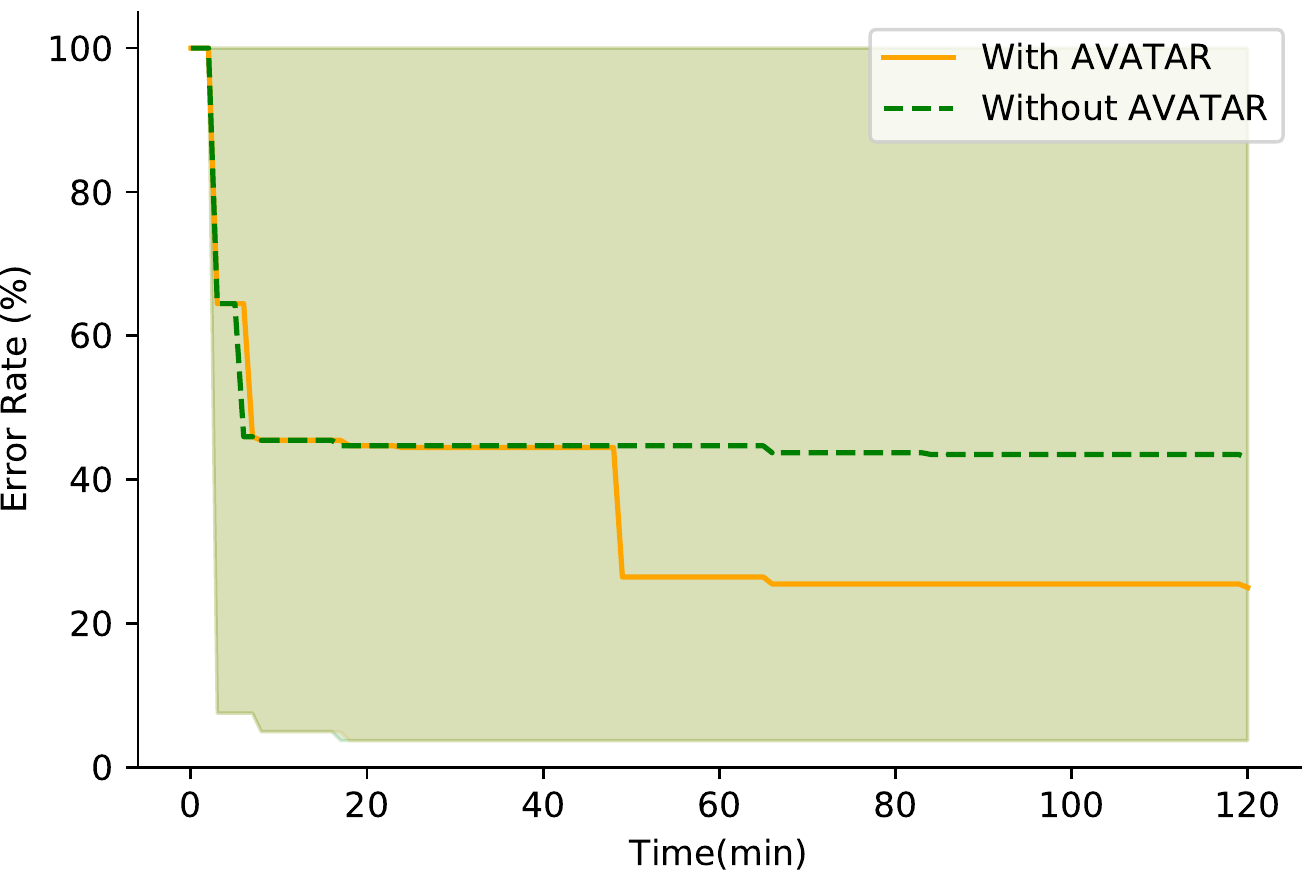}
      }%
      
       \caption{Compare the convergence of the error rate of the best pipelines found with and without the AVATAR.}
    \label{fig:exp_convergence_confidence_avatar_vs_noavatar}
\end{figure}

 \begin{figure}[!htbp]     
  \ContinuedFloat    
     \subfloat[gisette]{%
        \includegraphics[width=0.45\linewidth]{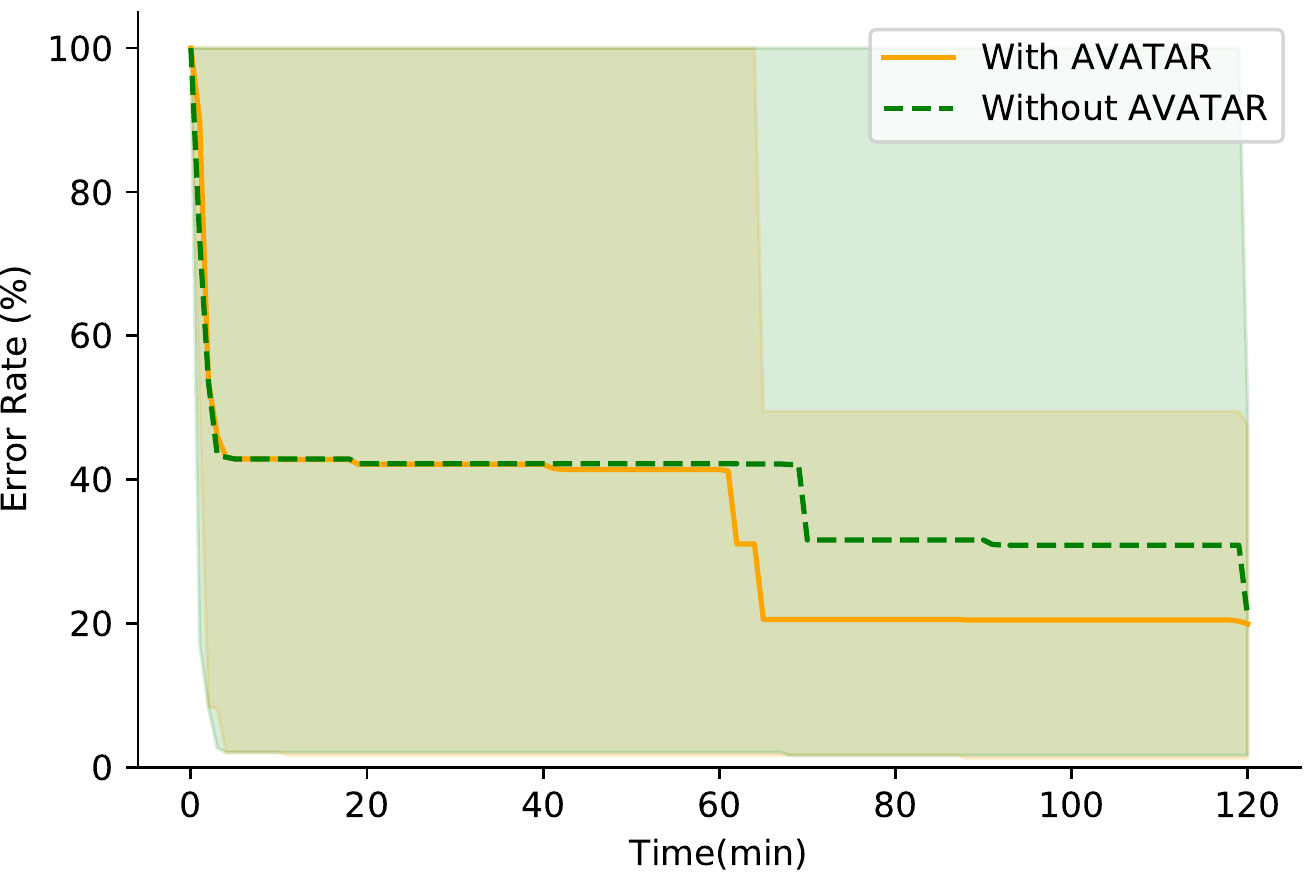}
      }%
    \hfill
    \subfloat[gcredit]{%
        \includegraphics[width=0.45\linewidth]{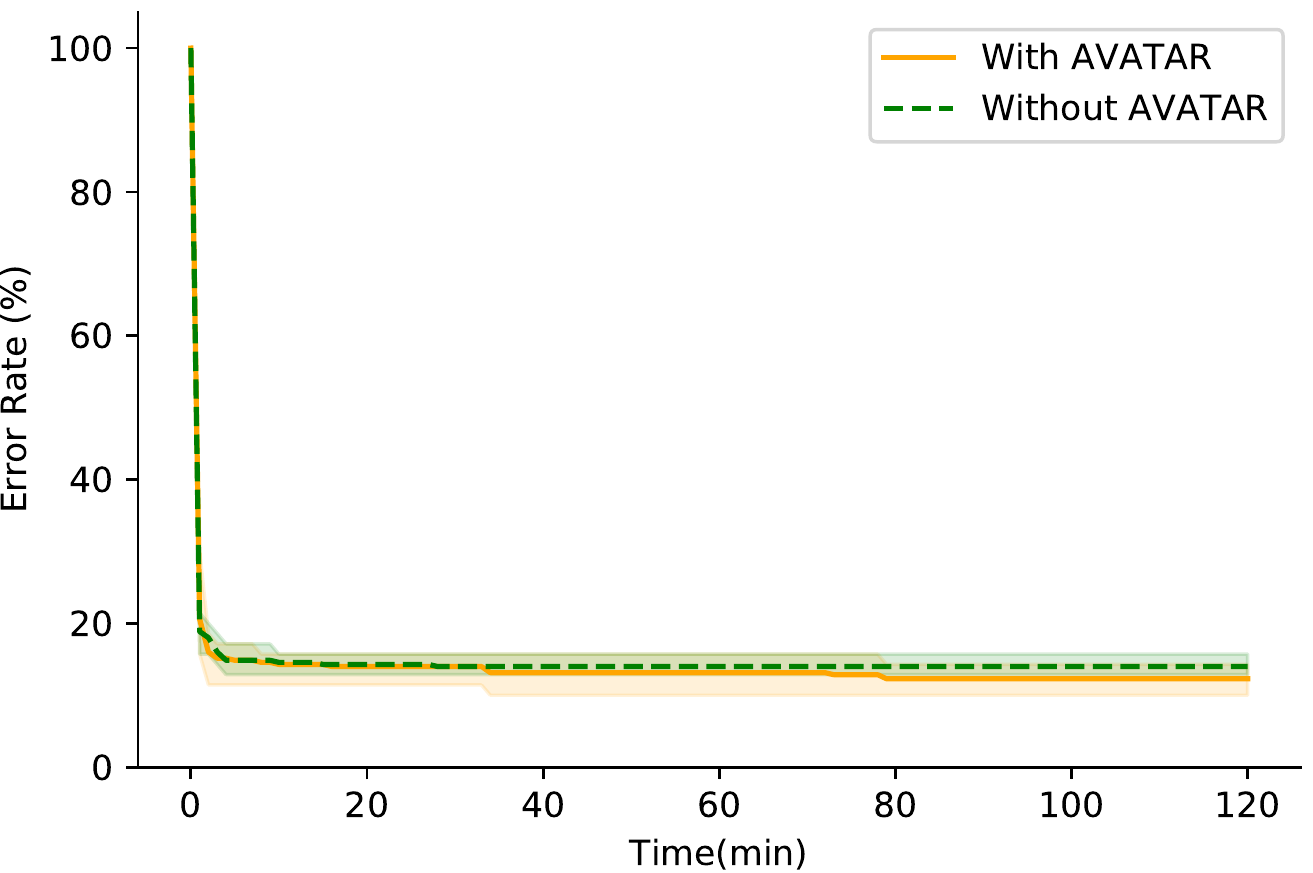}
        }%
    
    \subfloat[kddcup]{%
        \includegraphics[width=0.45\linewidth]{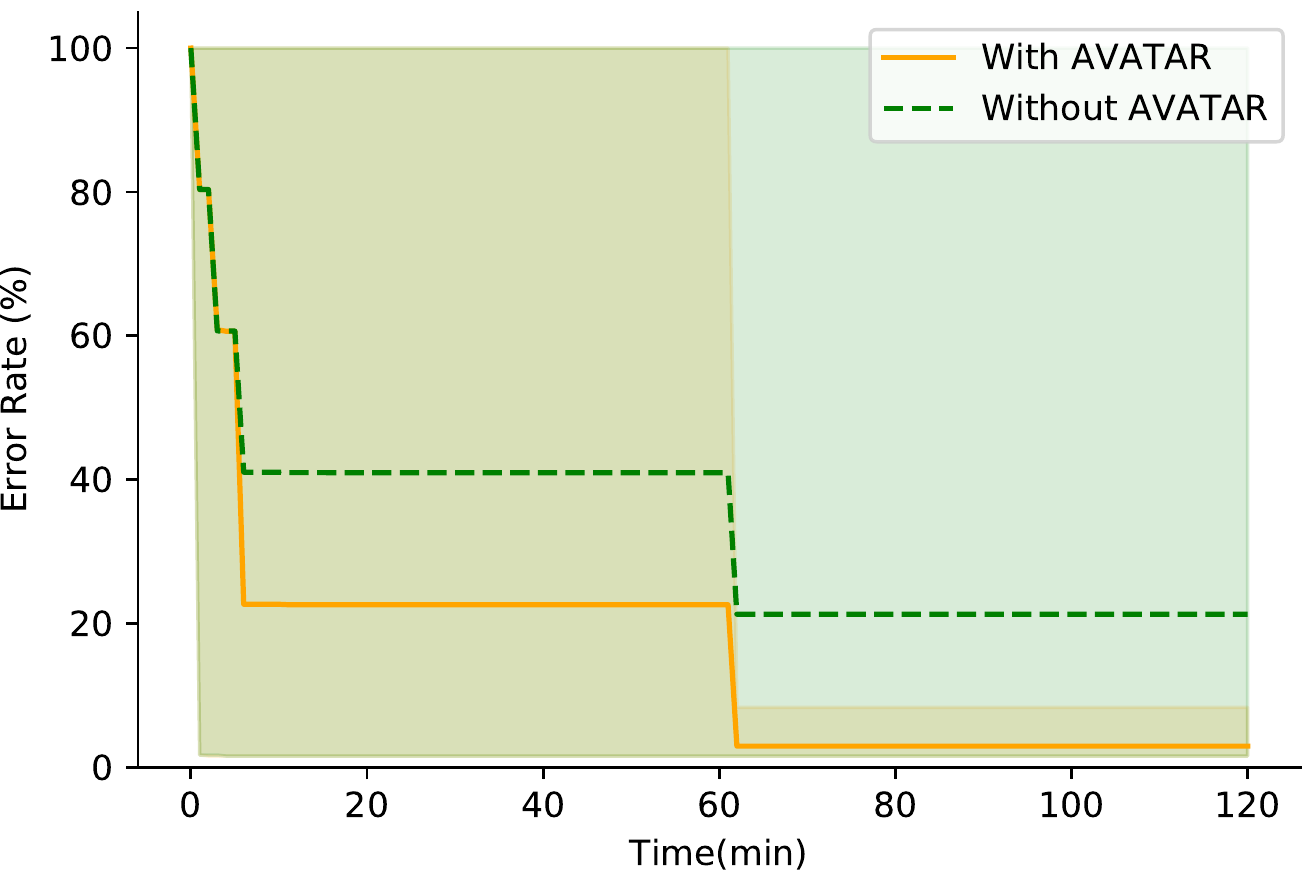}
        }%
     \hfill
     \subfloat[krvskp]{%
        \includegraphics[width=0.45\linewidth]{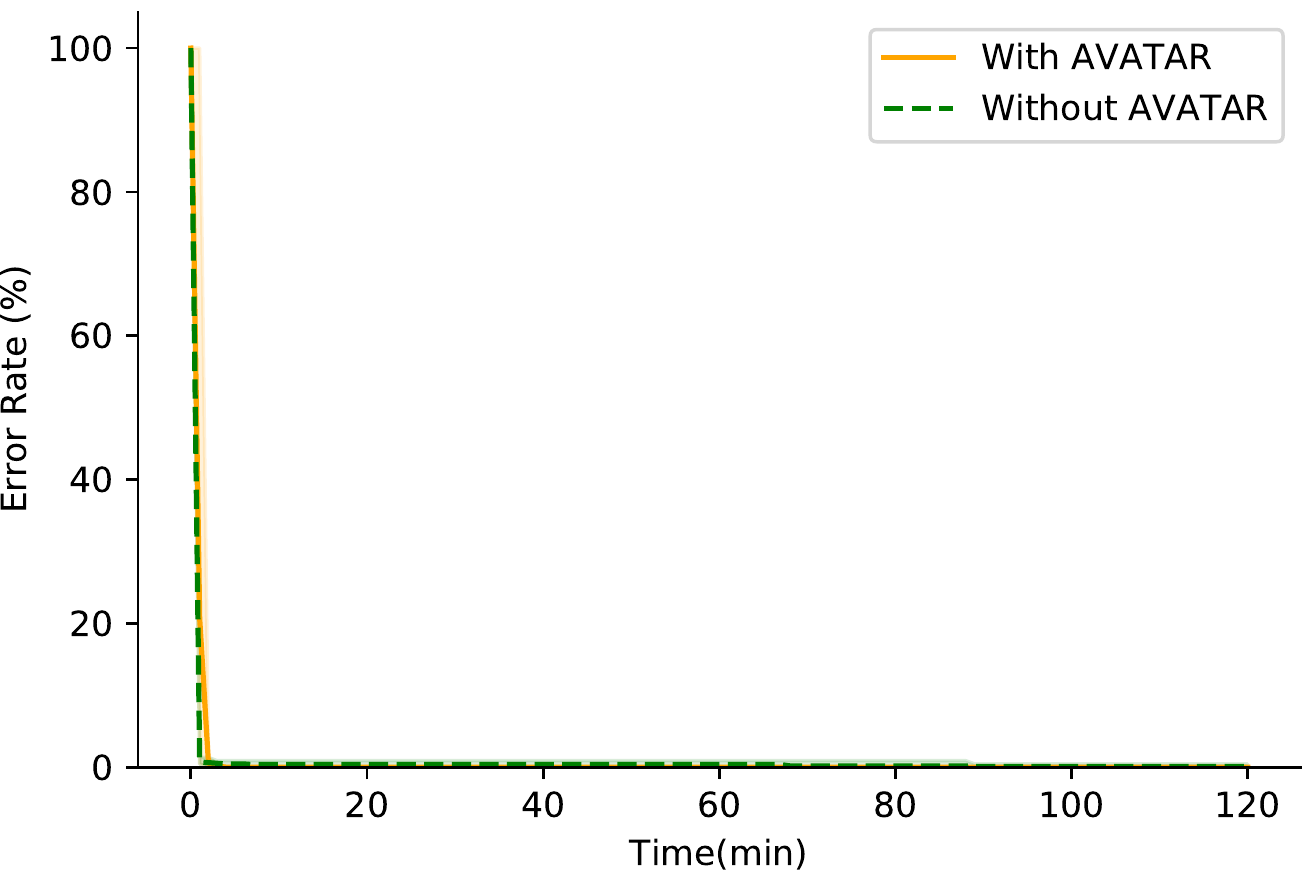}
      }%
    
    \subfloat[madelon]{%
        \includegraphics[width=0.45\linewidth]{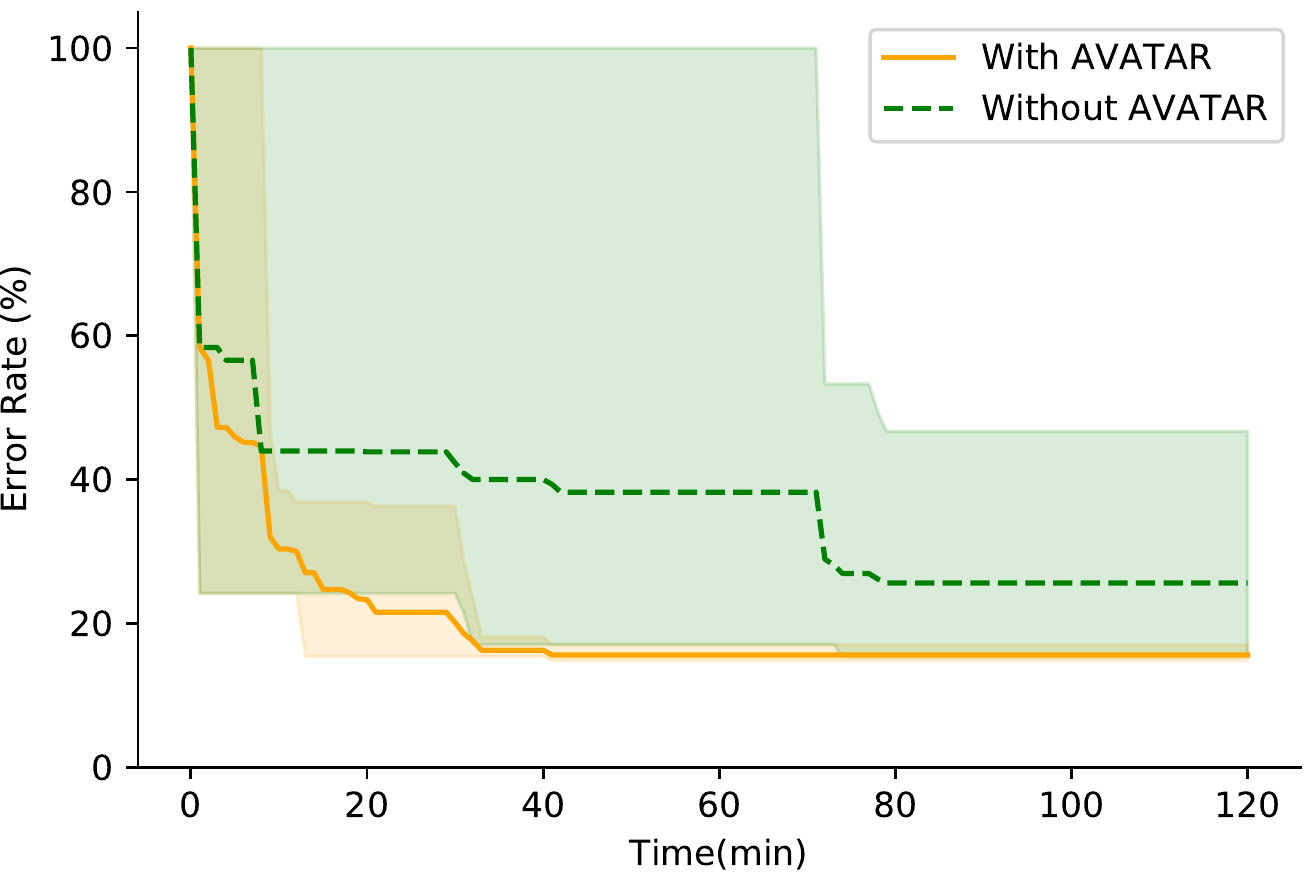}
        }%
     \hfill
     \subfloat[mnist]{%
        \includegraphics[width=0.45\linewidth]{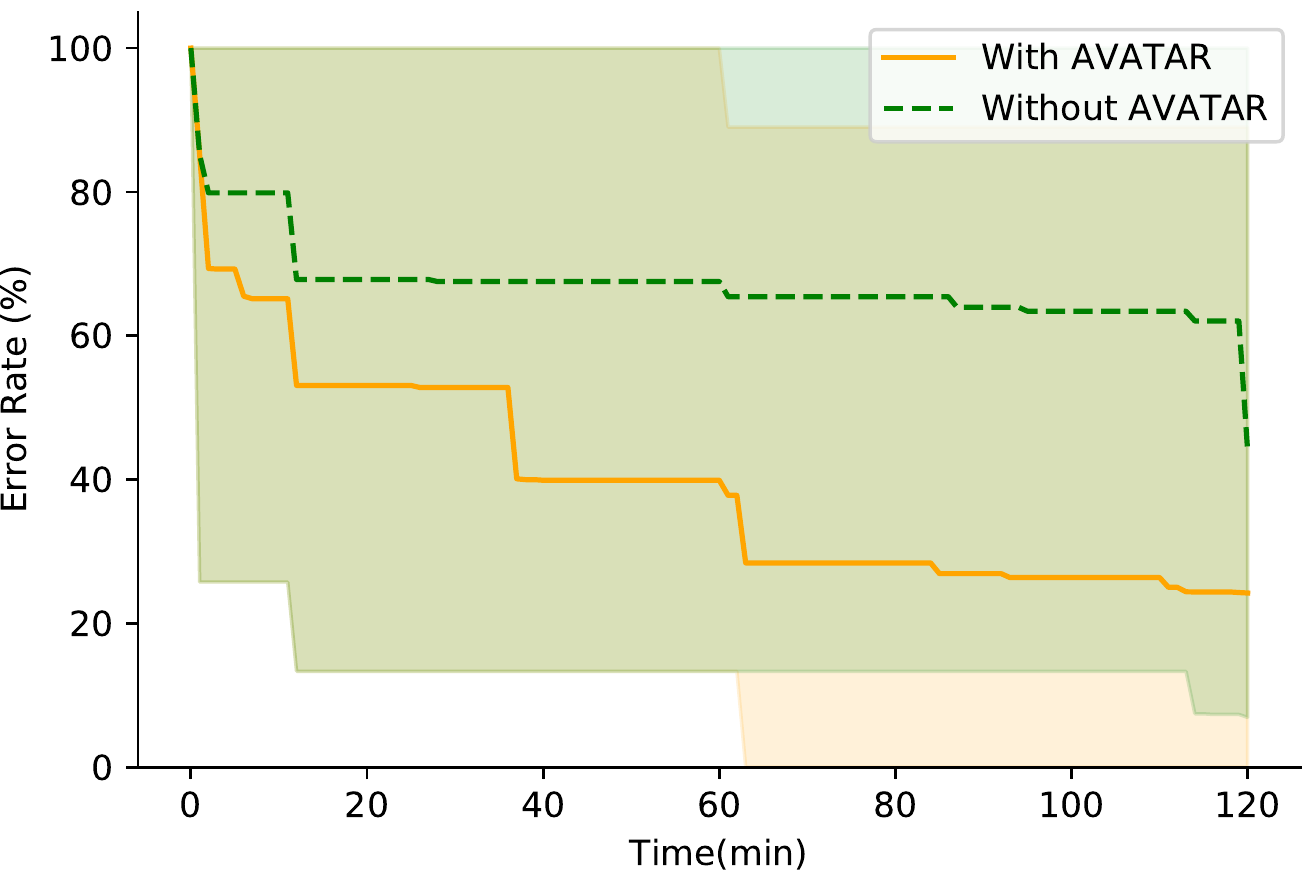}
      }%
    
    \subfloat[secom]{%
        \includegraphics[width=0.45\linewidth]{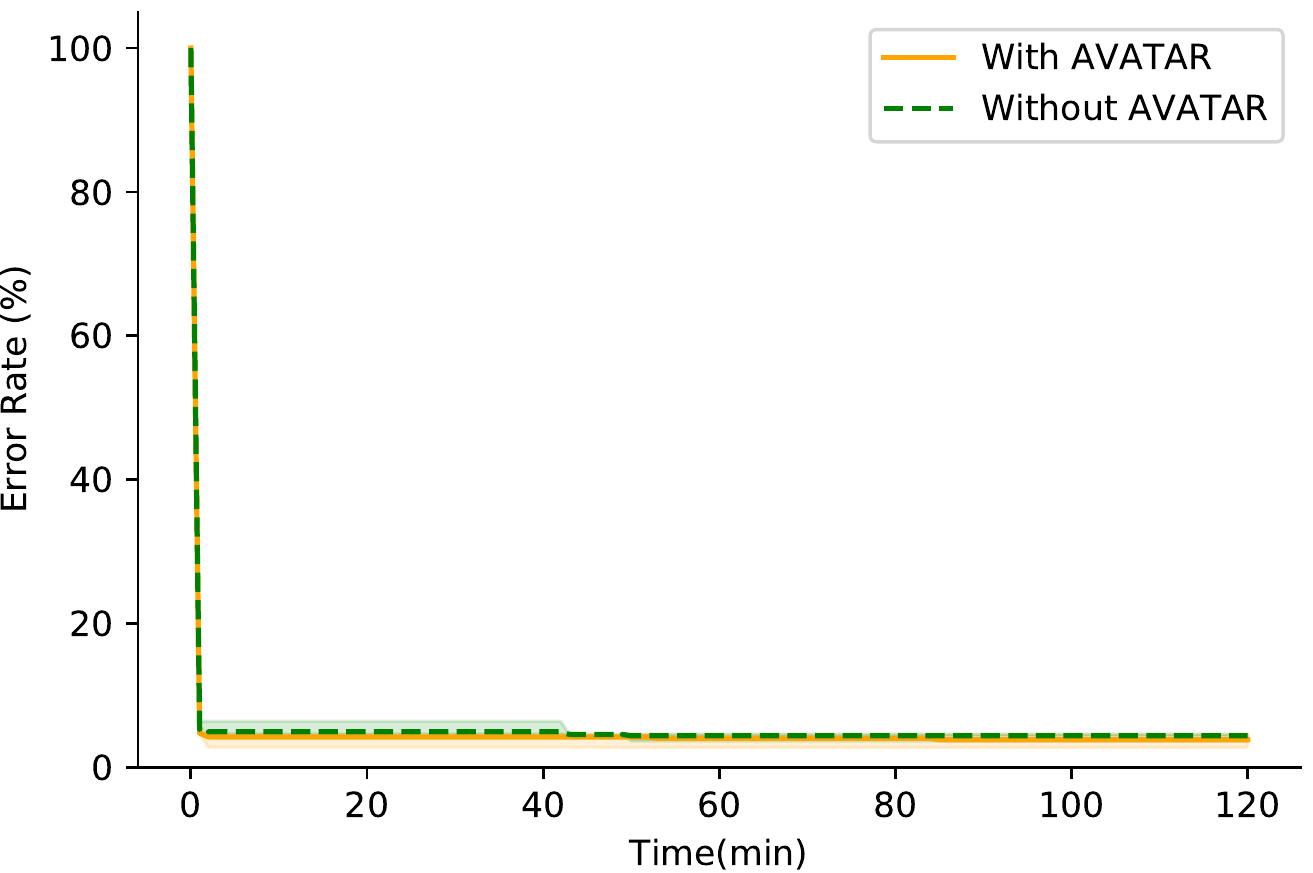}
        }%
     \hfill
     \subfloat[semeion]{%
        \includegraphics[width=0.45\linewidth]{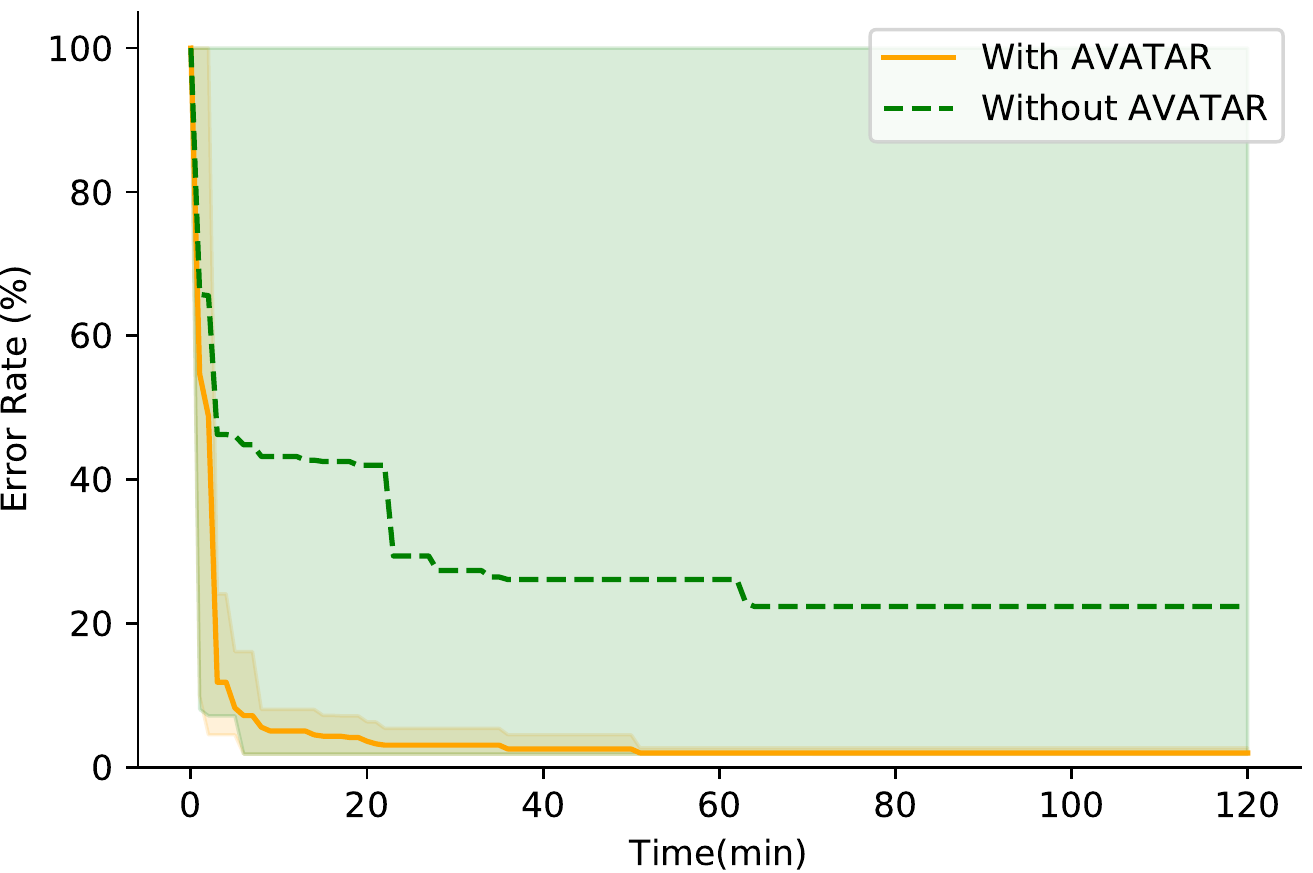}
      }%

 \caption{Comparison of the convergence of the error rate of the best pipelines found with and without the AVATAR.}
    \label{fig:exp_convergence_confidence_avatar_vs_noavatar}
\end{figure}

 \begin{figure}[!htbp]     
  \ContinuedFloat

    \subfloat[shuttle]{%
        \includegraphics[width=0.45\linewidth]{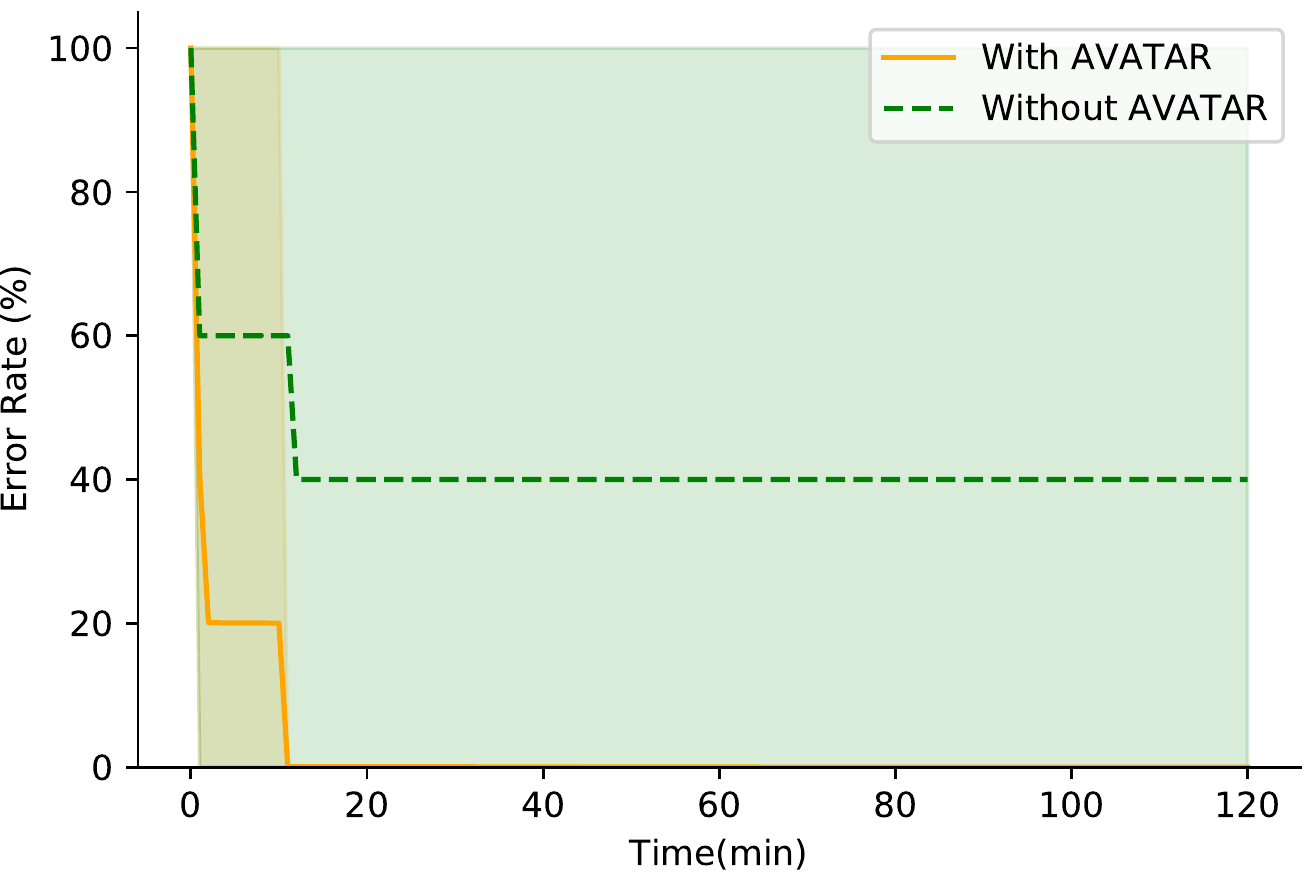}
        }%
    \hfill    
    \subfloat[waveform]{%
        \includegraphics[width=0.45\linewidth]{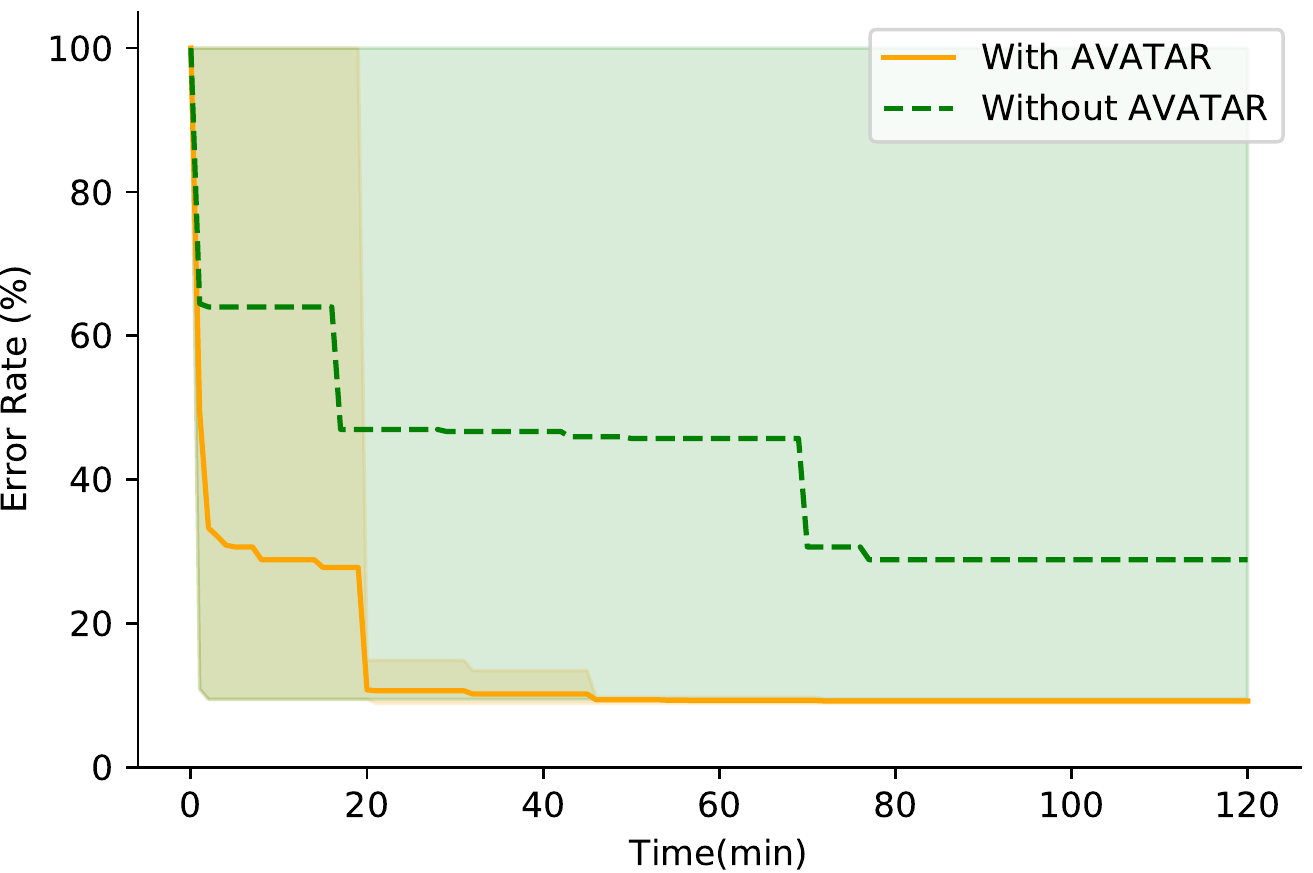}
      }%
    
    \subfloat[winequality]{%
        \includegraphics[width=0.45\linewidth]{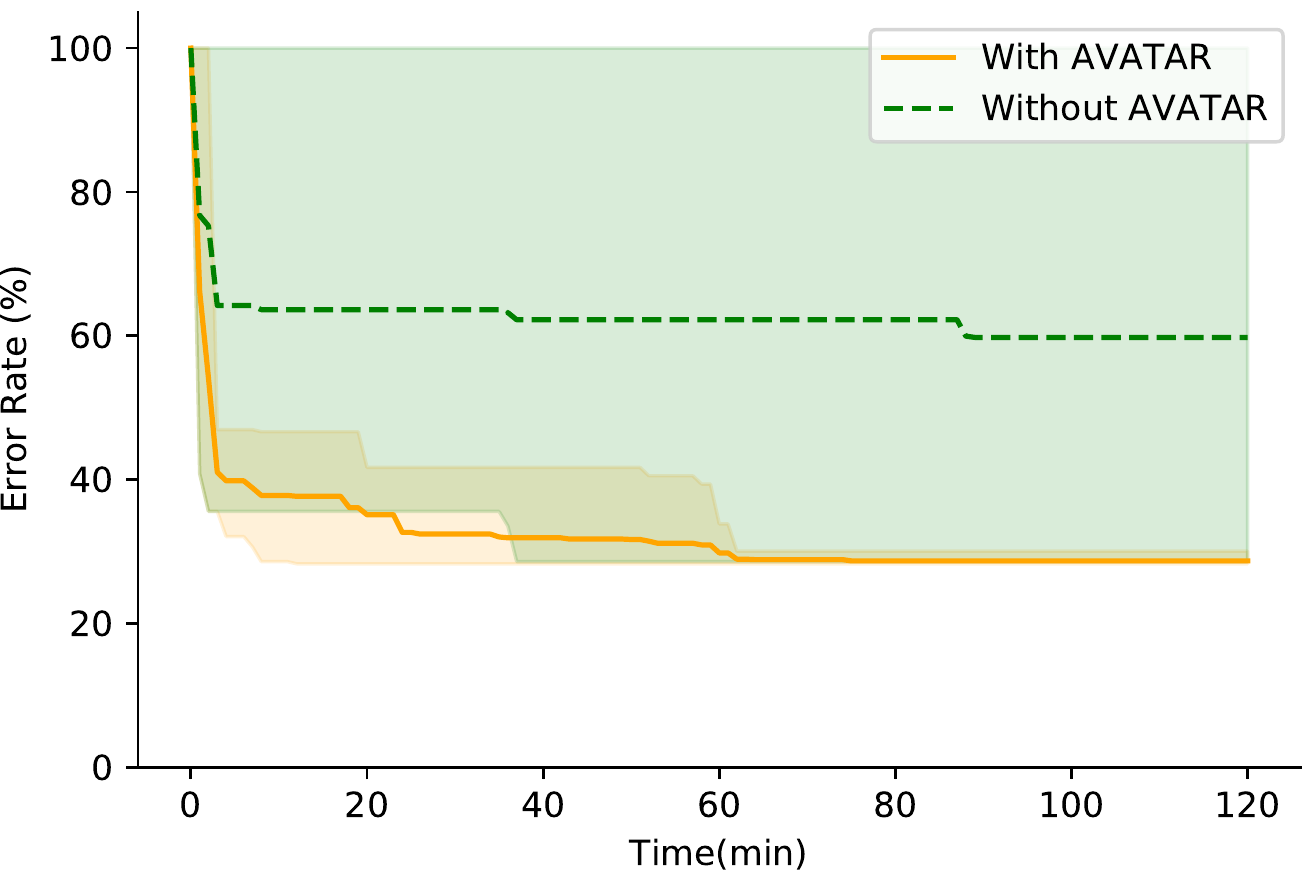}
        }%
        \hfill
       \subfloat[yeast]{%
        \includegraphics[width=0.45\linewidth]{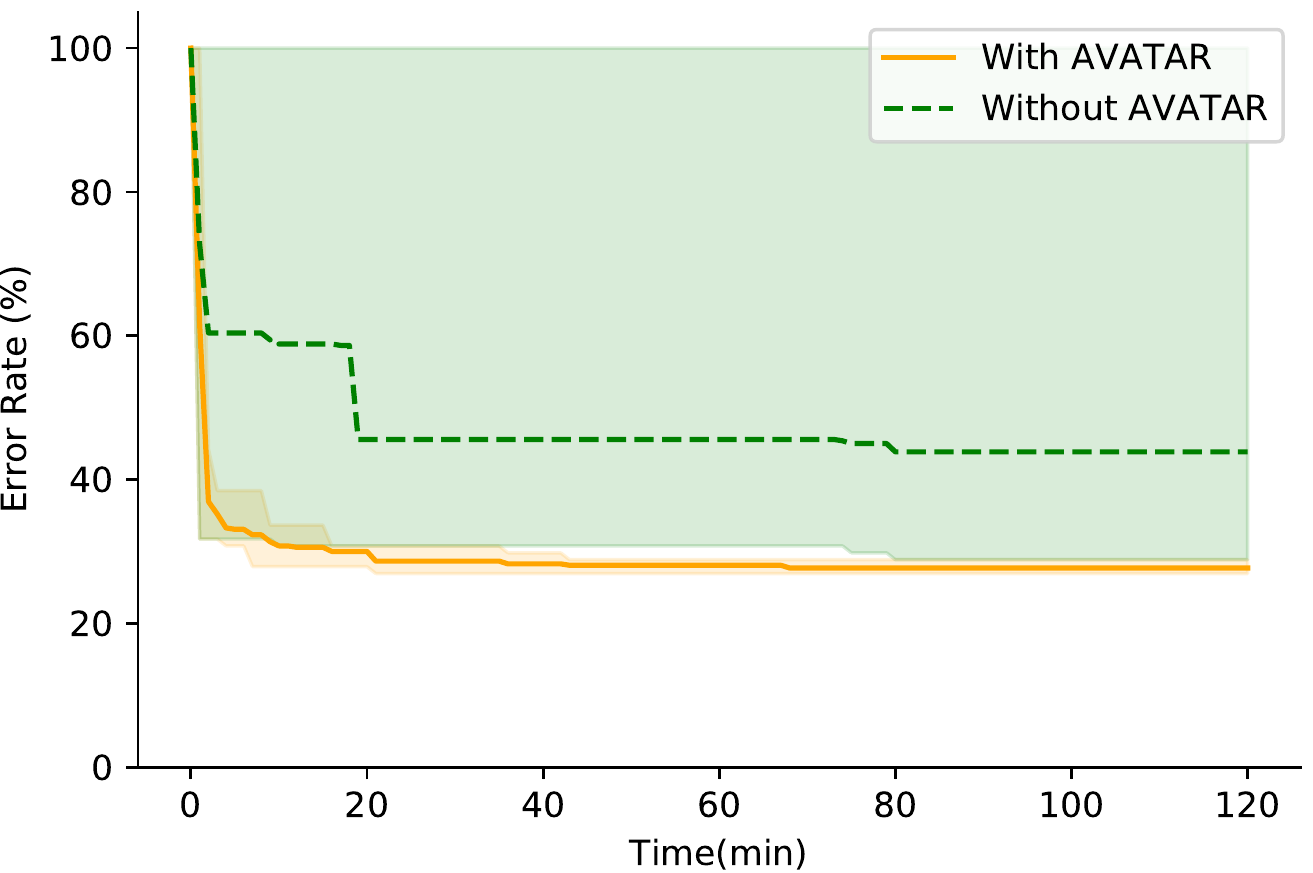}
      }%
        
    \caption{Compare the convergence of the error rate of the best pipelines found with and without the AVATAR.}
    \label{fig:exp_convergence_confidence_avatar_vs_noavatar}
\end{figure}

\begin{landscape}

\begin{table*}[htb!]

\vspace{-0.35cm}
\captionsetup{font=scriptsize}
\caption {Compare mean, minimum (min), difference of mean (dif\_mean error), difference of minimum (dif\_min error) and standard deviation (StdDev) of the error rate, and mean and standard deviation of the number of evaluated pipelines of SMAC integrated the AVATAR with 1 (SMAC1) and 5 (SMAC5) random initialising configurations. The better results are in bold.}
\label{tab:exp_smac1_vs_smac5}

\centering
\scriptsize

\begin{tabular}{|l|l|l|l|l|l|l|l|l|l|l|l|l|}
\hline
\multirow{2}{*}{\textbf{Dataset}} & \multicolumn{2}{l|}{\textbf{Mean error (\%)}} & \multicolumn{2}{l|}{\textbf{Min error (\%)}} & \multirow{2}{*}{\textbf{Dif\_mean error}} & \multirow{2}{*}{\textbf{Dif\_min error}} & \multicolumn{2}{l|}{\textbf{StdDev error}} & \multicolumn{2}{l|}{\textbf{Mean pipelines}} & \multicolumn{2}{l|}{\textbf{StdDev pipeline}} \\ \cline{2-5} \cline{8-13} 
                                  & \textbf{SMAC1}        & \textbf{SMAC5}        & \textbf{SMAC1}        & \textbf{SMAC5}       &                                           &                                          & \textbf{SMAC1}         & \textbf{SMAC5}         & \textbf{SMAC1}        & \textbf{SMAC5}       & \textbf{SMAC1}        & \textbf{SMAC5}        \\ \hline
\textbf{abalone}                  & 73.86                 & \textbf{73.44}        & \textbf{72.91}        & 73.26                & 0.42                                      & 0.35                                     & 0.94                   & \textbf{0.16}          & 106                   & 161                  & 73                    & 93                    \\ \hline
\textbf{adult}                  & 14.60                 & \textbf{14.27}        & \textbf{13.41}        & 13.77                & 0.33                                      & 0.36                                     & 0.60                   & \textbf{0.46}          & 60                   & 56                  & 28                    & 9                    \\ \hline
\textbf{amazon}                   & \textbf{46.03}        & 46.57                 & \textbf{38.48}        & \textbf{38.48}       & 0.54                                      & 0                                        & 10.68                  & \textbf{8.10}          & 44                    & 28                   & 38                    & 11                    \\ \hline
\textbf{car}                      & 0.58                  & \textbf{0.38}         & \textbf{0.33}         & \textbf{0.33}        & 0.2                                       & 0                                        & 0.30                   & \textbf{0.04}          & 1136                  & 1288                 & 122                   & 83                    \\ \hline
\textbf{cifar10small}             & 71.94                 & \textbf{70.72}        & \textbf{64.94}        & 70.45                & 1.22                                      &  5.51                               & 8.76                   & \textbf{0.27}          & 35                    & 19                   & 24                    & 10                    \\ \hline
\textbf{convex}                   & 33.52                 & \textbf{33.39}        & \textbf{25.69}        & 26.78                & 0.13                                      & 1.09                                     & 5.74                   & \textbf{4.56}          & 49                    & 44                   & 30                    & 15                    \\ \hline
\textbf{dexter}                   & \textbf{8.09}         & 10                    & \textbf{6.9}          & 7.86                 & 1.91                                      & 0.96                                     & \textbf{0.85}          & 3.34                   & 33                    & 75                   & 23                    & 26                    \\ \hline
\textbf{dorothea}                 & 8.02                  & \textbf{7.21}         & \textbf{6.71}         & 7.21                 & 0.81                                      & 0.5                                      & 1.31                   & \textbf{0.00}          & 17                    & 20                   & 14                    & 9                     \\ \hline
\textbf{gcredit}                  & \textbf{22.26}        & 22.37                 & \textbf{21.71}        & 22                   & 0.11                                      & 0.29                                     & 0.39                   & \textbf{0.23}          & 808                   & 948                  & 270                   & 331                   \\ \hline
\textbf{gisette}                  & \textbf{2.56}         & 3.66                  & 2.39                  & \textbf{2.04}        & 1.1                                       & 0.35                                     & \textbf{0.19}          & 2.05                   & 27                    & 35                   & 20                    & 18                    \\ \hline
\textbf{kddcup}        & \textbf{1.80}         & 1.84                  & 1.80                  & \textbf{1.78}        & 0.04                                      & 0.02                                     & \textbf{0.00}          & 0.06                   & 67                    & 35                   & 33                    & 16                    \\ \hline
\textbf{krvskp}                   & \textbf{0.44}         & 0.49                  & \textbf{0.40}         & \textbf{0.40}        & 0.05                                      & 0                                        & 0.04                   & \textbf{0.08}          & 898                   & 798                  & 128                   & 87                    \\ \hline
\textbf{madelon}                  & \textbf{22.86}        & 23.88                 & 22.75                 & \textbf{22.64}       & 1.02                                      & 0.11                                     & \textbf{0.22}          & 1.24                   & 279                   & 200                  & 59                    & 64                    \\ \hline
\textbf{mnist}                    & 16.38                 & \textbf{9.17}         & 16.38                 & \textbf{9.17}        &  7.21                               &  7.21                               & \textbf{0.00}          & \textbf{0.00}          & 23                    & 15                   & 27                    & 10                    \\ \hline
\textbf{secom}                    & \textbf{6.11}         & \textbf{6.11}         & \textbf{6.11}         & \textbf{6.11}        & 0                                         & 0                                        & \textbf{0.00}          & \textbf{0.00}          & 277                   & 339                  & 160                   & 145                   \\ \hline
\textbf{semeion}                  & \textbf{4.95}         & 5.32                  & \textbf{4.66}         & 4.84                 & 0.37                                      & 0.18                                     & \textbf{0.41}          & 0.57                   & 338                   & 328                  & 118                   & 88                    \\ \hline
\textbf{shuttle}                  & 0.09                  & \textbf{0.02}         & 0.03                  & \textbf{0.02}        & 0.07                                      & 0.005                                    & 0.10                   & \textbf{0.00}          & 91                    & 140                  & 48                    & 21                    \\ \hline
\textbf{waveform}                 & 12.53                 & \textbf{12.49}        & 12.46                 & \textbf{12.43}       & 0.04                                      & 0.03                                     & 0.13                   & \textbf{0.07}          & 333                   & 357                  & 138                   & 88                    \\ \hline
\textbf{winequality}              & \textbf{33.44}        & 33.61                 & 33.10                 & \textbf{33.04}       & 0.17                                      & 0.06                                     & \textbf{0.38}          & 0.89                   & 432                   & 428                  & 208                   & 148                   \\ \hline
\textbf{yeast}                    & \textbf{38.02}        & 38.09                 & 36.96                 & 37.25                & 0.07                                      & 0.29                                     & 0.60                   & \textbf{0.56}          & 1116                  & 1171                 & 349                   & 103                   \\ \hline
\end{tabular}

\end{table*}

\end{landscape}

\begin{figure}[!htbp] 
  
    \subfloat[abalone]{%
        \includegraphics[width=0.45\linewidth]{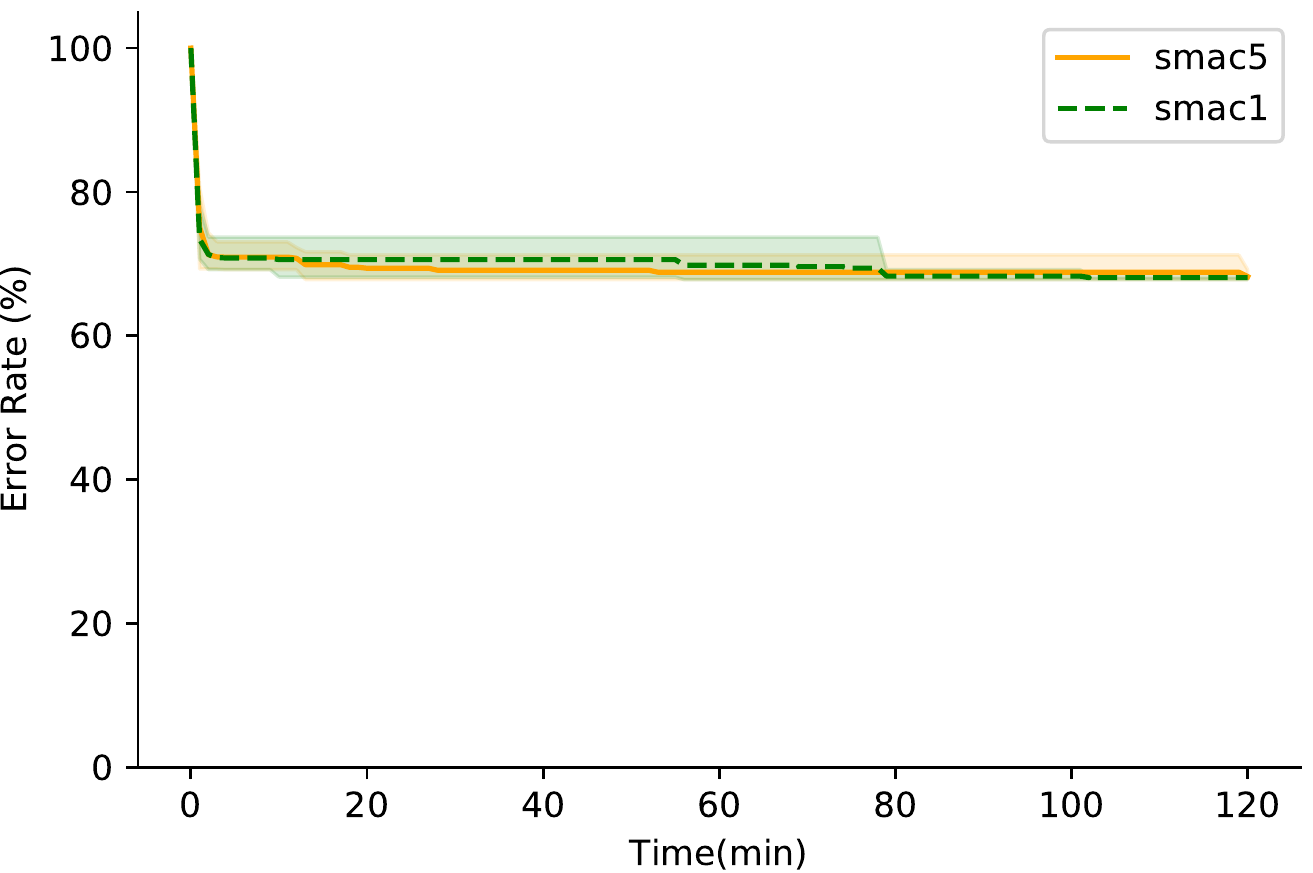}
      }%
    \hfill
    \subfloat[adult]{%
        \includegraphics[width=0.45\linewidth]{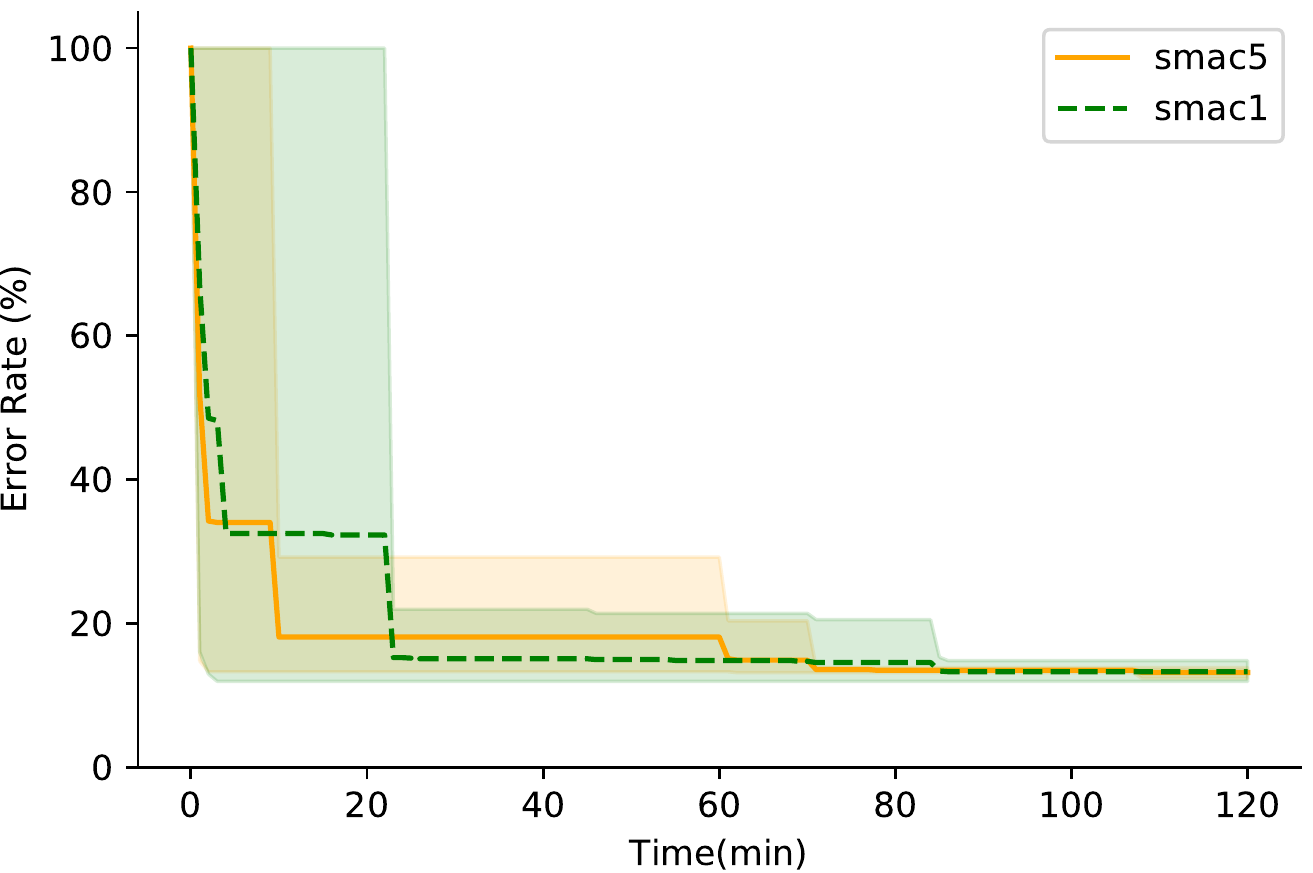}
      }%
      
    \subfloat[amazon]{%
        \includegraphics[width=0.45\linewidth]{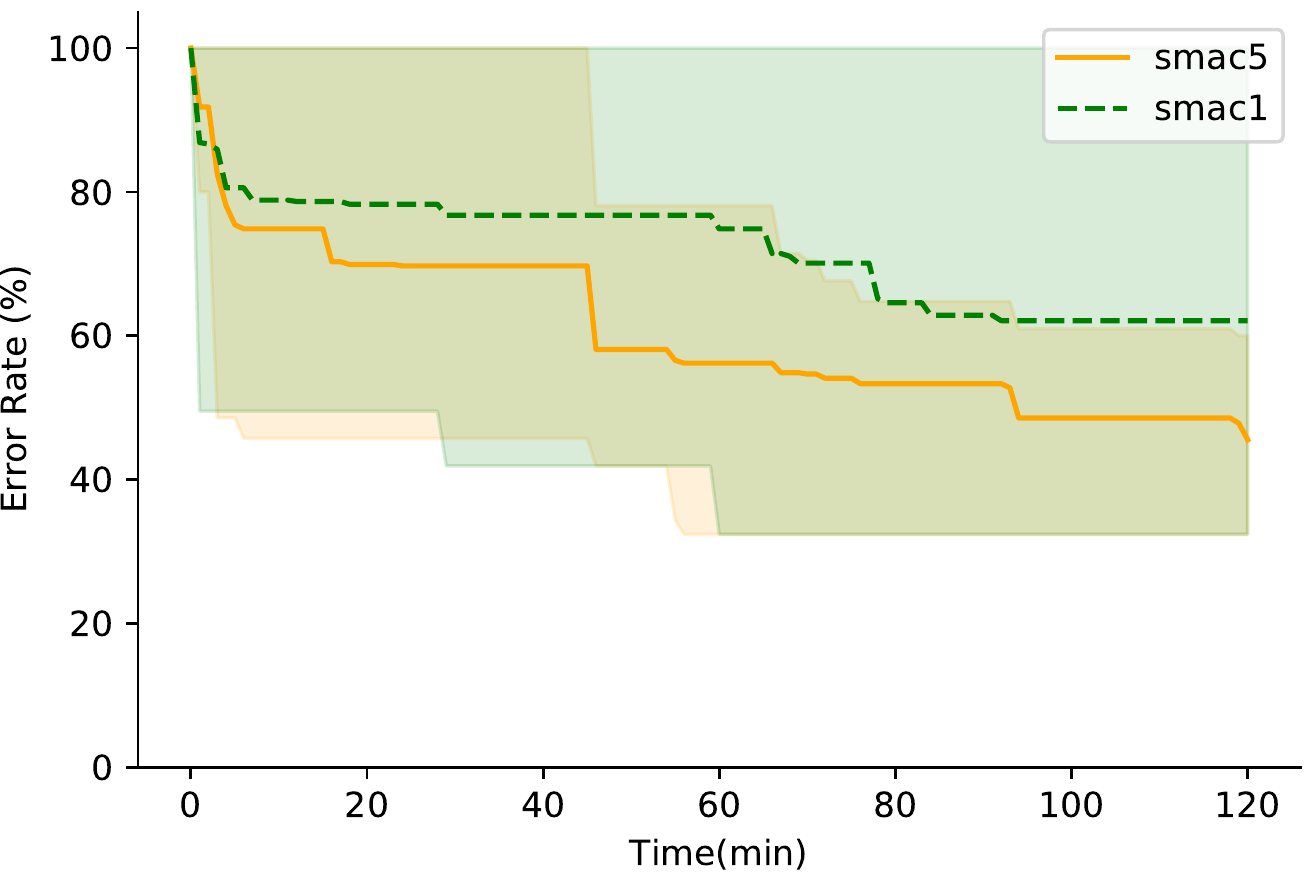}
        }%
    \hfill
    \subfloat[car]{%
        \includegraphics[width=0.45\linewidth]{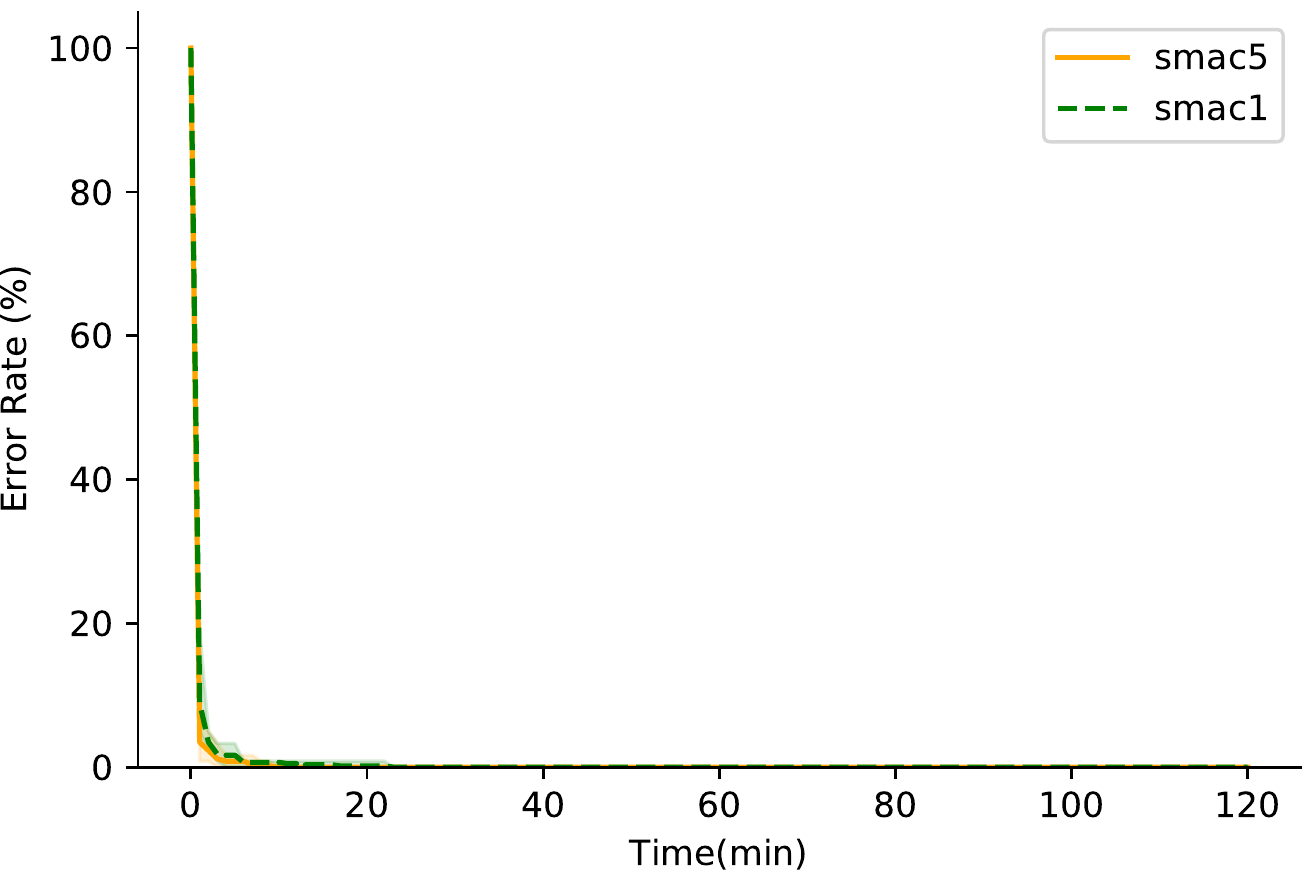}
      }%
    
    \subfloat[cifar10small]{%
        \includegraphics[width=0.45\linewidth]{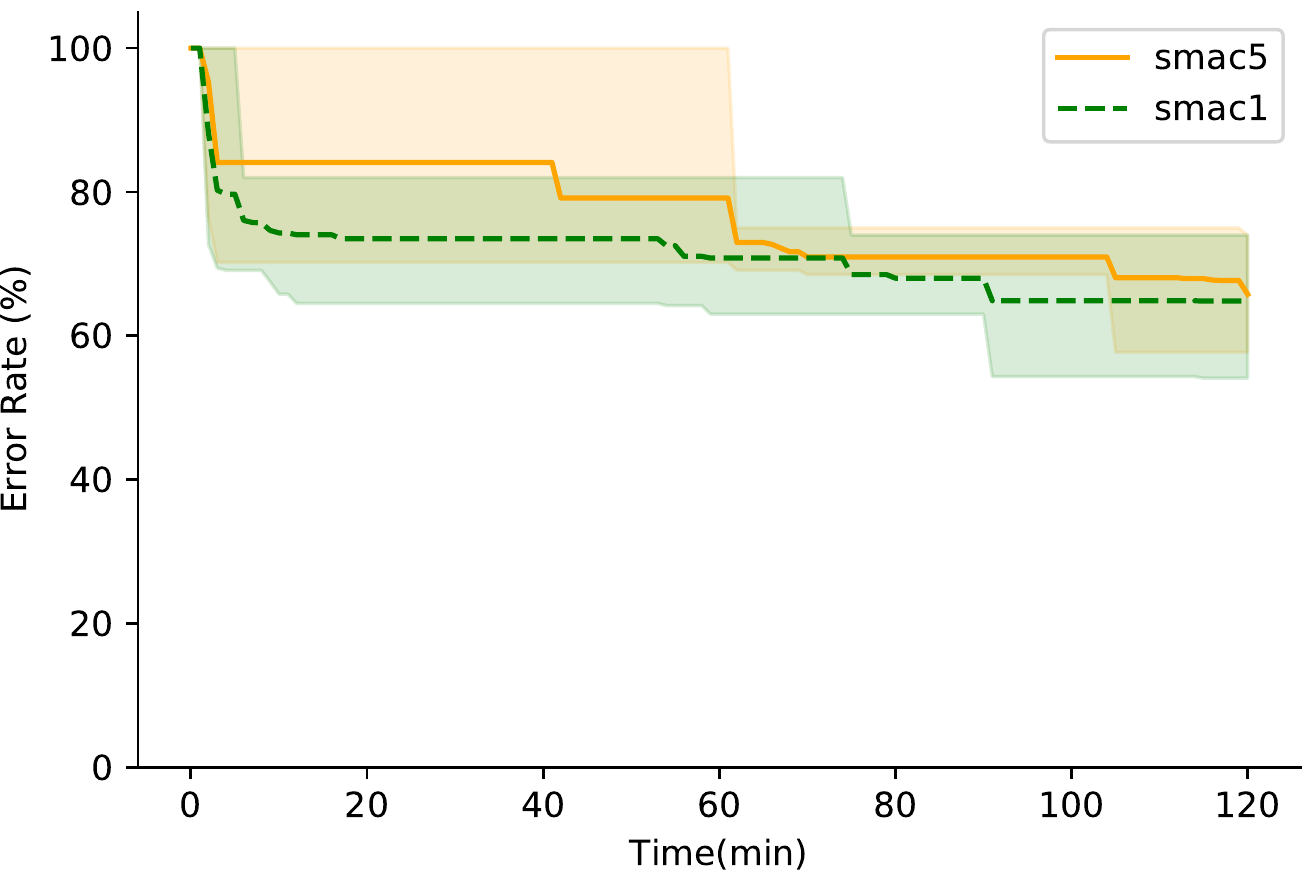}
        }%
    \hfill    
    \subfloat[convex]{%
        \includegraphics[width=0.45\linewidth]{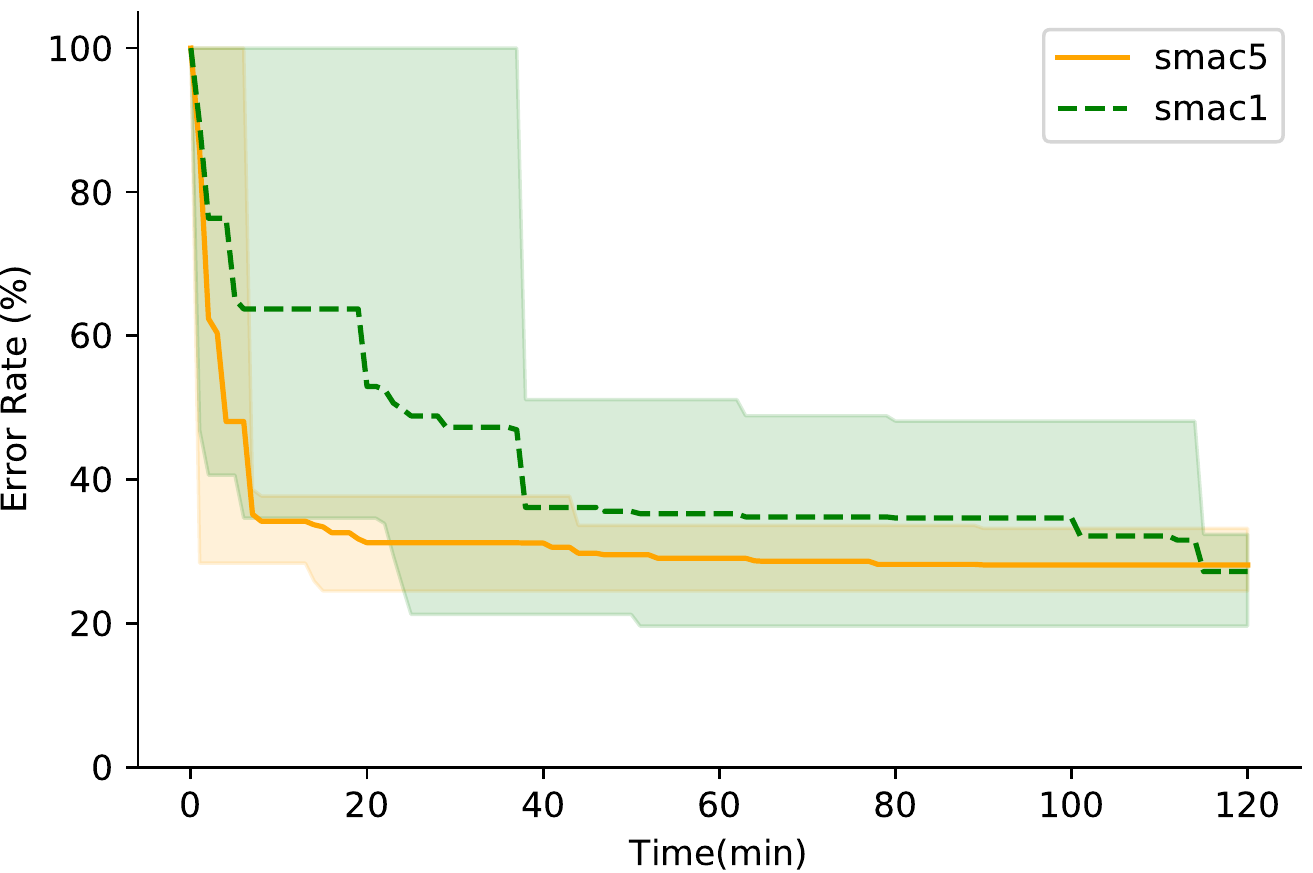}
      }%
    
    \subfloat[dexter]{%
        \includegraphics[width=0.45\linewidth]{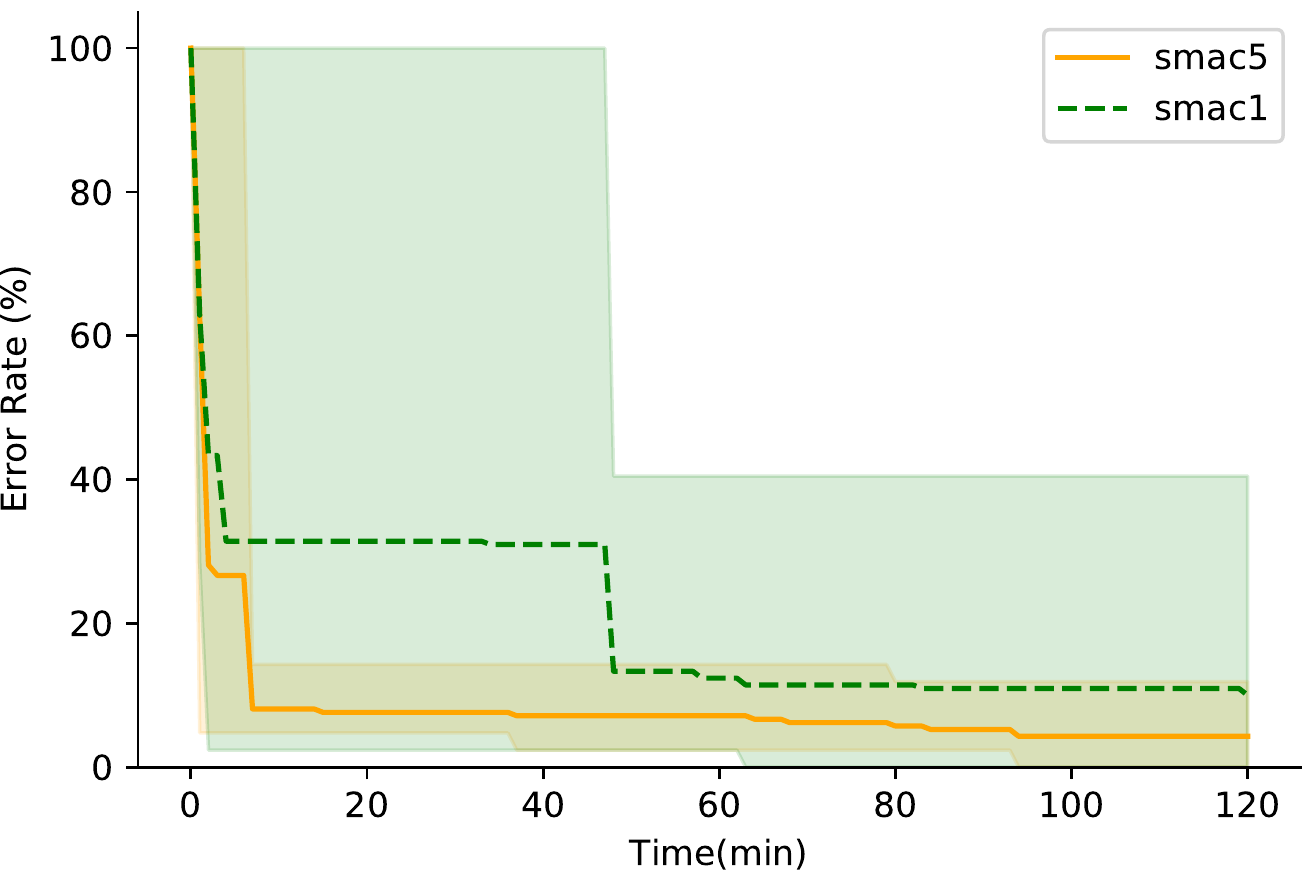}
        }%
     \hfill   
     \subfloat[dorothea]{%
        \includegraphics[width=0.45\linewidth]{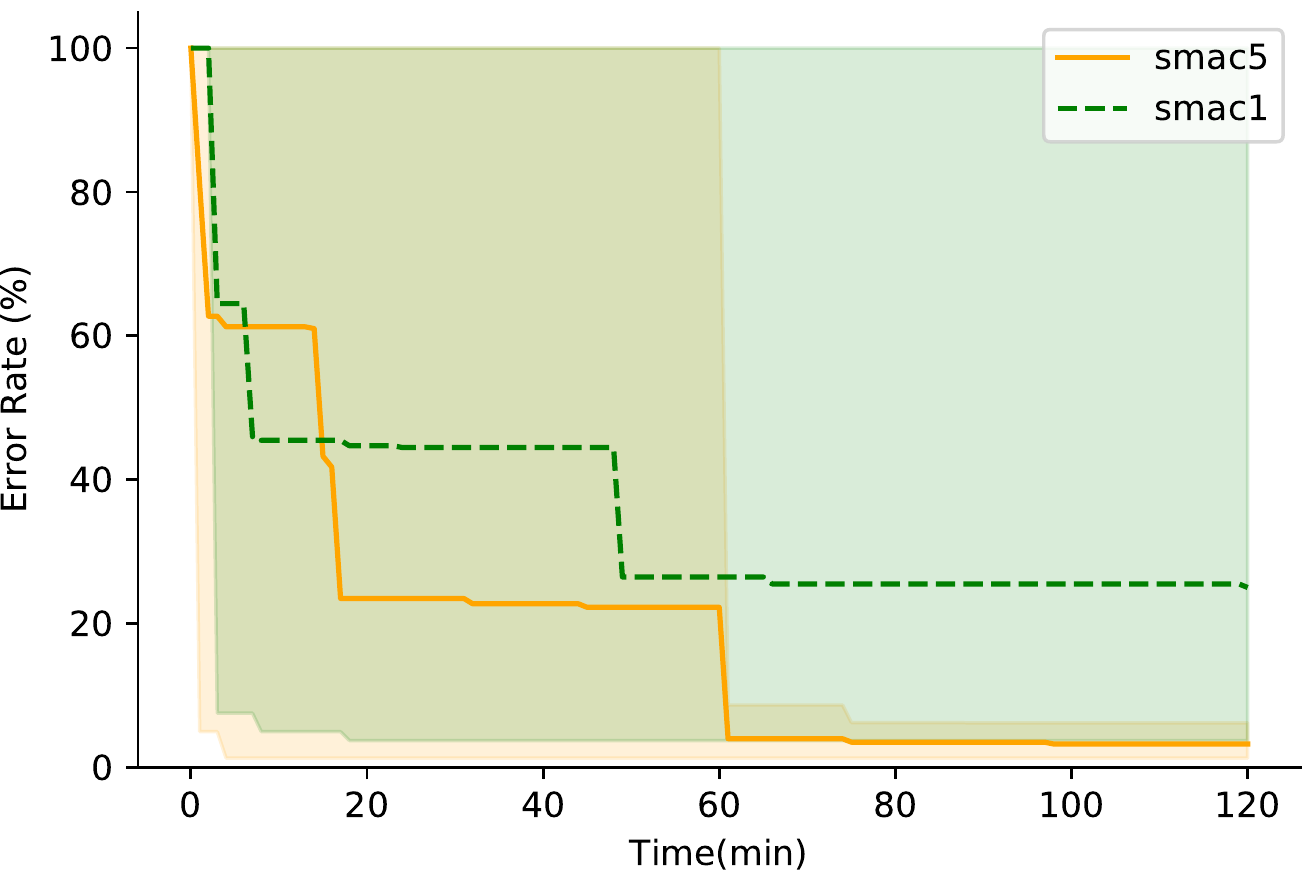}
      }%
      
       \caption{Compare the convergence of the error rate of the best pipelines found by SMAC1 and SMAC5.}
    \label{fig:exp_convergence_smac5_vs_smac1}
\end{figure}

 \begin{figure}[!htbp]     
  \ContinuedFloat    
     \subfloat[gisette]{%
        \includegraphics[width=0.45\linewidth]{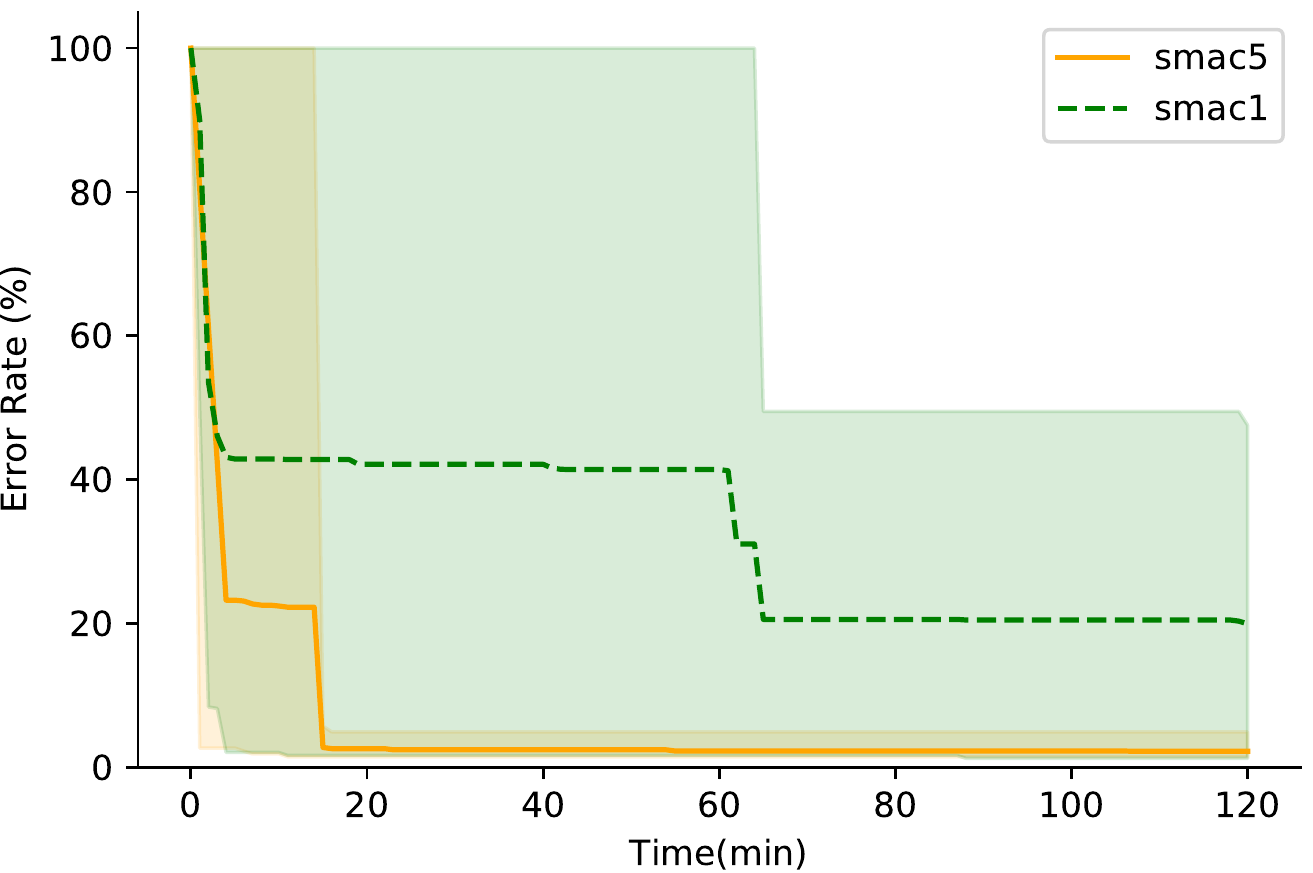}
      }%
    \hfill
    \subfloat[gcredit]{%
        \includegraphics[width=0.45\linewidth]{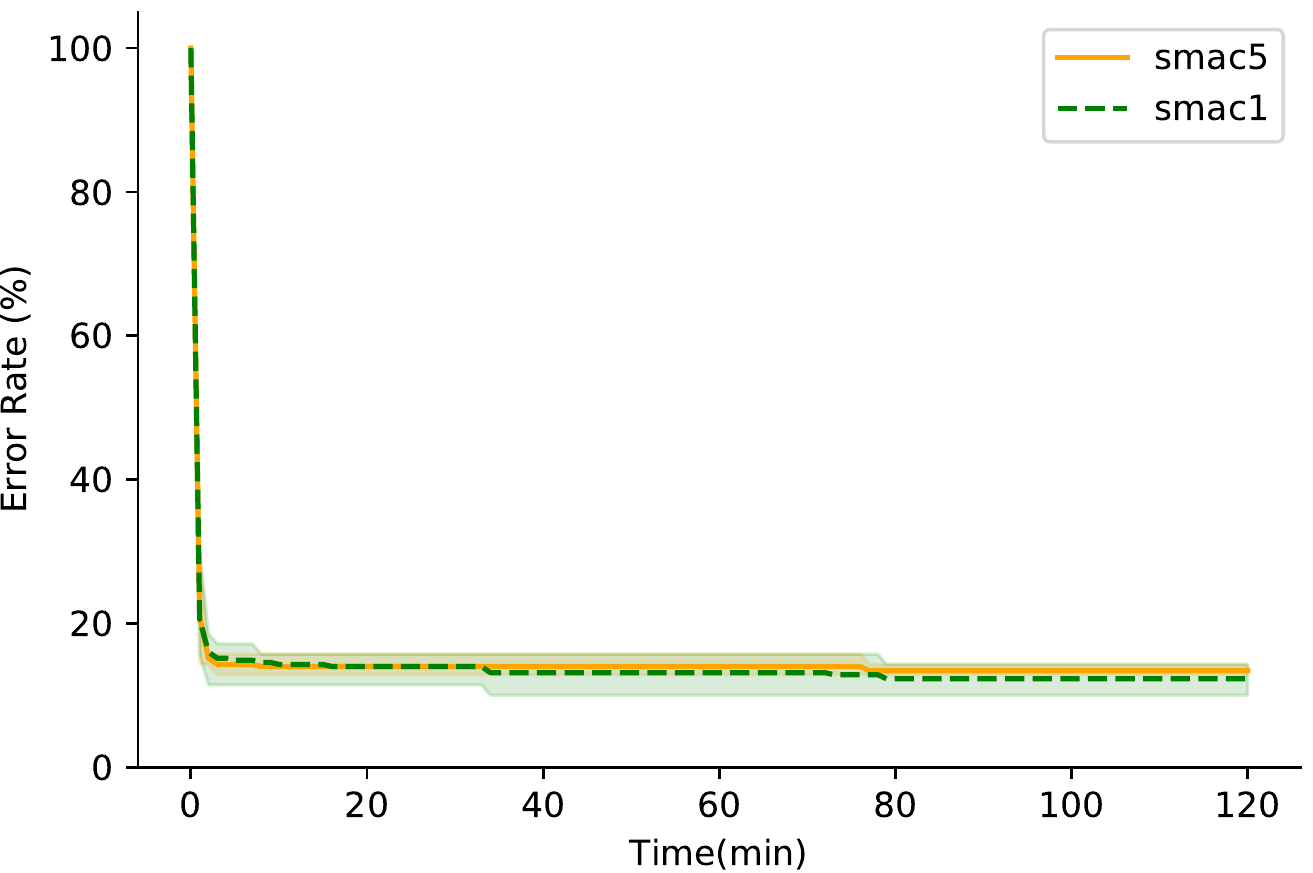}
        }%
    
    \subfloat[kddcup]{%
        \includegraphics[width=0.45\linewidth]{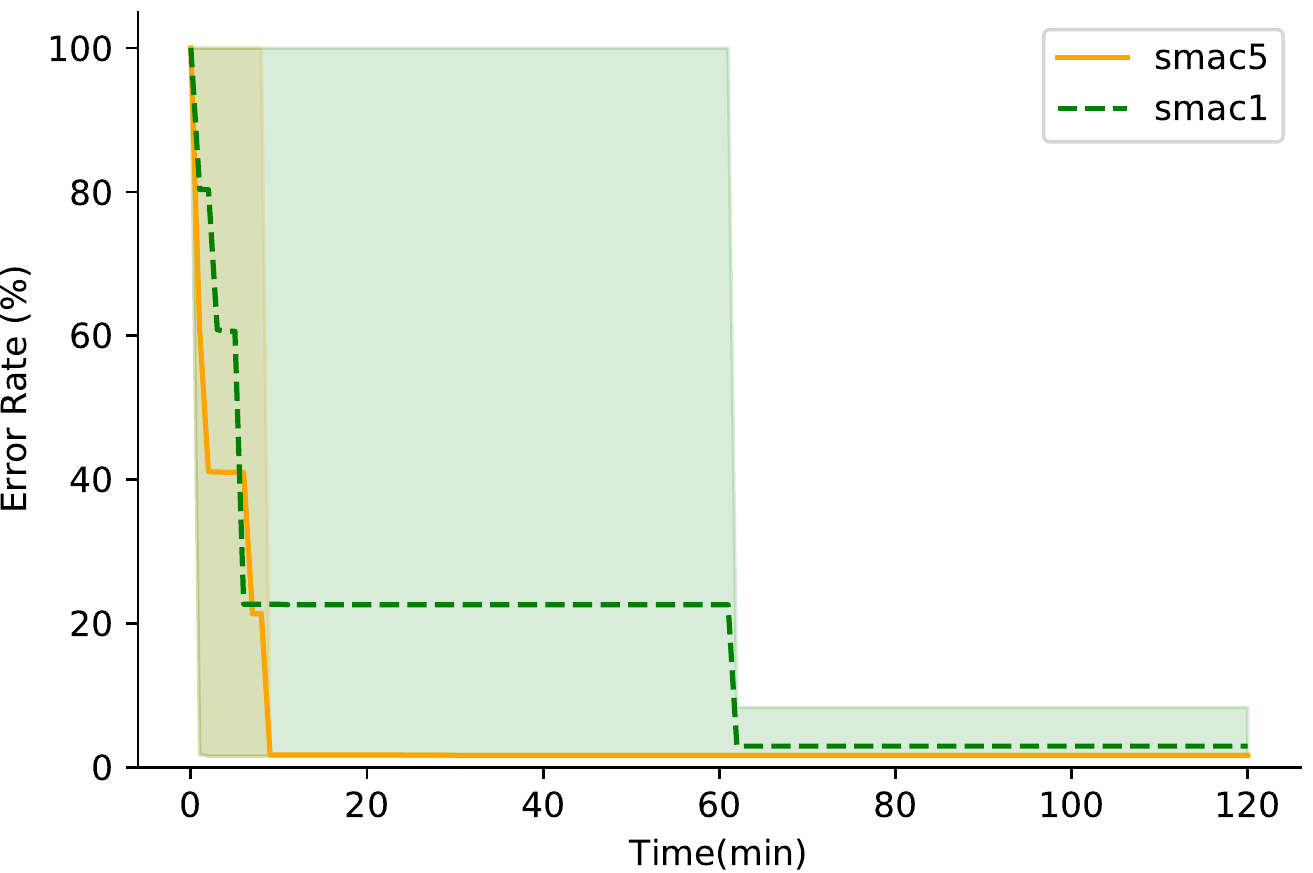}
        }%
     \hfill
     \subfloat[krvskp]{%
        \includegraphics[width=0.45\linewidth]{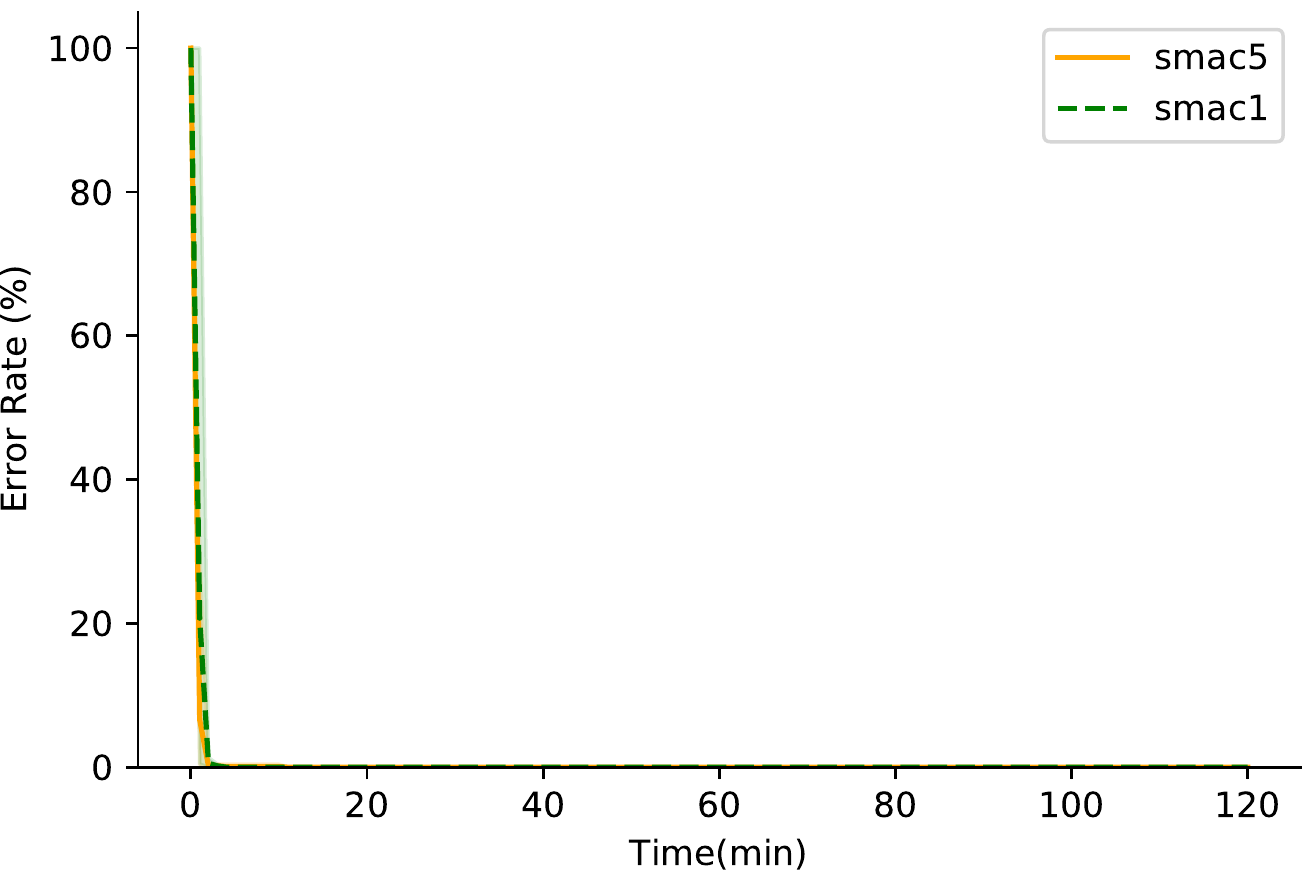}
      }%
    
    \subfloat[madelon]{%
        \includegraphics[width=0.45\linewidth]{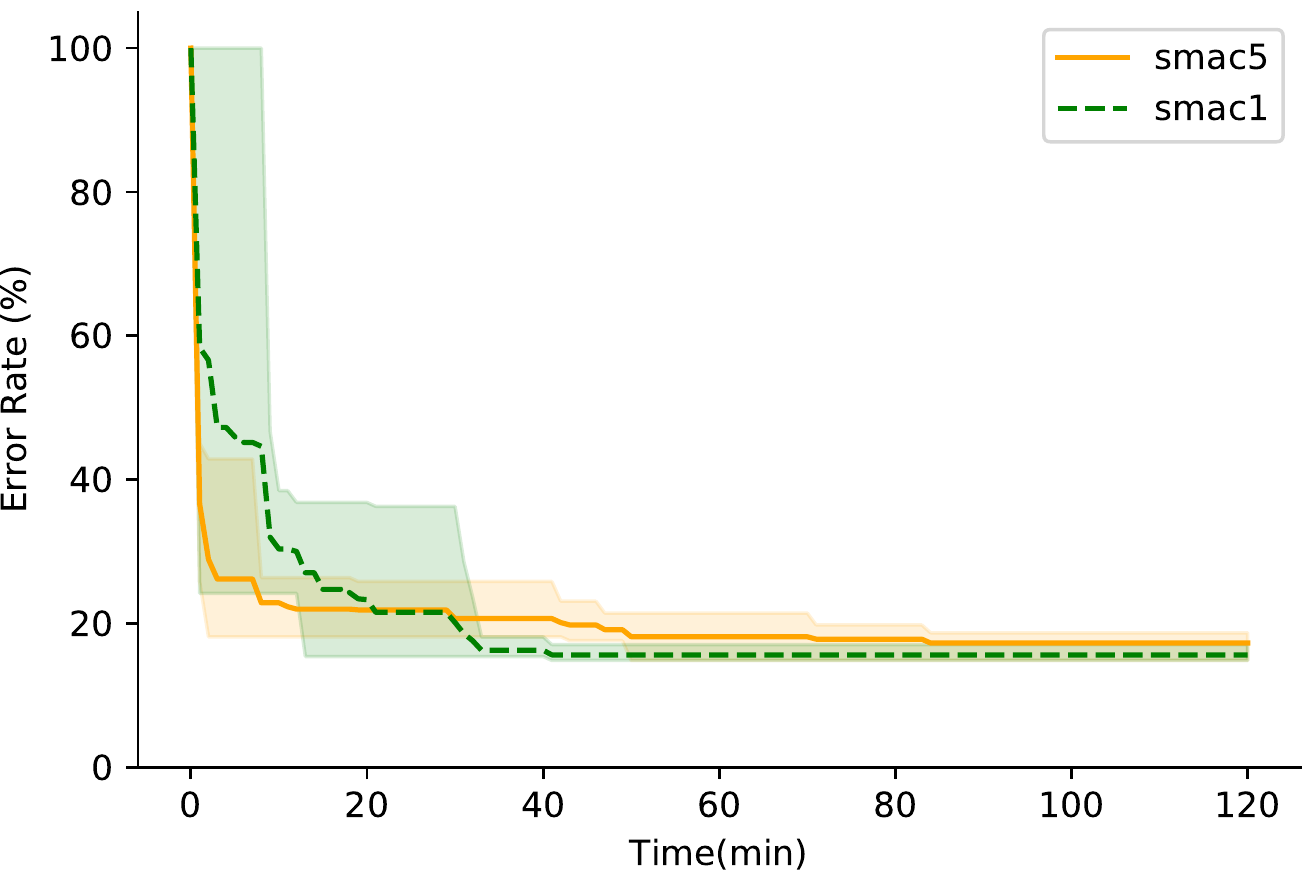}
        }%
     \hfill
     \subfloat[mnist]{%
        \includegraphics[width=0.45\linewidth]{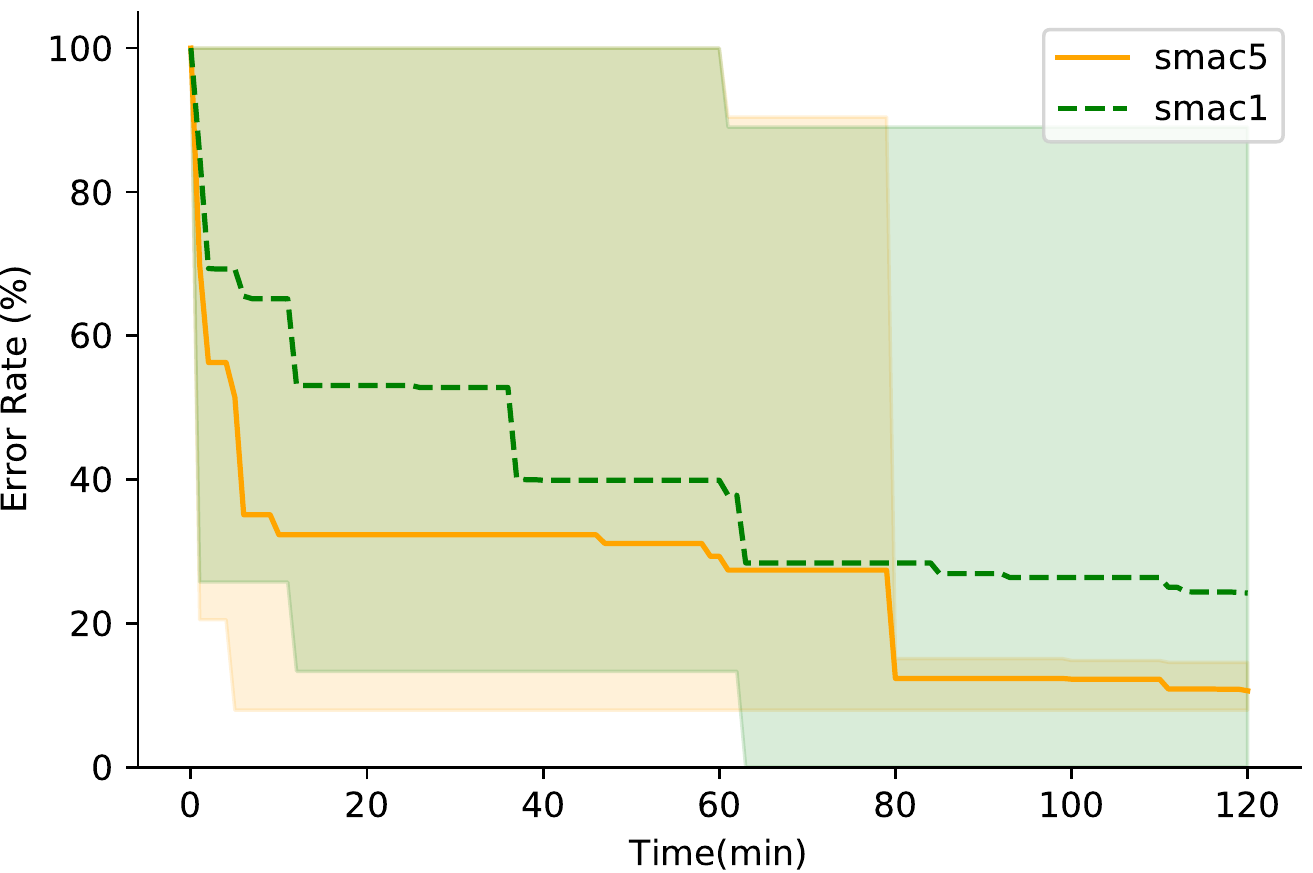}
      }%
    
    \subfloat[secom]{%
        \includegraphics[width=0.45\linewidth]{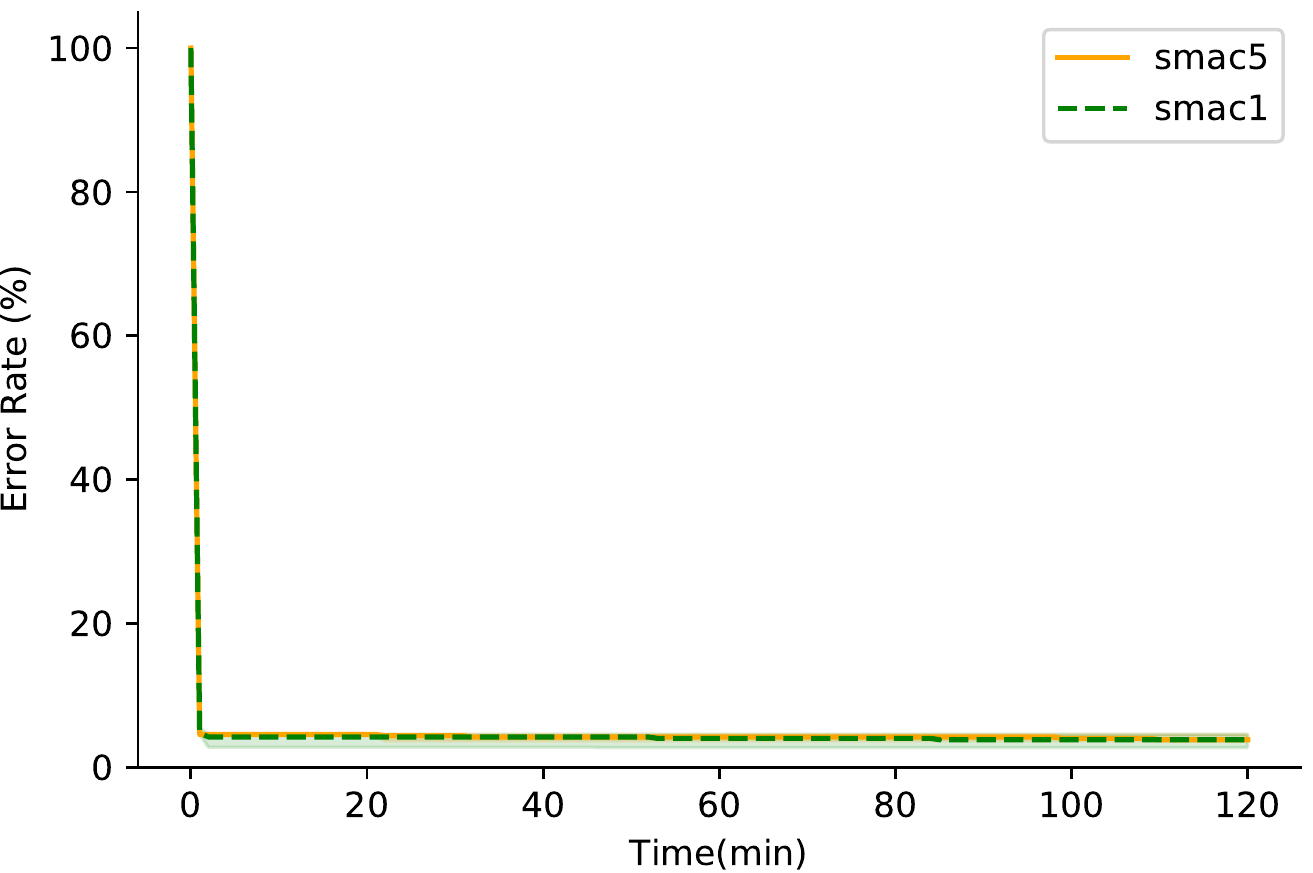}
        }%
     \hfill
     \subfloat[semeion]{%
        \includegraphics[width=0.45\linewidth]{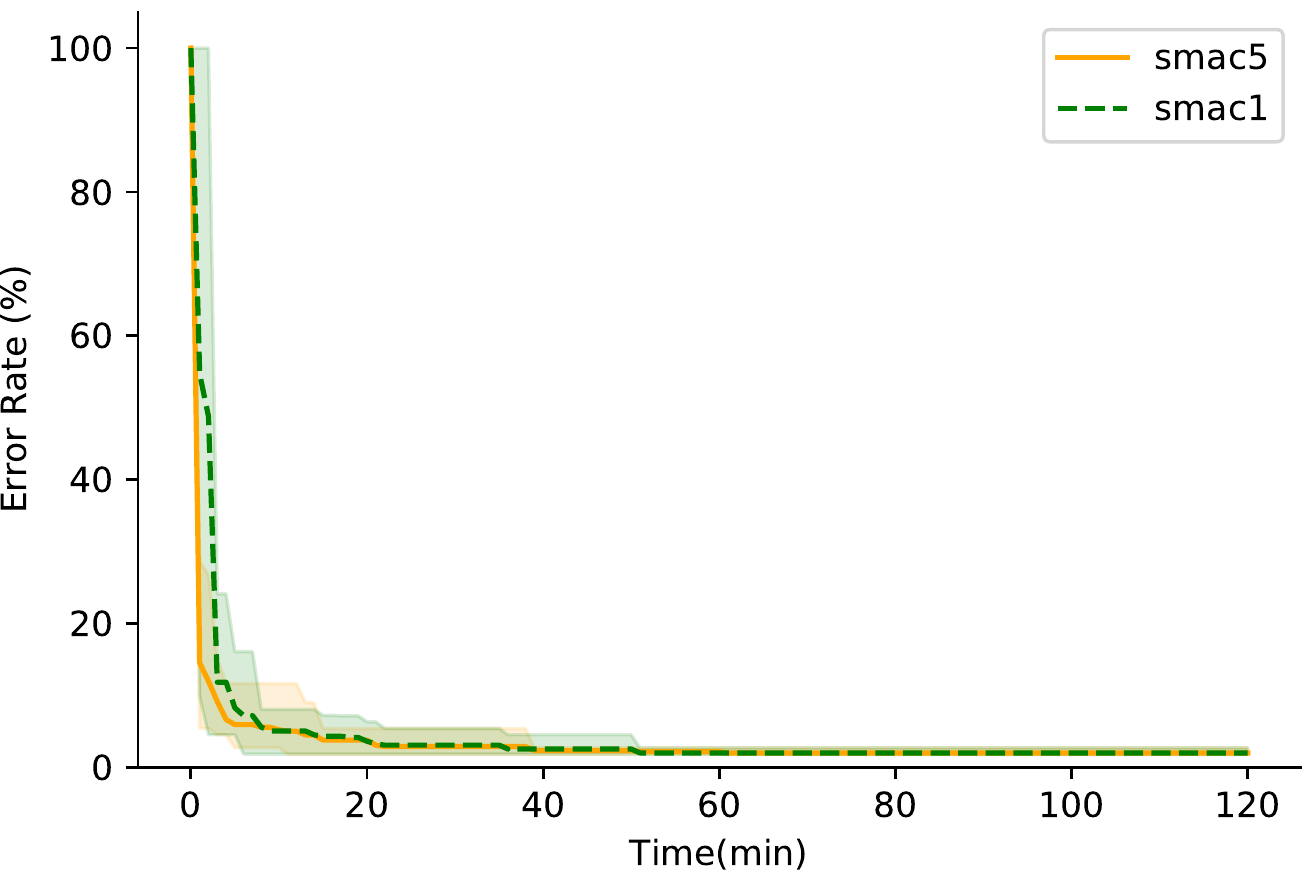}
      }%

 \caption{Compare the convergence of the error rate of the best pipelines found by SMAC1 and SMAC5.}
    \label{fig:exp_convergence_smac5_vs_smac1}
\end{figure}

 \begin{figure}[!htbp]     
  \ContinuedFloat

    \subfloat[shuttle]{%
        \includegraphics[width=0.45\linewidth]{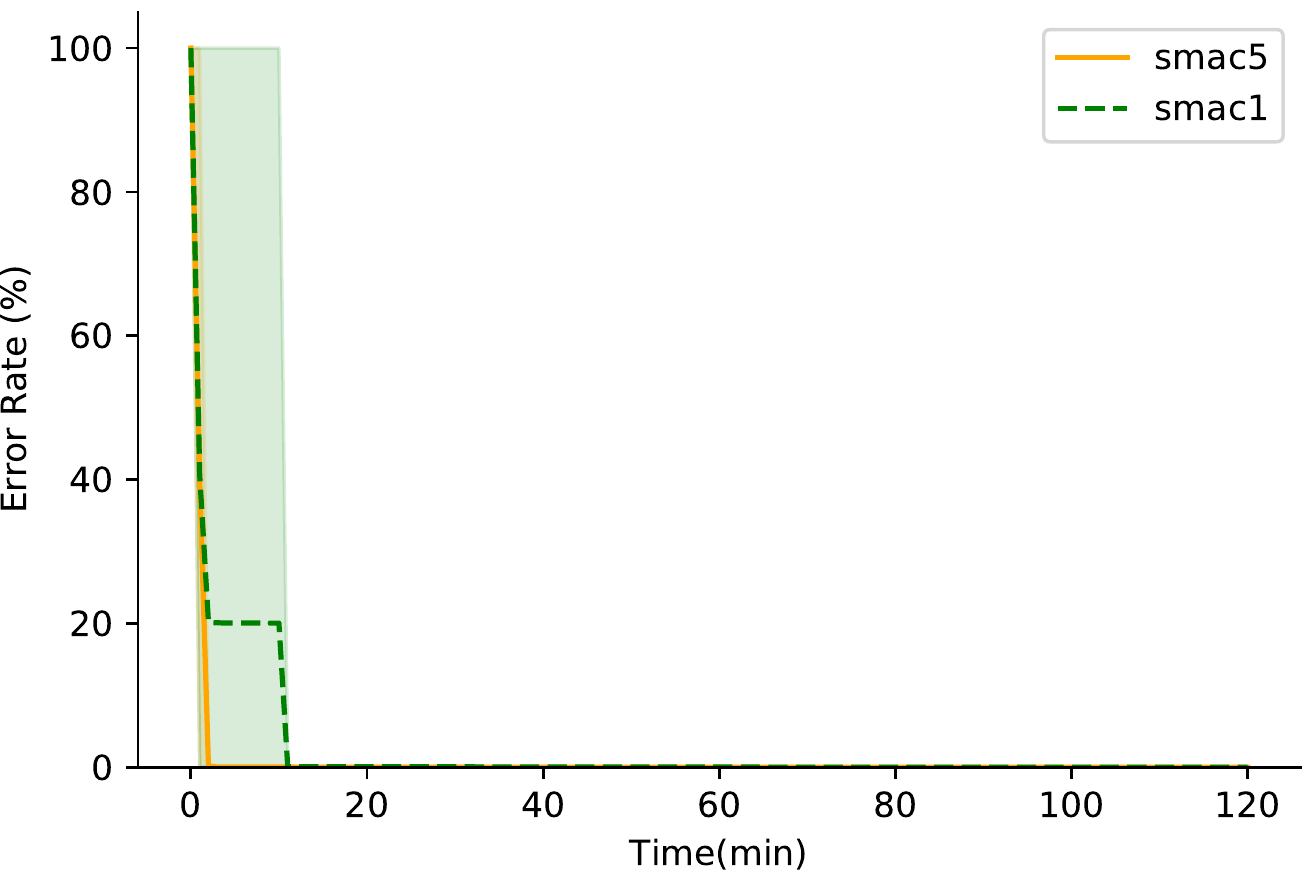}
        }%
    \hfill    
    \subfloat[waveform]{%
        \includegraphics[width=0.45\linewidth]{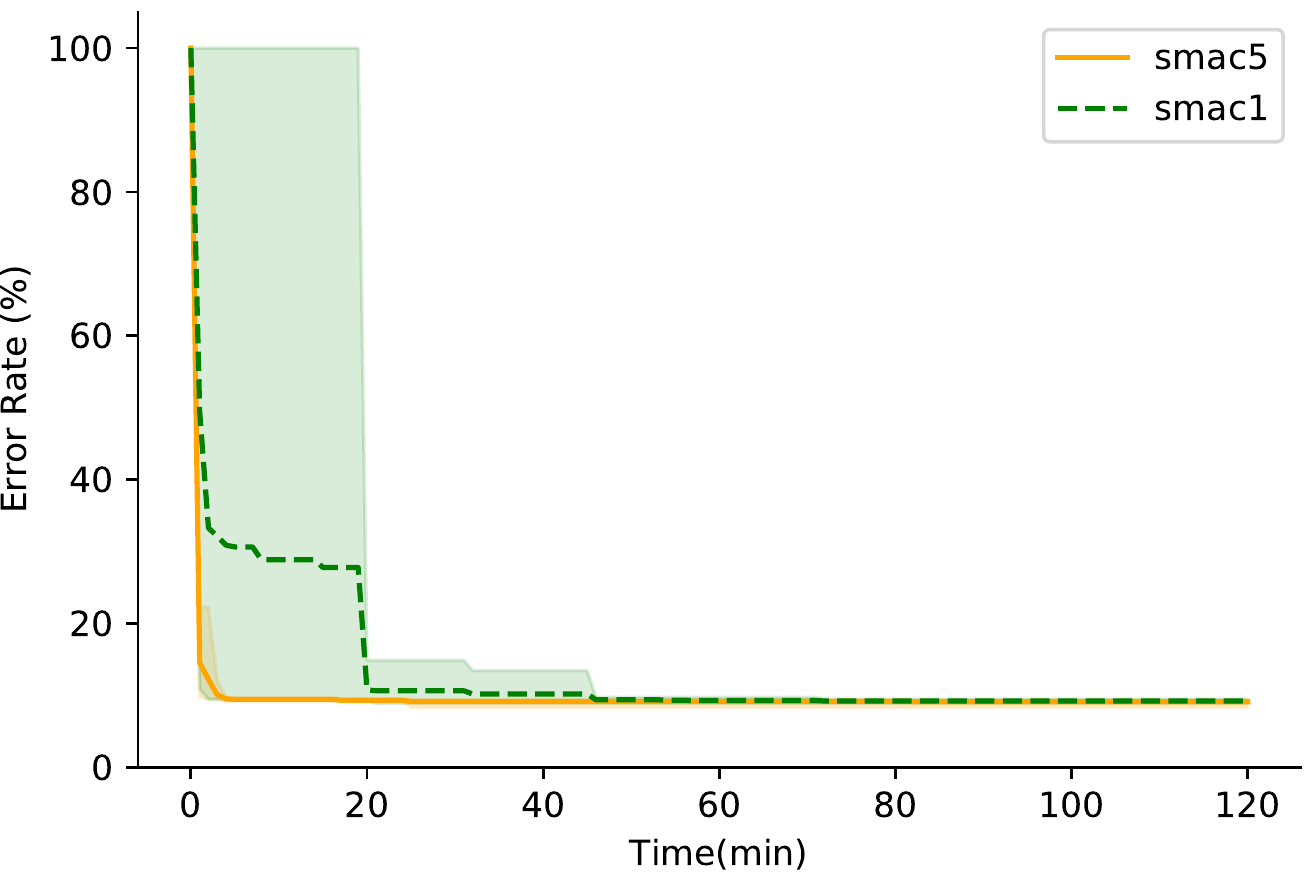}
      }%
    
    \subfloat[winequality]{%
        \includegraphics[width=0.45\linewidth]{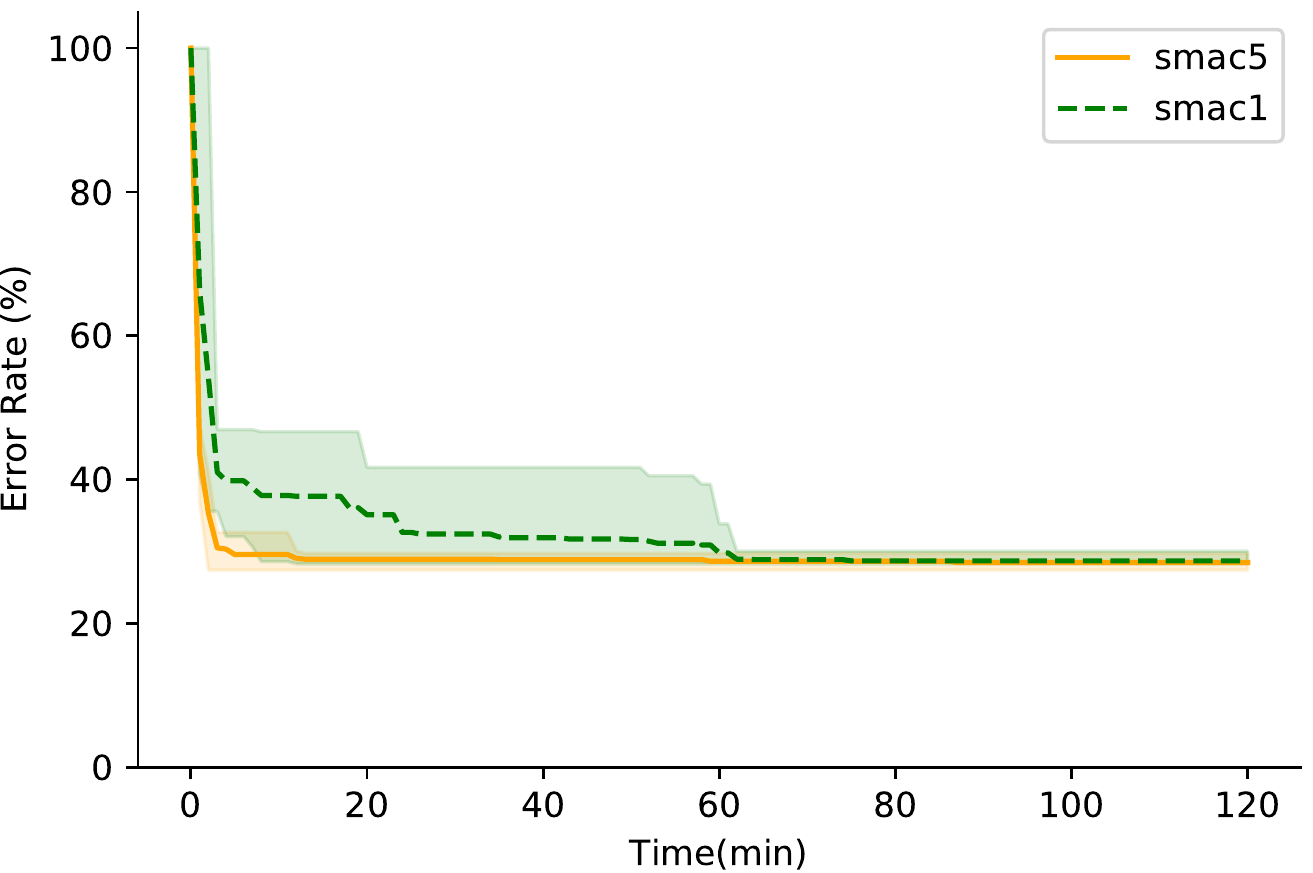}
        }%
        \hfill
       \subfloat[yeast]{%
        \includegraphics[width=0.45\linewidth]{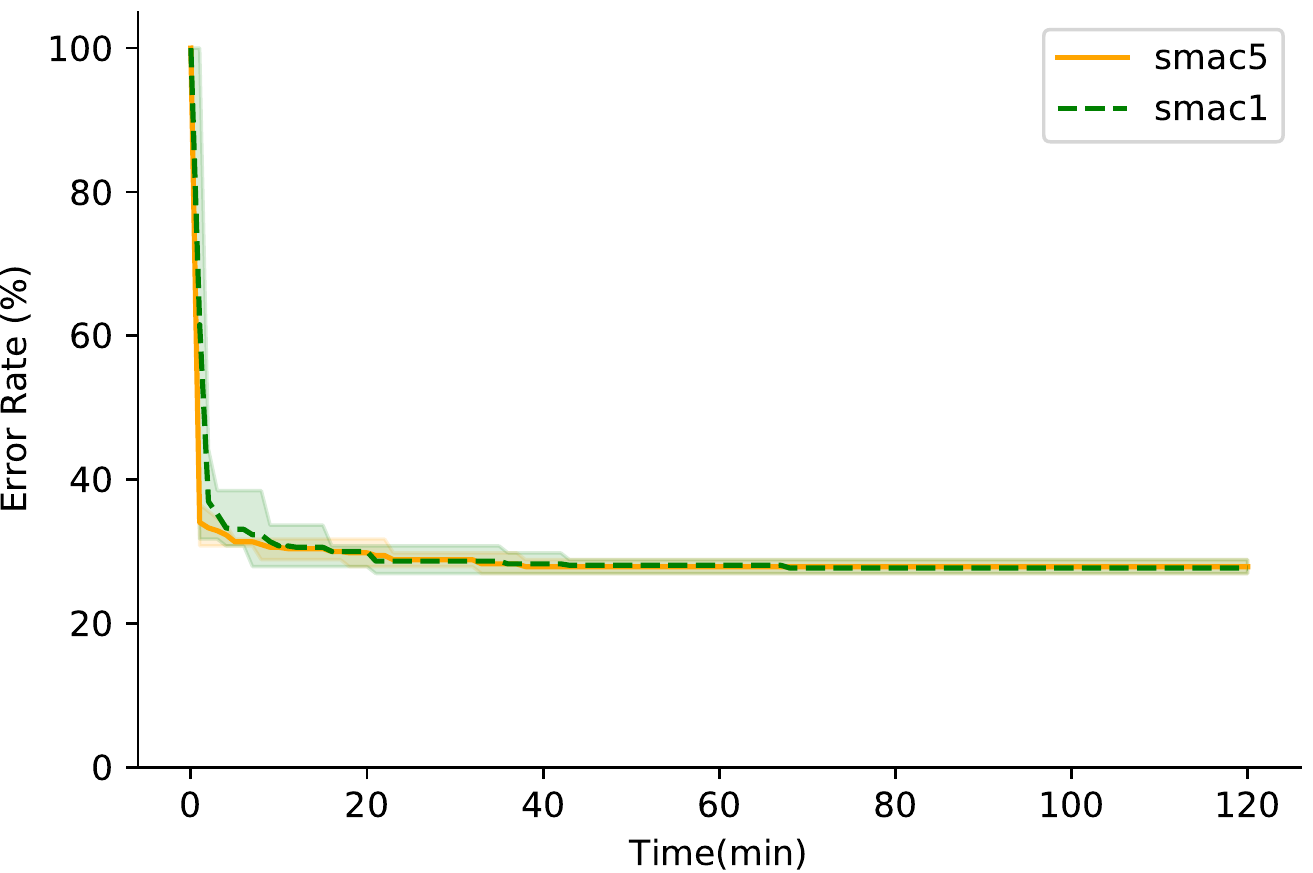}
      }%
        
    \caption{Compare the convergence of the error rate of the best pipelines found by SMAC1 and SMAC5.}
    \label{fig:exp_convergence_smac5_vs_smac1}
\end{figure}

\subsection{Experiments to investigate the impact the initialisation of SMAC with multiple configurations}
\subsubsection{Experiment Settings}
\label{sec:exp_settings}

Because the AVATAR enables SMAC to evaluate more promising pipelines within the same optimisation time budget and we have observed for a number of datasets that the convergence can be relatively quick, we take this advantage of the AVATAR to conduct the fourth set of experiments to investigate the impact of using multiple initialisations within a single 2-hour optimisation slot. In this set of experiments, we customise SMAC to be initialised with five random configurations instead of one and record the best found pipeline from such five optimisation runs. We use the same datasets as in the previous set of experiments. As previously, we also use SMAC as a pipeline composition and optimisation method. We evaluate the experiment results of the five iterations with and without the AVATAR. As in the previous experiments, we set the total time budget to 2 hours and the memory to 1GB per iteration. We compare the performance of SMAC with the initialisation of one (SMAC1) and five configurations (SMAC5) using the following criteria:

\begin{itemize}
    \item Mean, standard deviation and min of the error rate; 
    \item Mean, standard deviation and max number of the evaluated pipelines.
\end{itemize}

\subsubsection{Experiment Results}

Table \ref{tab:exp_smac1_vs_smac5} compares the performance of SMAC1 and SMAC5. We can see that the number of evaluated pipelines of SMAC1 and SMAC5 is similar.
In addition, the quality of the best pipelines found by SMAC1 and SMAC5 are also similar. Mean of the error rate of SMAC5 is higher than, equal and lower than SMAC1 in 11, 1 and 8 out of 20 datasets respectively. Minimum of the error rate of SMAC5 is higher than, equal and lower than SMAC1 in 8, 4 and 8 out of 20 datasets respectively.  Therefore, we can conclude that the quality of the best pipelines found by SMAC1 and SMAC5 is similar based on the mean and min error rate of the five iterations.
However, the standard deviation of the error rate for SMAC5 is lower than, equal and higher than SMAC1 in 12, 2 and 6 out of 20 datasets. It means that the convergence of SMAC5 is more stable than SMAC1. In other words, the initialisation of SMAC with multiple configurations within a single optimisation run is more likely to result in SMAC finding better ML pipelines in comparison to the single-configuration initialisation regardless of the input seed. Figure \ref{fig:exp_convergence_smac5_vs_smac1} visualises the convergence of the error rate of the best pipelines found by SMAC1 and SMAC5. The lower and upper boundary of the shaded area of the SMAC1 and SMAC5 are formed by the min and the max error rate of the five iterations respectively. The dashed and the solid lines represent the mean of the error rate of the best pipelines
of the five iterations found at the time points for SMAC1 and SMAC5 respectively.
These visualisations confirm that the stability of SMAC5 is better than SMAC1. A good example is the case of using the dataset \textit{kddcup}. Although the SMAC5 and SMAC1 can find the pipelines that have similar error rate after 120 minutes, SMAC5 found these pipelines after around 10 minutes, but SMAC1 requires more than 60 minutes to find pipelines that have a similar error rate.
Figure \ref{fig:exp_convergence_smac5_vs_smac1} also shows that the mean of the error rate of SMAC5 is lower than SMAC1 after the first 30 minutes of the optimisation time in 10 out of the 20 datasets (\textit{abalone}, \textit{amazon}, \textit{convex}, \textit{dexter}, \textit{dorothea}, \textit{gisette}, \textit{kddcup}, \textit{mnist}, \textit{waveform} and \textit{winequality}). For example, in case of the dataset \textit{mnist} , we can see that the mean of the error rate of SMAC5 is lower than SMAC1 after approximately 5 minutes of the time budget. After that, SMAC5 continues to find better pipelines that have lower mean error rate than pipelines found by SMAC1 after 30 minutes of the optimisation time. The reason is that the multiple configurations based initialisation enables SMAC to reduce a chance of selecting bad-performing pipelines at the initialisation stage which then leads to faster convergence of SMAC.
The mean of the error rate of SMAC5 is equal to SMAC1 after the first 30 minutes of the optimisation time in further 7 out of 20 datasets (\textit{car}, \textit{gcredit}, \textit{krvskp}, \textit{secom}, \textit{semeion}, \textit{shuttle} and \textit{yeast}). For example, in case of the dataset \textit{gcredit}, we can see that the means of the error rates of SMAC1 and SMAC5 are similar because this dataset is simple for SMAC to find well-performing pipelines regardless of the initialisation. 
The mean of the error rate of SMAC5 convergence is slower than SMAC1 after the first 30 minutes of the optimisation time in only 3 out of 20 datasets (\textit{adult}, \textit{cifar10small} and \textit{madelon}). This illustrates that using multiple configurations for the initialisation of SMAC is not a good solution in these cases as they require longer continuous optimisation time to find well-performing pipelines. This in fact is pointing to a more fundamental problem of identifying data sets which require longer optimisation time due to their complexity and/or size and better identification of potentially bad initialisation as the key reason for longer convergence times or variability of the quality of the found ML pipelines. This topic remains as one of our future research priorities.



\section{Conclusions}
\label{sec:conclusion}

In this study, we empirically demonstrated the challenges encountered when generating invalid pipelines during pipeline composition and optimisation process within automated machine learning. These challenges are how to find valid and well-performing pipelines in large configuration spaces.
The reason is that the configuration spaces of the pipelines' structures and hyperparameters of the pipelines' components expand significantly with the growing number of available methods and hyperparameters. For example, there are approximately eight billions unique pipelines if we limit the maximum length of a pipeline to eight components with the available methods in WEKA, one of the most popular ML libraries.
We showed that the wasted evaluation time of invalid pipelines is significant in the current ML composition and optimisation process. We propose the AVATAR which is a pipeline evaluation method using a surrogate model. The AVATAR can be used to accelerate pipeline composition and optimisation methods by quickly ignoring invalid pipelines to improve the effectiveness of the AutoML optimisation process. We demonstrate that SMAC can find better ML pipelines by using rather than not using the AVATAR within the same optimisation time. In addition, we leverage the advantage of the AVATAR's ability to quickly evaluate the validity of promising pipelines to initialise SMAC with multiple configurations. The experiments show that this approach can stabilise the quality of the best pipelines found by SMAC regardless of the random effects of the initialisation. 
In future, we intend to improve the AVATAR to estimate pipelines' quality besides their validity. Using various meta-learning approaches, as already illustrated for the purpose of warm start for AutoML in \cite{fekl15}, will be further explored to achieve that goal.

\bibliography{references}

\end{document}